\documentclass[letterpaper,11 pt]{article}
\usepackage[margin=1in]{geometry}
\usepackage{pifont}
\usepackage{soul}

\usepackage{xcolor}
\definecolor{mygreen}{rgb}{0.0, 0.5, 0.0}
\definecolor{winered}{rgb}{0.8,0,0}
\definecolor{myblue}{rgb}{0,0,0.8}
\definecolor{notered}{HTML}{d62728}
\definecolor{noteblue}{HTML}{1f77b4}
\definecolor{noteorange}{HTML}{ff7f0e}
\usepackage{multirow} 
\usepackage{booktabs}
\usepackage{threeparttable}
\usepackage{enumitem}
\usepackage{algorithm}
\usepackage{algpseudocode}
\usepackage{titletoc}

\usepackage{hyperref}
\hypersetup{
	colorlinks=true,
	linkcolor={winered},
	citecolor={mygreen}
}
\usepackage{graphicx}
\usepackage{wrapfig}
\usepackage{subcaption}
\usepackage{tikz}
\usepackage[round]{natbib}

\usepackage{amsmath,amsfonts,bm}
\usepackage{amsthm}
\usepackage{mathtools}

\def\1{\bm{1}}

\def\rvg{{\mathbf{g}}}

\def\rvx{{\mathbf{x}}}
\def\rvy{{\mathbf{y}}}

\def\vp{{\bm{p}}}
\def\vq{{\bm{q}}}

\def\vx{{\bm{x}}}
\def\vy{{\bm{y}}}
\def\vz{{\bm{z}}}

\def\mA{{\bm{A}}}

\def\mX{{\bm{X}}}
\def\mY{{\bm{Y}}}

\DeclareMathAlphabet{\mathsfit}{\encodingdefault}{\sfdefault}{m}{sl}
\SetMathAlphabet{\mathsfit}{bold}{\encodingdefault}{\sfdefault}{bx}{n}

\def\gA{{\mathcal{A}}}

\def\gD{{\mathcal{D}}}
\def\gE{{\mathcal{E}}}

\def\gK{{\mathcal{K}}}

\def\gO{{\mathcal{O}}}

\newcommand{\E}{\mathbb{E}}

\newcommand{\R}{\mathbb{R}}

\newcommand{\Var}{\mathrm{Var}}

\newcommand{\Cov}{\mathrm{Cov}}

\DeclareMathOperator*{\argmin}{arg\,min}

\newcommand\inp[2]{\left\langle #1, #2 \right\rangle} 
\newcommand{\norm}[1]{\left\lVert#1\right\rVert} 
\newcommand{\Norm}[1]{\lVert#1\rVert} 
\newcommand{\sqn}[1]{{\left\lVert#1\right\rVert}^2} 

\newcommand{\pfcomment}[1]{
	\tag*{$\triangleright$ #1}
}

\newtheorem{assumption}{Assumption}

\newtheorem{theorem}{Theorem}
\newtheorem{corollary}{Corollary}
\newtheorem{lemma}{Lemma}

\newcommand{\fronorm}[1]{\ensuremath{\norm{#1}_{\tiny{\mbox{F}}}}}

\title{\LARGE Convergence Analysis of Sequential Split Learning \\
	on Heterogeneous Data}
\author{Yipeng Li and Xinchen Lyu
	\thanks{Email: {\tt \{liyipeng, lvxinchen\}@bupt.edu.cn}. Beijing University of Posts and Telecommunications, China. \textcolor{notered}{Please refer to the new version ``Convergence Analysis of Sequential Federated Learning on Heterogeneous Data'', updated in a new location \url{https://arxiv.org/abs/2311.03154}, where we i) framed the whole paper into Sequential Federated Learning, ii) fixed some typos in the proof, and iii) updated the experiments (with detailed settings and public code). It's worth noting that the convergence analysis of Sequential Split Learning is identical to Sequential Federated Learning completely.}}}
\date{}

\begin{document}
	\maketitle
	\thispagestyle{empty}
	\pagestyle{empty}
	
\begin{abstract}
	Federated Learning (FL) and Split Learning (SL) are two popular paradigms of distributed machine learning. By offloading the computation-intensive portions to the server, SL is promising for deep model training on resource-constrained devices, yet still lacking of rigorous convergence analysis. In this paper, we derive the convergence guarantees of Sequential SL (SSL, the vanilla case of SL that conducts the model training in sequence) for strongly/general/non-convex objectives on heterogeneous data. Notably, the derived guarantees suggest that SSL is better than Federated Averaging (FedAvg, the most popular algorithm in FL) on heterogeneous data. We validate the counterintuitive analysis result empirically on extremely heterogeneous data.
\end{abstract}

\section{Introduction}
Federated Learning (FL) and Split Learning (SL) are two popular distributed machine learning paradigms where multiple clients collaborate to train a global model. The optimization problems of both paradigms with $M$ clients can be given by
\begin{align}
	\min_{\rvx\in \R^d} \left\{ F(\rvx) \coloneqq \frac{1}{M}\sum_{m=1}^M F_m(\rvx) \right\},\label{eq:sl optimization problem}
\end{align}
where $\rvx$ is the model parameter, $F$ denotes the global objective, $F_m$ denotes the local objective on local dataset $\gD_m$ at client $m$. In particular, $F_m$ is defined as
\[\textstyle
F_m(\rvx) \coloneqq \E_{\xi_m \sim \gD_m}[f_m(\rvx; \xi_m)]=\frac{1}{\lvert\gD_m\rvert}\sum_{i=1}^{\lvert\gD_m\rvert}f_m(\rvx;\xi_m^i),
\]
where $f_m$ is the loss and $\xi_m$ is a data sample randomly chosen from the local dataset $\gD_m$. Note that the unweighted global objective in Equation~\eqref{eq:sl optimization problem} can be extended to the weighted case readily.

In FL, after receiving the global model from the parameter server, each client would perform multiple local updates and then send the updated local parameters to the server. The server generates the global parameter by taking the weighted average on the local parameters (Alg.~\ref{algorithm:FedAvg}). In particular, each client needs to train a full (possibly) complex AI model locally. This is preferable in \textit{cross-silo settings} (small-scale clients with adequate resources, e.g., organizations), but may be unaffordable in \textit{cross-device settings} (massive resource-constrained clients, e.g., IoT devices) \citep{kairouz2021advances}.

 

To address the model-training resource bottleneck on resource-constrained devices, SL has emerged \citep{gupta2018distributed, vepakomma2018split}, where the AI model is split to be trained at the clients and server collaboratively. The computation-intensive portions are offloaded to the server. In particular, there are two typical algorithms in SL: (i) Sequential Split Learning (\texttt{SSL}) \citep{gupta2018distributed}, where clients train their local models in sequence; (ii) Split Federated Learning (\texttt{SFL}) \citep{thapa2020splitfed}, where clients train their local models in parallel and the global model is generated by federated averaging. The first version of \texttt{SFL} (\texttt{SFLV1}) is considered in this work.

\paragraph{Motivation.} Due to its advantage on model training on resource-constrained devices, recent work has sought to study the learning performance of SL on heterogeneous data. In particular, \cite{gao2020end, gao2021evaluation} compared the performance of FL and SL on heterogeneous data empirically, and found that (i) \texttt{FedAvg} (Federated Averaging \citep{mcmahan2017communication}) outperforms \texttt{SSL}; (ii) the performance of \texttt{SFLV1} is close to \texttt{FedAvg}. Yet their work was empirical, lacking theory, which prompts us to study the convergence theory of SL on heterogeneous data.

\paragraph{Our focus: convergence theory of \texttt{SSL}.} Convergence theory is critical for analyzing the learning performance of FL and SL. The existing convergence theory of FL \citep{li2019convergence, karimireddy2020scaffold} can apply to \texttt{SFLV1} readily given the same model averaging mechanism, thereby achieving similar performances as empirically validated by \cite{gao2020end, gao2021evaluation}. However, the convergence theory of \texttt{SSL} on heterogeneous data is much more complicated given the sequential training manner, is not investigated in the literature yet, and is the research question of our paper. To go further, we aim to compare the performance of \texttt{SSL} with \texttt{FedAvg} (Algorithm~\ref{algorithm:FedAvg}, the most popular algorithm in FL) by the convergence theory. In the following, we provide some preliminaries.

\paragraph{Update rule of \texttt{SSL}.} We give a simplified version of \texttt{SSL} in Algorithm~\ref{algorithm:SSL} to clarify the update rule, while deferring operation details to Appendix~\ref{sec:algs}. At the beginning of each training round, the indices $\pi_1, \pi_2, \ldots, \pi_M$ are sampled \textbf{without replacement} from $\{1,2,\ldots,M\}$ randomly (i.e., $\pi_1, \pi_2, \ldots, \pi_M$ is one random permutation) as clients' training order. In each round, the first client (i.e., $\pi_1$) receives the global model parameter. Then it performs $K$ steps of \textit{local updates} with its local data. The updated local parameter will be passed to the next client. This process continues until all the clients finish their local training. Let $\rvx_m^{(r, k)}$ denote the local parameter at client $\pi_m$ after $k$ local updates in the $r$-th round, and $\rvx^r$ denote the global parameter. With SGD (Stochastic Gradient Descent) as the local solver, the update rule of \texttt{SSL} can be written as
\begin{equation}
	\begin{aligned}
		&\textrm{Local update}: \rvx_{m}^{(r,k+1)} = \rvx_{m}^{(r,k)} - \eta \rvg_{\pi_m}^{(r,k)},\;\textrm{initializing as}\; \rvx_{m}^{(r,0)} =
		\begin{cases}
			\rvx^{r}, m=1\\
			\rvx_{m-1}^{(r,K)}, m>1
		\end{cases}\\
		&\textrm{Global model}: \rvx^{r+1} = \rvx_{M}^{(r,K)}
	\end{aligned}
\end{equation}
where $\rvg_{\pi_m}^{(r,k)} \coloneqq \nabla f_{\pi_m}(\rvx_{m}^{(r,k)};\xi_m)$ denotes the stochastic gradient of $F_{\pi_m}$ regarding parameters $\rvx_{m}^{(r,k)}$ and $\eta$ denotes the learning rate (or stepsize). Notations are summarized in Appendix~\ref{apx:notation}.

\begin{minipage}[t]{0.5\linewidth}
	\begin{algorithm}[H]
		\caption{Sequential Split Learning (\texttt{SSL})}\label{algorithm:SSL}
		\begin{algorithmic}[1]
			\Ensure $\bar{\rvx}^R$: weighted average on $\rvx^r$
			\For{training round $r = 0, 1,\ldots, R-1$}
			\State Sample a permutation $\pi_1, \pi_2, \ldots, \pi_{M}$ of $\{1,2,\ldots,M\}$
			\For{$m = 1,2,\ldots,M$ \textbf{\textcolor{winered}{in sequence}}}
			\State Initialize
			$
			\rvx_{m}^{(r,0)} =
			\begin{cases}
				\rvx^{r}, m=1\\
				\rvx_{m-1}^{(r,K)}, m>1
			\end{cases}
			$
			\For{local step $k = 0,\ldots, K-1$}
			\State
			$\rvx_{m}^{(r,k+1)} = \rvx_{m}^{(r,k)} - \eta \rvg_{\pi_m}^{(r,k)}$
			\EndFor
			\EndFor
			\State Global model: $\rvx^{r+1} = \rvx_{M}^{(r,K)}$
			\EndFor
		\end{algorithmic}
	\end{algorithm}
\end{minipage}
\hfill\hspace{0.5em}
\begin{minipage}[t]{0.49\linewidth}
	\begin{algorithm}[H]
		\caption{Federated Averaging (\texttt{FedAvg})}\label{algorithm:FedAvg}
		\begin{algorithmic}[1]
			\Ensure $\bar{\rvx}^R$: weighted average on $\rvx^r$
			\For{training round $r = 0,1,\ldots, R-1$}
			\For{$m = 1,2,\ldots,M$ \textbf{\textcolor{winered}{in parallel}}}
			\State Initialize $\rvx_{m}^{(r,0)} = \rvx^r$
			\For{local step $k = 0,\ldots, K-1$}
			\State
			$\rvx_{m}^{(r,k+1)} = \rvx_{m}^{(r,k)} - \eta \rvg_{m}^{(r,k)}$
			\EndFor
			\EndFor
			\State Global model: \(\displaystyle \rvx^{r+1} = \frac{1}{M} \sum_{m=1}^M \rvx_m^{(r,K)} \)
			\EndFor
		\end{algorithmic}
	\end{algorithm}
\end{minipage}

\section{Contributions}

\paragraph{Brief literature review.} The most relevant work is the convergence theory of Random Reshuffling and \texttt{FedAvg}. Random Reshuffling (RR) is one sibling algorithm of SGD, where the training samples are sampled without replacement. RR is deemed to be more practical than SGD (sampling with replacement). However, it would cause a challenge that the gradients are not (conditionally) unbiased. Recently, much work \citep{safran2020good, mishchenko2020random, safran2021random} tried to derive the convergence guarantee of RR. There is a wealth of work has analyzed the convergence of \texttt{FedAvg} (or local SGD) on homogeneous data \citep{stich2019local, zhou2017convergence, khaled2020tighter}, heterogeneous data \citep{li2019convergence, khaled2020tighter, karimireddy2020scaffold} and unbalanced data \citep{wang2020tackling}.

\paragraph{Challenges.} This is the first work to derive the convergence guarantee of \texttt{SSL} on heterogeneous data. The guarantee of SSL on homogeneous data is trival (can be reduced to the case of SGD). However, on heterogeneous data, the stochastic gradient at any client $m$ is not an (conditionally) unbiased estimater of the global objective, i.e., $\E[\nabla f_{\pi_m}(\rvx_m^{(r,k)};\xi_m)\vert\rvx_m^{(r,k)}] =\nabla F_{\pi_m}(\rvx_m^{(r,k)}) \neq \nabla F(\rvx_m^{(r,k)})$, thus it cannot follow the theory of SGD. The challenges of \texttt{SSL} mainly arise from (i) sequential training across clients and (ii) multiple local updates with SGD on each client.

Sequential training across clients (vs \texttt{FedAvg}). In \texttt{FedAvg}, models are updated independently within each round and synchronized in the end of each round to generate the global model. At this case, clients update their local model parameters on the global model. However, in \texttt{SSL}, clients (except the first client) update their local model parameters on the local model of their previous client. This complicates our derivation of per-round recursion when conditioning on the global model.

Multiple local updates with SGD on each client (vs Random Reshuffling). Recall that RR samples the training samples without replacement, which is similar to \texttt{SSL}'s sampling the clients. In fact, RR can be regarded as a special case of \texttt{SSL} where only one local update with GD is performed on each training sample (client in \texttt{SSL}). Multiple local updates with SGD would significantly complicate our derivation of convergence guarantees.

\paragraph{Contributions.}
The main contributions are summarized as follows:
\begin{itemize}[leftmargin=1em]
	\item We derive the convergence guarantee of \texttt{SSL} for strongly convex, general convex and non-convex objectives on heterogeneous data with the standard assumptions in Section~\ref{sec:analyses on SSL}. As far as we know, this work is the first to give the convergence guarantee of \texttt{SSL}.
	\item We compare the convergence guarantees of \texttt{FedAvg} and \texttt{SSL}, and provide a \textit{counterintuitive} comparison result that the guarantee of \texttt{SSL} is better than \texttt{FedAvg} with full participation and partial participation in terms of training rounds on heterogeneous data in Section~\ref{sec:SSL vs FedAvg}.
	\item We validate our counterintuitive comparison results with simulations on quadratic functions (Section~\ref{sec:feasibility analysis}) and experiments on real datasets (Section~\ref{sec:exp}). The experimental results show that \texttt{SSL} outperforms \texttt{FedAvg} on extremely heterogeneous data, especially in cross-device settings.
\end{itemize}

\section{Convergence theory}\label{sec:convergence}
In our convergence theory, three cases are considered, i.e., the strongly convex case, the general convex case and the non-convex cases, where the global objective $F$ and local objectives $\{F_m\}_{m=1}^M$ are \textbf{$\mu$-strongly convex}, \textbf{general convex} ($\mu=0$) and \textbf{non-convex}.

\subsection{Assumptions}
We assume that (i) $F$ is lower bounded by $F^\ast$ for all cases and there exists a minimizer $\rvx^\ast$ such that $F(\rvx^\ast)=F^\ast$ for strongly and general convex cases; (ii) each local objective function is $L$-smooth (Assumption~\ref{asm:L-smoothness}). Furthermore, we need to make assumptions on the diversities: (iii) the assumptions on the stochasticity bounding the diversities of $\{f_m(\cdot;\xi_m^i)\}_{i=1}^{\lvert\gD_m\rvert}$ with respect to $i$ inside any client $m$ (Assumption~\ref{asm:stochasticity}); (4) the assumptions on the heterogeneity bounding the diversities of $\{F_m\}_{m=1}^M$ with respect to $m$ (Assumption~\ref{asm:heterogeneity}).

\begin{assumption}[$L$-Smoothness]\label{asm:L-smoothness}
	Each local objective function $F_m$ is $L$-smooth, $m \in \{1,2,\ldots,M\}$, i.e., there exists a constant $L>0$ such that $\norm{\nabla F_m(\rvx) - \nabla F_m(\rvy)} \leq L \norm{\rvx - \rvy}$ for all $\rvx,\rvy \in \R^d$.
\end{assumption}

\textbf{Assumptions on the stochasticity.} Note that both in Algorithms~\ref{algorithm:SSL} and \ref{algorithm:FedAvg}, the local model is updated with multiple steps of SGD, i.e., the data samples are sampled with replacement. As a result, the stochastic gradient $\nabla f_m(\rvx_m;\xi_m)$ generated at any client $m$ is an (conditionally) unbiased estimate of the gradient of local objective function $\E_{\xi_m} \left[\nabla f_m(\rvx_m;\xi_m)\vert\rvx_m\right] = \nabla F_m(\rvx_m)$. Then we have Assumption~\ref{asm:stochasticity}, where $\sigma^2$ measures the level of stochasticity.

\begin{assumption}\label{asm:stochasticity}
	The variance of the stochastic gradient at each client $m$ is bounded:
	\begin{align}
		\E_{\xi_m}\left[\norm{\nabla f_m(\rvx_m;\xi_m) - \nabla F_m(\rvx_m)}^2\vert\rvx_m\right] \leq \sigma^2, \quad\forall\, m \in \{1,2,\ldots,M\}\label{eq:asm:stochasticity}
	\end{align}
\end{assumption}

\textbf{Assumptions on the heterogeneity.} Now we make assumptions on the dissimilarities of the local objective functions in Assumption~\ref{asm:heterogeneity}, also known as the heterogeneity in FL. The assumption~\eqref{eq:asm:heterogeneity:everywhere} is made for non-convex cases, where the constants $B^2$ and $\zeta^2$ measure the heterogeneity of the local objective functions, and they equal zero when all the local objective functions are identical to each other. Further, if the local objective functions are strongly or general convex, we use one weaker assumption \eqref{eq:asm:heterogeneity:optimum} as \cite{koloskova2020unified} did, which only bounds the dissimilarities at the optima.

\begin{assumption}\label{asm:heterogeneity}
	There exist constants $B^2$ and $\zeta^2$ such that
	\begin{align}
		\textstyle
		\frac{1}{M}\sum_{m=1}^M \norm{\nabla F_m(\rvx)-\nabla F(\rvx)}^2 \leq B^2\norm{\nabla F(\rvx)}^2 + \zeta^2\label{eq:asm:heterogeneity:everywhere}
	\end{align}
	Or further, it only holds at the optimum. Formally, there exists one constant $\zeta_\ast^2$ such that
	\begin{align}
		\textstyle
		\frac{1}{M}\sum_{m=1}^M \norm{\nabla F_m(\rvx^\ast)}^2 = \zeta_\ast^2\label{eq:asm:heterogeneity:optimum}
	\end{align}
	where $\rvx^\ast \in \argmin_{\rvx\in \R^d} F(\rvx)$ is one global minimizer.
\end{assumption}

\subsection{Convergence analysis of \texttt{SSL}}\label{sec:analyses on SSL}
\begin{theorem}\label{thm:SSL convergence}
	For \texttt{SSL}, there exists a constant effective learning rate $\tilde\eta \coloneqq MK\eta$, making the weighted average of the model parameters $\bar{\rvx}^R\coloneqq \frac{1}{W_R}\sum_{r=0}^{R}w_r\rvx^r$ ($W_R = \sum_{r=0}^Rw_r$) satisfy: 
	\begin{itemize}[leftmargin=1em]
		\item \textbf{Strongly convex}: Under Assumptions~\ref{asm:L-smoothness}, \ref{asm:stochasticity}, \eqref{eq:asm:heterogeneity:optimum} of \ref{asm:heterogeneity}, there exist one constant effective learning rate $\frac{1}{\mu R}\leq \tilde\eta \leq \frac{1}{6L}$ and weights $w_r=(1-\frac{\mu\tilde\eta}{2})^{-(r+1)}$, making it hold that
		\begin{flalign}
			\E\left[F(\bar\rvx^R)-F(\rvx^\ast)\right] \leq 3\mu\norm{\rvx^{0}- \rvx^*}^2 \exp\left(-\frac{\mu\tilde\eta R}{2}\right) + \frac{6\tilde\eta\sigma^2}{MK} + \frac{45L\tilde\eta^2\sigma^2}{4MK} + \frac{15L\tilde\eta^2\zeta_\ast^2}{2M}\label{eq:thm:ssl strongly convex} &&
		\end{flalign}
		\item \textbf{General convex}: Under Assumptions~\ref{asm:L-smoothness}, \ref{asm:stochasticity}, \eqref{eq:asm:heterogeneity:optimum} of \ref{asm:heterogeneity}, there exist one constant effective learning rate $\tilde\eta \leq \frac{1}{6L}$ and weights $w_r=1$, making it hold that
		\begin{flalign}
			\E\left[F(\bar\rvx^R)-F(\rvx^\ast)\right] \leq \frac{2\norm{\rvx^{0}- \rvx^\ast}^2}{\tilde\eta R} + \frac{6\tilde\eta\sigma^2}{MK} + \frac{45L\tilde\eta^2\sigma^2}{4MK} + \frac{15L\tilde\eta^2\zeta_\ast^2}{2M} &&
		\end{flalign}
		\item \textbf{Non-convex}: Under Assumptions~\ref{asm:L-smoothness}, \ref{asm:stochasticity}, \eqref{eq:asm:heterogeneity:everywhere} of \ref{asm:heterogeneity}, there exist one constant effective learning rate $\tilde\eta \leq \frac{1}{6L(B^2+1)}$ and weights $w_r=1$, making it hold that
		\begin{flalign}
			\min_{0\leq r\leq R}\E\norm{\nabla F(\rvx)}^2 \leq \frac{3\E [F(\rvx^{0}) - F^\ast]}{\tilde\eta R} + \frac{3L\tilde\eta\sigma^2}{MK} + \frac{27L^2\tilde\eta^2\sigma^2}{8MK} + \frac{9L^2\tilde\eta^2\zeta^2}{4M} &&
		\end{flalign}
	\end{itemize}
\end{theorem}

\textbf{The effective learning rate} $\tilde\eta \coloneqq MK\eta$ is used in the upper bound as \cite{karimireddy2020scaffold, wang2020tackling} did. All these upper bounds consist of two parts: the optimization part (the 1-st term) and the error part (the 2, 3, 4-th terms). Setting $\tilde\eta$ large can make the optimization part vanishes at a higher rate, while causing the error part to be larger. This implies that we need to choose an appropriate $\tilde\eta$ to achieve a balance between these two parts, which is actually done in Corollary~\ref{cor:SSL convergence}. Here we adopt a prior knowledge of the total training rounds $R$ as done in the previous work \citep{karimireddy2020scaffold, reddi2020adaptive} to choose the learning rate.

\begin{corollary}\label{cor:SSL convergence}
	 Applying the results of Theorem~\ref{thm:SSL convergence}. By choosing a appropriate learning rate (see the proof of Theorem~\ref{thm:SSL convergence} in Appendix~\ref{sec:proof SSL}), we can obtain the convergence rates for \texttt{SSL} as follows:
	\begin{itemize}[leftmargin=0.8em]
		\item \textbf{Strongly convex}: Under Assumptions~\ref{asm:L-smoothness}, \ref{asm:stochasticity}, \eqref{eq:asm:heterogeneity:optimum} of \ref{asm:heterogeneity}, there exist one constant effective learning rate $\frac{1}{\mu R}\leq \tilde\eta \leq \frac{1}{6L}$ and weights $w_r=(1-\frac{\mu\tilde\eta}{2})^{-(r+1)}$, making it hold that
		\begin{flalign}
			\E\left[F(\bar\rvx^R)-F(\rvx^\ast)\right] = \tilde\gO\left(\frac{\sigma^2}{\mu MKR} + \frac{L\sigma^2}{\mu^2MKR^2} + \frac{L\zeta_\ast^2}{\mu^2MR^2} + \mu D^2 \exp\left(-\frac{\mu R}{12L}\right)\right)\label{eq:cor:ssl strongly convex} &&
		\end{flalign}
		\item \textbf{General convex}: Under Assumptions~\ref{asm:L-smoothness}, \ref{asm:stochasticity}, \eqref{eq:asm:heterogeneity:optimum} of \ref{asm:heterogeneity}, there exist one constant effective learning rate $\tilde\eta \leq \frac{1}{6L}$ and weights $w_r=1$, making it hold that
		\begin{flalign}
			\E\left[F(\bar\rvx^R)-F(\rvx^\ast)\right] = \gO\left(\frac{\sigma D}{\sqrt{MKR}} + \frac{\left(L\sigma^2D^4\right)^{1/3}}{(MK)^{1/3}R^{2/3}} + \frac{\left(L\zeta_\ast^2D^4\right)^{1/3}}{M^{1/3}R^{2/3}} + \frac{LD^2}{R}\right) &&
		\end{flalign}
		\item \textbf{Non-convex}: Under Assumptions~\ref{asm:L-smoothness}, \ref{asm:stochasticity}, \eqref{eq:asm:heterogeneity:everywhere} of \ref{asm:heterogeneity}, there exist one constant effective learning rate $\tilde\eta \leq \frac{1}{6L(B^2+1)}$ and weights $w_r=1$, making it hold that
		\begin{flalign}
			\min_{0\leq r\leq R}\E\Norm{\nabla F(\rvx^r)}^2 = \gO\left(\frac{ \left(\sigma^2LA\right)^{1/2}}{\sqrt{MKR}} + \frac{\left(L^2\sigma^2A^2\right)^{1/3}}{(MK)^{1/3}R^{2/3}} + \frac{\left(L^2\zeta^2A^2\right)^{1/3}}{M^{1/3}R^{2/3}} + \frac{LB^2A}{R}\right) &&
		\end{flalign}
	\end{itemize}
	where $\gO$ omits absolute constants, $\tilde\gO$ omits absolute constants and polylogarithmic factors, $D\coloneqq\norm{x^0-x^\ast}$ for convex cases and $A \coloneqq F(\rvx^0) - F^\ast$ for the non-convex case.
\end{corollary}

\paragraph{Convergence rate.} By Corollary~\ref{cor:SSL convergence}, for sufficiently large $R$, the convergence rate is determined by the first term for all cases, resulting in convergence rates of $\tilde\gO(1/MKR)$ for strongly convex cases, $\gO(1/\sqrt{MKR})$ for general convex cases and $\gO(1/\sqrt{MKR})$ for non-convex cases. 

As said before, Random Reshuffling can be seen as one special case of \texttt{SSL}, where one step of GD is performed on each local objective $F_m$, i.e, $K=1$ and $\sigma=0$. Let us borrow the convergence guarantee from \cite{mishchenko2020random} (their Corollary 1),
\[
\text{Strongly convex:} \quad\E\norm{\rvx^R-\rvx^\ast}^2 = \tilde\gO\left(\frac{L\zeta_\ast^2}{\mu^3MR^2} + \norm{x^0-x^\ast}^2\exp\left(-\frac{\mu {\color{red}M}R}{L}\right)\right).\\
\]
As we can see, our bound turns to $\tilde\gO\left(\frac{L\zeta_\ast^2}{\mu^2MR^2} + \mu \norm{x^0-x^\ast}^2 \exp\left(-\frac{\mu R}{L}\right)\right)$ when $K=1$ and $\sigma=0$. The bound of Random Reshuffling only has an advantage on the second term (marked in red). The difference on the constant $\mu$ is because their bound is for $\E\norm{\rvx^R-\rvx^\ast}^2$ (see \cite{stich2019unified}). As a result, following a similar analysis in \cite{mishchenko2020random}, our bound matches the lower bound $\Omega\left(\min \{1, \frac{\zeta_\ast^2}{\mu^2 MR^2}\}\right)$ given by \cite{safran2020good}. For the general convex case, we also match the result in \cite{mishchenko2020random} (see their Corollary 2). These all suggest \textbf{our bounds are tight}. Yet a specialized lower bound for \texttt{SSL} is still required.


\paragraph{Effect of local steps.} Two comments are included: (i) It can be seen that \textit{local updates can help the convergence} with proper learning rate choices (small enough) by Corollary~\ref{cor:SSL convergence}. Yet this increases the total steps (iterations), leading to a higher computation cost. (ii) \textit{Excessive local updates do not benefit the convergence rate}. Take the strongly convex case as an example. It can be seen that large value of $K$ benefits the dominated term in \eqref{eq:cor:ssl strongly convex}, yet this will not hold when $\frac{\sigma^2}{\mu MKR} \leq \frac{L\zeta_\ast^2}{\mu^2MR^2}$, i.e., the latter turns decisive. In other words, when the value of $K$ exceed $\tilde\Omega\left(\sigma^2/\zeta_\ast^2\cdot \mu/L \cdot R\right)$, increasing local updates will not help the convergence rate. Besides, the maximum value of $K$ is affected by $\sigma^2/\zeta_\ast^2$, $\mu/L$ and $R$. This analysis follows \cite{reddi2020adaptive,khaled2020tighter}.

\subsection{\texttt{SSL} vs \texttt{FedAvg} on heterogeneous data}\label{sec:SSL vs FedAvg}

\begin{table}[h]
	\centering
	\begin{threeparttable}[b]
		\renewcommand{\arraystretch}{1.2}
		\setlength{\tabcolsep}{1.5em}
		\caption{Guarantees in the strongly convex case with absolute constants and polylogarithmic factors omitted. All results are for \textbf{heterogeneous settings}.}
		\label{table:summary for strongly convex case}
		\begin{tabular}{ll}
			\toprule
			Method &Bound ($D=\norm{x^0-x^\ast}$)\\
			\midrule
			SGD \citep{stich2019unified} &$\frac{\sigma^2}{\mu MKR} + LD^2\exp\left(-\frac{\mu R}{L}\right)$ \tnote{\color{red}(1)} \\[1ex]
			\midrule
			\texttt{FedAvg} &\\[-1ex]
			\;\qquad \citep{karimireddy2020scaffold} &$\frac{\sigma^2}{\mu MKR} + \frac{L\sigma^2}{\mu^2KR^2} + \frac{L\zeta^2}{\mu^2 R^2} + \mu D^2 \exp\left(-\frac{\mu R}{L}\right)
			$\tnote{\color{red}(2)}\\[1.2ex]
			\;\qquad \citep{koloskova2020unified} &$\frac{\sigma_\ast^2}{\mu MKR} + \frac{L\sigma_\ast^2}{\mu^2KR^2} + \frac{L\zeta_\ast^2}{\mu^2 R^2} + LKD^2 \exp\left(-\frac{\mu R}{L}\right)
			$\tnote{\color{red}(3)}\\[1.2ex]
			\;\qquad Theorem~\ref{thm:fedavg convergence} &$\frac{\sigma^2}{\mu MKR} + \frac{L\sigma^2}{\mu^2KR^2} + \frac{L\zeta_\ast^2}{\mu^2 R^2} + \mu D^2 \exp\left(-\frac{\mu R}{L}\right)
			$\\[1.2ex]
			\midrule
			\texttt{SSL} &\\[-1ex]
			\;\qquad Theorem~\ref{thm:SSL convergence} &$\frac{\sigma^2}{\mu MKR} + \frac{L\sigma^2}{\mu^2{\color{red}M}KR^2} + \frac{L\zeta_\ast^2}{\mu^2{\color{red}M}R^2} + \mu D^2 \exp\left(-\frac{\mu R}{L}\right)$\\[1.2ex]
			\bottomrule
		\end{tabular}
		\begin{tablenotes}
			\begin{small}
				\item [\tnote{\color{red}(1)}] SGD with a large batch size of $MK$, by letting $\sigma=\sigma/MK$ in the result of \cite{stich2019unified}.
				\item [\tnote{\color{red}(2)}] Here we do not consider the global learning rate. They use $\frac{1}{M}\sum_{m=1}^M \norm{\nabla F_m(\rvx)}^2 \leq B^2\norm{\nabla F(\rvx)}+ \zeta^2$ to bound the heterogeneity, which is equivalent to \eqref{eq:asm:heterogeneity:everywhere}.
				\item [\tnote{\color{red}(3)}] $\sigma_\ast^2$ is defined as $\sigma_\ast^2 \coloneqq \frac{1}{M}\sum_{m=1}^M \E_{\xi_m} \norm{\nabla f_m(\rvx^\ast;\xi_m)-\nabla F_m(\rvx^\ast)}^2$, to bound the stochasticity.
			\end{small}
		\end{tablenotes}
	\end{threeparttable}
\end{table}

\paragraph{Fair comparison in terms of training rounds.} Without otherwise stated, our comparison is in terms of training rounds, which also adopted in \cite{thapa2020splitfed, gao2020end, gao2021evaluation}. This comparison (running for the same total training rounds $R$) is fair given the same total computation cost (including the computation cost on client-side and server-side). We do not compare the communication cost, since there is communication in the local update stage in all SL algorithms, including \texttt{SSL} and \texttt{SFLV1}. The communication cost varies for different applications and settings \citep{singh2019detailed}. We do not compare the training time, since federated algorithms (including \texttt{FedAvg}, \texttt{SFLV1}) are trained in parallel, showing an inherent advantage over sequential algorithms. At last, we would like to stress that our comparison results also apply to the case \texttt{SFLV1} vs \texttt{SSL}, as \texttt{FedAvg} and \texttt{SFLV1} share the same update rules and convergence theory. 

\paragraph{Convergence results of \texttt{FedAvg}.} We summarize the existing convergence results of \texttt{FedAvg} for strongly convex cases in Table~\ref{table:summary for strongly convex case}. Here we slightly improve the convergence result for strongly convex cases by combining the works of \cite{karimireddy2020scaffold,koloskova2020unified}. \cite{woodworth2020minibatch} provided a tighter bound for general convex cases with a much stronger assumption on the heterogeneity (their Theorem 3), so we do not include it. Besides, we note that to derive a unified theory of Decentralized SGD, the proofs of \cite{koloskova2020unified} are different from other works focusing on \texttt{FedAvg}. So we reproduce the bound for general convex and non-convex cases based on \cite{karimireddy2020scaffold}. All our results on \texttt{FedAvg} are in Theorem~\ref{thm:fedavg convergence} deferred to Appendix~\ref{sec:proof FedAvg}. As a result, our comparison is persuasive.



\paragraph{The guarantee of \texttt{SSL} is better than \texttt{FedAvg} on heterogeneous data.} Take the strongly convex case as an example. From Table~\ref{table:summary for strongly convex case} (the 4, 5-th rows), (i) it can be seen that the guarantee of \texttt{SSL} has an improvement over \texttt{FedAvg} (marked in red in the 5-th row). Besides, (ii) both \texttt{FedAvg} and \texttt{SSL} are worse than Minibatch SGD (see the discussion in \cite{woodworth2020minibatch}).

We note that the global learning rate is adopted in \cite{karimireddy2020scaffold}, which shows a same-level improvement over \texttt{FedAvg}. However, this technology is still immature in practice \citep{jhunjhunwala2023fedexp} and can be adopted in \texttt{SSL} too. In addition, some variants like \texttt{SCAFFOLD} \citep{karimireddy2020scaffold} show much better than \texttt{FedAvg}, which will be our future work.

\paragraph{Partial participation.} \texttt{SSL} actually works on the more challenging cross-device setting. In this setting, only a small fraction of clients participate in each round. Following the work \citep{li2019convergence, yang2021achieving}, we provide the upper bound of \texttt{FedAvg} with partial participation as follows (Corollary~\ref{cor:without replacement:fedavg strongly convex}, Corollary~\ref{cor:with replacement:fedavg strongly convex} in Appendix~\ref{sec:proof FedAvg}):
\begin{align}
	F(\bar\rvx^R)-F(\rvx^\ast) &= \tilde\gO\left(\frac{\sigma^2}{\mu SKR} + {\color{red}\frac{\zeta_\ast^2}{\mu R}\frac{M-S}{S(M-1)}} + \frac{L\sigma^2}{\mu^2KR^2} + \frac{L\zeta_\ast^2}{\mu^2 R^2} + \mu D^2 \exp\left(-\frac{\mu}{12L}R\right)\right)\nonumber\\
	F(\bar\rvx^R)-F(\rvx^\ast) &= \tilde\gO\left(\frac{\sigma^2}{\mu SKR}+ {\color{red}\frac{\zeta_\ast^2}{\mu SR}} + \frac{L\sigma^2}{\mu^2KR^2} + \frac{L\zeta_\ast^2}{\mu^2 R^2}+ \mu D^2 \exp\left(-\frac{\mu}{12L}R\right)\right),\label{eq:fedavg partial participation}
\end{align}
where $S$ clients are selected randomly and unbiasedly per round without replacement (the 1-st equation) and with replacement (the 2-nd equation). It can be seen that for both sampling schemes, there are additional terms (marked in red), which come from the variance caused by partial participation \citep{yang2021achieving}. For sufficiently large $R$, the upper bounds will \textbf{be dominated by} the first two terms, $\tilde\gO\left(\frac{\sigma^2}{SKR} + {\frac{\zeta_\ast^2}{R}\frac{M-S}{S(M-1)}}\right)$ and $\tilde\gO\left(\frac{\sigma^2}{SKR} + {\frac{\zeta_\ast^2}{SR}}\right)$. For fair comparison, to keep the same computation cost, letting \texttt{SSL} run $S/M \cdot R$ rounds in total, we get the bound of $\tilde\gO\left(\frac{\sigma^2}{SKR}\right)$ for \texttt{SSL}. Note that we only keep the dominated terms and omit other terms and constants $\mu$, $L$ here. Therefore, \textbf{\texttt{SSL} shows better than \texttt{FedAvg} with partial client participation}. At last, this analysis applies to the general convex and non-convex cases.

\section{Feasibility analysis of counterintuitive comparison results}\label{sec:feasibility analysis}
According to Table~\ref{table:summary for strongly convex case}, \texttt{SSL} outperforms \texttt{FedAvg} on heterogeneous data (in the worst case), which contradicts the general-sense understanding via empirical results \citep{gao2020end, gao2021evaluation}. This section analyzes the feasibility of this counterintuitive comparison result, and provides simulation results on quadratic functions. The feasibility analysis also guides experimental settings on extremely heterogeneous data in Section~\ref{sec:exp}.

\paragraph{Feasibility analysis.} The counterintuitive comparison result is caused by the assumption on data heterogeneity (Assumption~\ref{asm:heterogeneity}). Notably, Assumption~\ref{asm:heterogeneity} is developed from the convergence theory of SGD \citep{bottou2018optimization}, which is tight to capture the gradient diversities of the sequential training manner (e.g., \texttt{SSL}). However, it omits the correlations between local objectives, thus might be too pessimistic for \texttt{FedAvg} \citep{wang2022unreasonable}. The negative ``drift'' caused by local updates can be typically canceled each other out in the weighted-average operation of \texttt{FedAvg} in moderately heterogeneous settings. In particular, \cite{wang2022unreasonable} have provided rigorous analyses showing that \texttt{FedAvg} performs much better than theories based on Assumption~\ref{asm:heterogeneity} suggest in moderately heterogeneous settings and real-world FL training tasks (e.g., FEMNIST, \citep{caldas2018leaf}). However, in extremely heterogeneous settings, the ``drift'' can not be canceled out and Assumption~\ref{asm:heterogeneity} turns tight, and \texttt{SSL} can outperform \texttt{FedAvg} in practice.

\begin{figure}[htbp]
	\vspace{-2ex}
	\centering
	\begin{subfigure}{0.32\linewidth}
		\centering
		\includegraphics[width=1\linewidth]{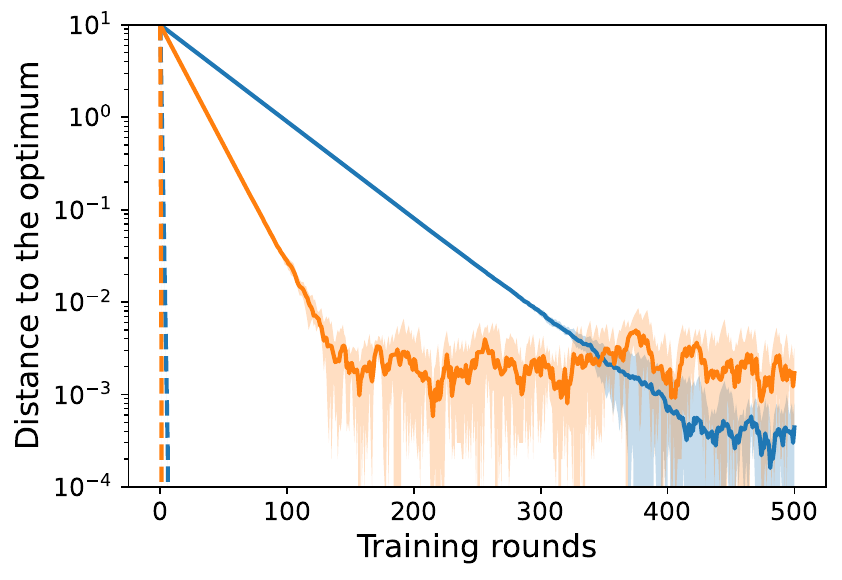}
	\end{subfigure}
	\begin{subfigure}{0.32\linewidth}
		\centering
		\includegraphics[width=1\linewidth]{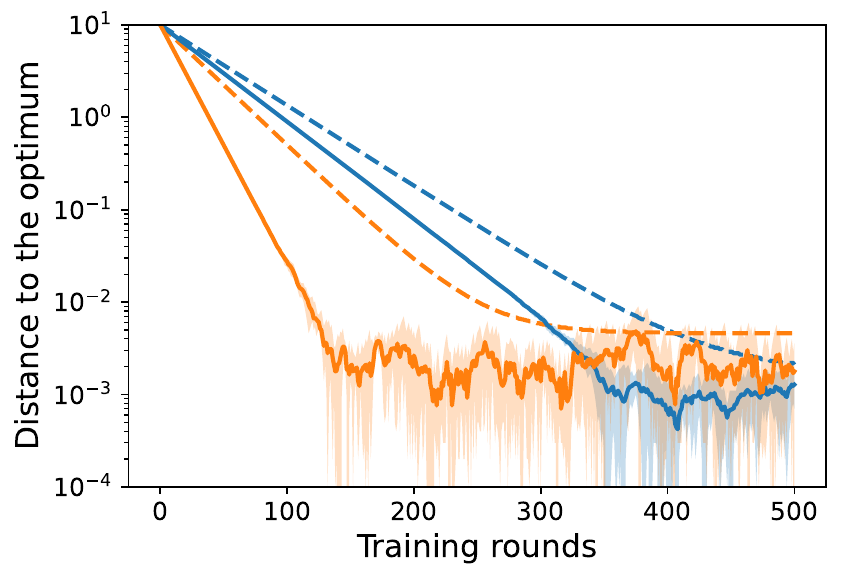}
	\end{subfigure}
	\begin{subfigure}{0.32\linewidth}
		\centering
		\includegraphics[width=1\linewidth]{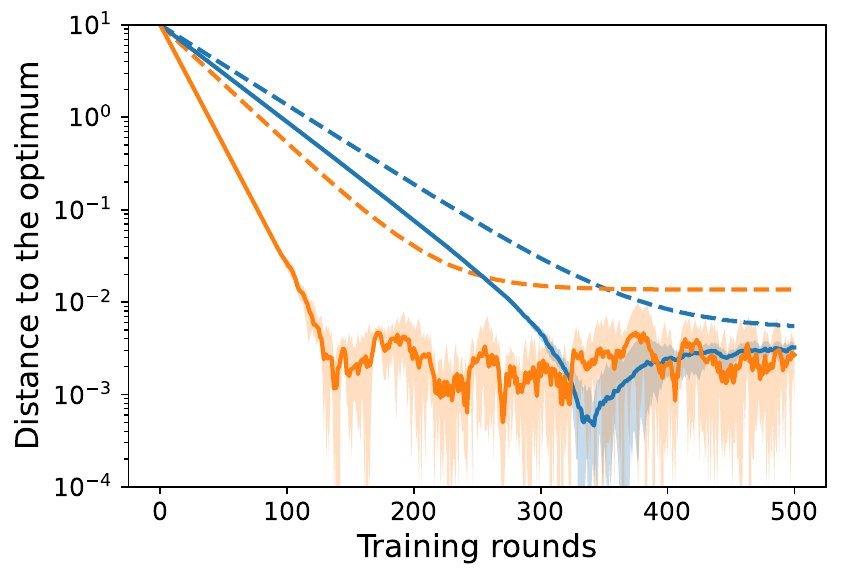}
	\end{subfigure}
	\caption{Simulations on quadratic functions. The left plot, middle plot, right plot are for Group 1, Group 2, Group 3 in Table~\ref{table:simulation settings} respectively. Shaded areas show the min-max values.}
	\label{fig:simulation}
\end{figure}

\paragraph{Simulation validation on quadratic functions.} We adopt the one-dimensional quadratic functions as in \cite{karimireddy2020scaffold} to validate the analysis above. Three groups of experiments with the same global objective $F(x)=\frac{1}{2}x^2$ are considered. The detailed settings are in Table~\ref{table:simulation settings}. Notably, we use the distance between the average of local optima and the global optimum $\norm{\frac{1}{M}\sum_{m=1}^Mx_m^\ast-x^\ast}$ to measure heterogeneity, with a larger value meaning higher heterogeneity. Figure~\ref{fig:simulation} plots the results of \texttt{FedAvg} and \texttt{SSL} with $K=10$. The results validate that \texttt{FedAvg} performs much better than \texttt{SSL} (also better than Theorem~\ref{thm:fedavg convergence} suggests) in moderately heterogeneous settings (Group 1) and  worse than \texttt{SSL} in extremely heterogeneous settings (Groups 2 and 3).
\begin{table}[h!]
	\renewcommand{\arraystretch}{1}
	\centering
	\caption{Settings of simulated experiments. Each group has two local objectives (i.e., $M=2$). The heterogeneity increases from Group 1 to 3.}
	\label{table:simulation settings}
	\normalsize{
		\begin{tabular}{cccc} 
			\toprule
			Settings &Group 1 &Group 2 &Group 3\\
			\midrule
			$F_1(x)$ &$F_1(x) =\frac{1}{2}x^2 + x$ &$F_1(x) =\frac{2}{3}x^2 + x$ &$F_1(x) = x^2 + x$\\[1ex]
			$F_2(x)$ &$F_2(x) =\frac{1}{2}x^2 - x$ &$F_2(x) =\frac{1}{3}x^2 - x$ &$F_2(x) = - x$\\[0.5ex]
			$\norm{\frac{1}{M}\sum_{m=1}^Mx_m^\ast-x^\ast}$  &$0$ &$\frac{3}{8}$ &$+\infty$\\
			\bottomrule
	\end{tabular}}
\end{table}

\section{Experiments}\label{sec:exp}


In this section, we validate our theory empirically in two settings: (1) cross-silo settings with full client participation; (2) cross-device settings with partial client participation \citep{kairouz2021advances}. We adopt the following models and datasets: (i) training LeNet-5 \citep{lecun1998gradient} on the MNIST dataset \citep{lecun1998gradient}; (ii) training LeNet-5 on the FMNIST dataset \citep{xiao2017fashion}; (iii) training VGG-11 \citep{simonyan2014very} on the CIFAR-10 dataset \citep{krizhevsky2009learning}. We partition the training sets artificially with \textit{Extended Dirichlet strategy} and spare the original test sets for computing test accuracy of the global model after each round.

\paragraph{Extended Dirichlet Strategy.} This is to generate arbitrarily heterogeneous data across clients by extending the popular Dirichlet-based data partition strategy \citep{yurochkin2019bayesian,hsu2019measuring}. The difference is to add a step of allocating classes to determine the number of classes per client (denoted by $C$) before allocating samples via Dirichlet distribution (with concentrate parameter $\alpha$). Thus, the extended strategy can be denoted by $\text{Exdir}(C,\alpha)$. Suppose that there are $M$ clients. More details are deferred to Appendix~\ref{sec:apx exps}. The implementation is as follows:
\begin{itemize}[leftmargin=1em]
	\item \textit{Allocating classes}. Select $C$ classes for each client until each class is allocated to at least one client. Then we can obtain the prior distribution $\vq_c \in \R^M$ over $M$ clients for any class $c$.
	\item \textit{Allocating samples}. For any class $c$, we draw $\vp_c\sim \text{Dir}(\alpha \vq_c)$ and then allocate a $\vp_{c,m}$ proportion of the samples of class $c$ to client $m$. For example, $\vq_c = [1, 1, 0, 0, \ldots,]$ means that the samples of class $c$ are only allocated to the first 2 clients.
\end{itemize}
In this work, we use two partitions $\text{Exdir}(1,10.0)$ and $\text{Exdir}(2,10.0)$, where most clients own samples from one class in the former, and from two classes in the latter. So we call $\text{Exdir}(1,10.0)$ as \textit{extremely heterogeneous data} and $\text{Exdir}(2,10.0)$ as \textit{moderately heterogeneous data}. We note that the partition where clients owing samples from one class is not rare \citep{yu2020federated, yang2021achieving, li2022federated}. We note that the partition where clients owning samples from two classes is often seen as pathological \citep{mcmahan2017communication}, so it needs more attention on how much data partitions can affect local objectives.

\begin{figure}[htbp]
	\centering
	\begin{subfigure}{0.45\linewidth}
		\centering
		\includegraphics[width=1\linewidth]{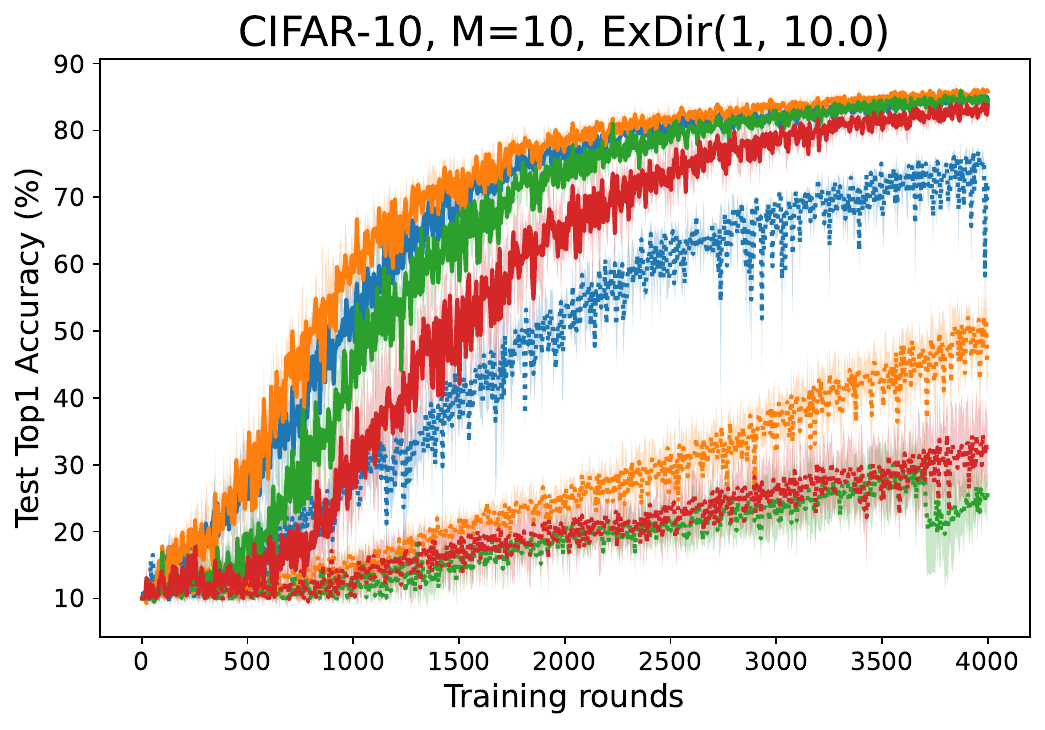}
	\end{subfigure}
	\hspace{2em}
	\begin{subfigure}{0.45\linewidth}
		\centering
		\includegraphics[width=1\linewidth]{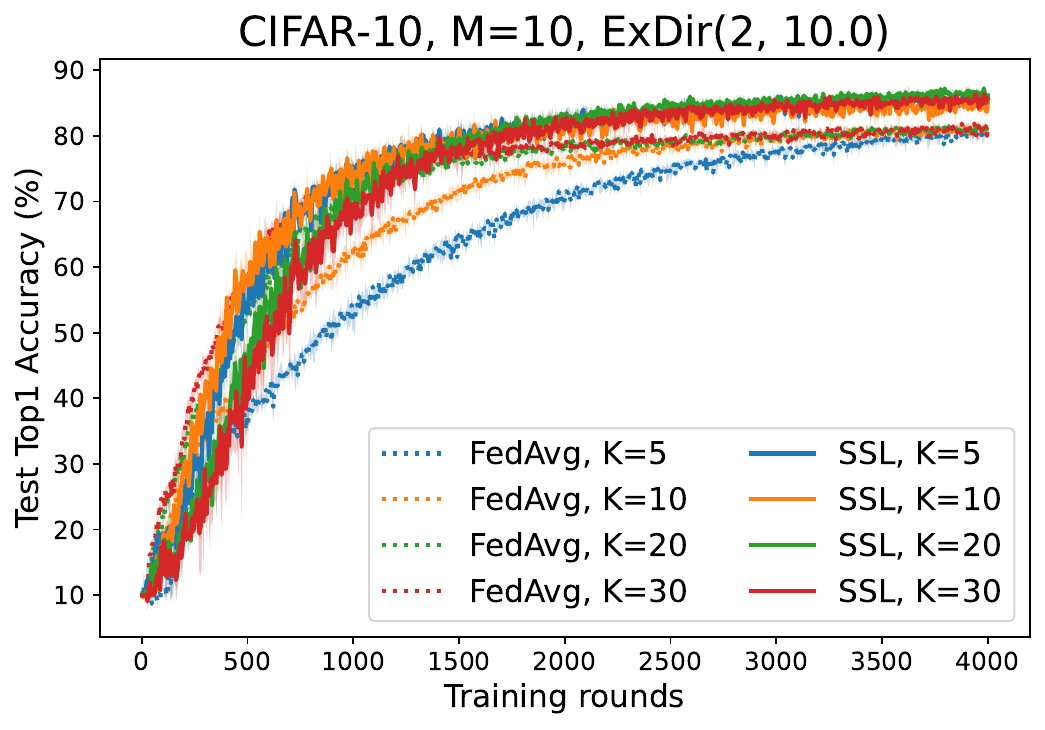}
	\end{subfigure}
	\caption{Test accuracy results with varying number of local steps on CIFAR-10 in cross-silo settings. The shaded areas show the standard deviation.}
	\label{fig:cross-silo settings}
\end{figure}

\paragraph{Cross-silo settings.} The training data is partitioned to 10 clients according to $\text{Exdir}(1,10.0)$ and $\text{Exdir}(2,10.0)$.
Test accuracy results with varying local steps on CIFAR-10 are shown in Figure~\ref{fig:cross-silo settings}. We have the following observations: (i) \textit{data heterogeneity hurts the performance of \texttt{SSL}}. The accuracy curves on $\text{Exdir}(1,10.0)$ exhibit unstable spikes and slower convergence rate than on $\text{Exdir}(2,10.0)$. This phenomenon is more obvious on curves with a large number of local steps. (ii) \textit{Increasing the number of local steps can help the convergence, yet excessive steps can even have negative impacts.} From $K=5$ to $K=10$, the performance of \texttt{SSL} improves, yet from $K=10$ to $K=20$, $30$, it drops consistently. (iii) It can be seen that \texttt{SSL} \textit{outperforms} \texttt{FedAvg} \textit{significantly on extremely heterogeneous data, yet slightly on moderately heterogeneous data}. Interestingly, \texttt{SSL} shows more robust to the choice of $K$ in the extreme case than \texttt{FedAvg} while the opposite is true in the moderate case. From $K=5$ to $K=30$, the performance of \texttt{FedAvg} drops heavily in the left plot yet improves consistently in the right plot. These observations validate our theory.

\paragraph{Cross-device settings.} We highlight the comparison in cross-device settings considering the practical scenario of SL (resource-constrained edge devices). The training data is partitioned to 500 clients according to $\text{Exdir}(1,10.0)$ and $\text{Exdir}(2,10.0)$ for all datasets MNIST, FMNIST and CIFAR-10. For \texttt{FedAvg}, 10 clients are selected in each round. \texttt{SSL} only runs for 1/10 number of the total rounds of \texttt{FedAvg} for fair comparisons. Note that we evaluate the performance of the global model after the training of every 10 clients, which is seen as one ``training rounds'' here. The results with varying local steps are reported in Table~\ref{table:cross-device settings}.
It can be seen that (i) \texttt{SSL} \textit{outperforms} \texttt{FedAvg} \textit{significantly in extremely heterogeneous data on cross-device settings} (i.e., $\text{Exdir}(1,10.0)$). For CIFAR-10, the training of \texttt{FedAvg} almost dies when $K=10$, $20$. Even when $K$ is small, \texttt{FedAvg} struggle to converge. It is not surprising that \texttt{FedAvg} performs awfully in this extremely heterogeneous data. Because there may be only a subset of the total labels selected for training in each round, causing a large sample variance when partial clients are selected (see \eqref{eq:fedavg partial participation}). \texttt{FedAvg} shows better in cross-silo settings with the same local steps (left plot in Figure~\ref{fig:cross-silo settings}) validate it.
In this case, client selection turns vital for \texttt{FedAvg} \citep{fraboni2021clustered}. The experiments in \cite{li2022federated} can validate our results. These match our theory in Section~\ref{sec:SSL vs FedAvg}. Besides, we find that (ii) \texttt{SSL} \textit{has no clear advantage on moderately heterogeneous data}. On FMNIST, \texttt{FedAvg} even shows better than \texttt{SSL} and more local steps seem to enlarge the gap. In addition, the degradation of \texttt{SSL} on MNIST, FMNIST from $\text{Exdir}(1,10.0)$ to $\text{Exdir}(2,10.0)$ is because of the difference on the number of total training rounds. These match our analysis in Section~\ref{sec:feasibility analysis}.

\begin{table}[ht]
	\renewcommand{\arraystretch}{1}
	\centering
	\caption{Test Accuracy results in cross-device settings. 1000 rounds are run for $\text{ExDir}(1, 10.0)$ and 400 rounds for $\text{ExDir}(2, 10.0)$ on MNIST, FMNIST. 4000 rounds are for both two partitions on CIFAR-10. Results are averaged across 3 random seeds and the last 20 rounds. The column 30\% and 75\% show the minimum number of rounds required to reach 30\% and 75\% accuracies.}
	\label{table:cross-device settings}
	\resizebox{\linewidth}{!}{
	\begin{tabular}{llllllllll}
		\toprule
		\multicolumn{2}{c}{\multirow{3}{*}{Setup}} &\multicolumn{4}{c}{$\text{ExDir}(1, 10.0)$} &\multicolumn{4}{c}{$\text{ExDir}(2, 10.0)$} \\
		 & &\multicolumn{2}{c}{FedAvg}  &\multicolumn{2}{c}{SSL} &\multicolumn{2}{c}{FedAvg} &\multicolumn{2}{c}{SSL} \\ \cmidrule(lr){3-6}\cmidrule(lr){7-10}
		 & &ACC (\%) &30\%  &ACC (\%) &30\% &ACC (\%) &75\% &ACC (\%) &75\% \\ \cmidrule(lr){1-2}\cmidrule(lr){3-6}\cmidrule(lr){7-10}
		\multirow{4}{*}{CIFAR-10} 
		 &$K=5$ &$29.06\pm{\color{gray}6.46}$ &3666 &$\textbf{81.34}\pm{\color{gray}1.35}$ &604 &$75.92\pm{\color{gray}1.17}$ &3210 &$\textbf{84.57}\pm{\color{gray}0.90}$ &1004\\
		 &$K=10$ &$14.00\pm{\color{gray}5.66}$ &- &$\textbf{82.20}\pm{\color{gray}1.33}$ &483 &$81.17\pm{\color{gray}0.55}$ &2153 &$\textbf{82.95}\pm{\color{gray}1.70}$ &966\\
		 &$K=20$ &$10.00\pm{\color{gray}0.00}$ &- &$\textbf{80.60}\pm{\color{gray}1.62}$ &673 &$81.26\pm{\color{gray}0.60}$ &1614 &$\textbf{85.24}\pm{\color{gray}0.79}$ &998\\
		 &$K=30$ &- &- &- &- &$80.47\pm{\color{gray}0.69}$ &1618 &$\textbf{84.52}\pm{\color{gray}0.86}$ &966 \\\midrule
		 \multirow{2}{*}{FMNIST}
		 &$K=10$ &$39.90\pm{\color{gray}6.08}$ &492 &$\textbf{80.38}\pm{\color{gray}2.80}$ &19 &$\textbf{81.77}\pm{\color{gray}1.15}$ &127 &$78.25\pm{\color{gray}2.28}$ &162\\
		 &$K=20$ &$38.78\pm{\color{gray}6.51}$ &490 &$\textbf{76.39}\pm{\color{gray}3.15}$ &25 &$\textbf{83.55}\pm{\color{gray}1.13}$ &87 &$76.58\pm{\color{gray}3.18}$ &103\\\midrule
		 \multirow{2}{*}{MNIST}
		 &$K=10$ &$56.43\pm{\color{gray}11.54}$ &274 &$\textbf{98.84}\pm{\color{gray}0.19}$ &16 &$98.15\pm{\color{gray}0.13}$ &25 &$98.57\pm{\color{gray}0.36}$ &10\\
		 &$K=20$ &$51.86\pm{\color{gray}10.66}$ &274 &$\textbf{98.85}\pm{\color{gray}0.12}$ &14 &$98.35\pm{\color{gray}0.14}$ &18 &$98.56\pm{\color{gray}0.33}$ &8\\
		\bottomrule
	\end{tabular}
}
\end{table}

\section{Conclusion}
In this work, we have derived the convergence guarantee of \texttt{SSL} for strongly convex, general convex and non-convex objectives on heterogeneous data. In particular, we have compared \texttt{SSL} against \texttt{FedAvg}, showing that the guarantee of \texttt{SSL} is better than \texttt{FedAvg} on heterogeneous data. Experimental results show that \texttt{SSL} outperforms \texttt{FedAvg} on extremely heterogeneous data, especially in cross-device settings. We believe that this work can bridge the gap between FL and SL, provide deep understanding of both approaches and guide the application deployment in real world.

\bibliographystyle{plainnat}
\bibliography{refs}

\clearpage
\begin{center}
	\LARGE {Appendix}
\end{center}
\appendix
\vskip 3ex\hrule\vskip 2ex
{
	\hypersetup{linktoc=page}
	\parskip=0ex
	\startcontents[sections]
	\printcontents[sections]{l}{1}{\setcounter{tocdepth}{3}}
}
\vskip 3ex\hrule\vskip 5ex
\clearpage

\section{Notations and technical lemmas}

\subsection{Notations}\label{apx:notation}
Table~\ref{table:summary of notations} summarizes the notations appearing in this paper.

\begin{table}[ht]
	\renewcommand{\arraystretch}{1.2}
	\centering
	\caption{Summary of key notations.}
	\label{table:summary of notations}
	\begin{tabular}{cl}
		\toprule
		Symbol &Description\\ \midrule
		$R, r$ &number, index of training rounds \\
		$M, m$ &number, index of clients \\
		$K, k$ &number, index of local update steps \\
		$\pi$ &$\{\pi_1, \pi_2, \ldots, \pi_M\}$ is a permutation of $\{1,2,\ldots,M\}$\\ 
		$S$ &number of clients selected for training per round with partial client participation \\
		$\eta$ &learning rate (or stepsize) \\
		$\tilde\eta$ &effective learning rate ($\tilde\eta\coloneqq MK\eta$ in \texttt{SSL} and $\tilde\eta\coloneqq K\eta$ in \texttt{FedAvg})\\
		$\mu$ &$\mu$-strong convexity constant\\
		$L$ &$L$-smoothness constant (Asm.~\ref{asm:L-smoothness})\\
		$\sigma$ &upper bound on variance of stochastic gradients (Asm.~\ref{asm:stochasticity} on stochasticity)\\
		$B, \zeta$ &constants in Asm.~\ref{asm:heterogeneity} to bound heterogeneity everywhere\\
		$\zeta_\ast$ &constants in Asm.~\ref{asm:heterogeneity} to bound heterogeneity at the optimum \\
		$F/F_m$ &global objective/local objective of client $m$\\
		$\rvx^{r}$ &global model parameters in the $r$-th round \\
		$\rvx_m^{(r, k)}$ &\begin{tabular}[l]{@{}l@{}}local model parameters \\ \quad after the $k$-th local update with client $\pi_m$ in the $r$-th round \end{tabular} \\
		$\rvg_{\pi_m}^{(r, k)}$ &\begin{tabular}[l]{@{}l@{}}$\rvg_{\pi_m}^{(r,k)} \coloneqq \nabla f_{\pi_m}(\rvx_{m}^{(r,k)};\xi_m)$ denotes the stochastic gradients of $F_{\pi_m}$ \\ \quad regarding model parameters $\rvx_m^{(r, k)}$, on the sample $\xi_m$ \end{tabular} \\
		$\text{ExDir}(C,\alpha)$ &\begin{tabular}[l]{@{}l@{}}Extended Dirichlet strategy with parameters $C$ and $\alpha$. \\
		\quad Decreasing the values of $C$, $\alpha$ results in higher data heterogeneity. (Sec.~\ref{subsec:extented dirichlet}).\end{tabular}\\
		\bottomrule
	\end{tabular}
\end{table}

\subsection{Basic identities and inequalities}
These identities and inequalities are mostly from \cite{mishchenko2020random, karimireddy2020scaffold, zhou2018fenchel, garrigos2023handbook}.

For any random variable $X$, letting the variance can be decomposed as
\begin{align}
	\E{\norm{X-\E{X}}^2} = \E{\norm{X}^2} - \norm{\E{X}}^2 \label{eq:variance decomposition}
\end{align}
For the discrete random variables, it holds that 
\begin{equation}
	\frac{1}{m} \sum_{i=1}^{m} \sqn{\vx_i - \bar{\vx}} = \frac{1}{m} \sum_{i=1}^{m} \sqn{\vx_{i}} - \sqn{ \frac{1}{m} \sum_{i=1}^{m} \vx_i },\label{eq:variance decomposition discrete}
\end{equation}
for given vectors $\vx_{1}, \ldots, \vx_{m} \in \R^d$ and their average $\bar{\vx}$.

\paragraph{Jensen's inequality.} For any convex function $h$ and any vectors $\vx_1,\dotsc, \vx_m$ we have
\begin{align}
	h\left(\frac{1}{m}\sum_{i=1}^m \vx_i\right) \leq \frac{1}{m}\sum_{i=1}^m f(\vx_i). \label{eq:jensen's inquality}
\end{align}
As a special case with $f(x)=\norm{x}^2$, we obtain
\begin{align}
	\sqn{\frac{1}{m}\sum_{i=1}^m \vx_i} \leq \frac{1}{m}\sum_{i=1}^m \norm{\vx_i}^2. \label{eq:jensen norm}
\end{align}

\paragraph{Smoothness and (general) convexity, strong convexity.} There are some useful inequalities with respect to $L$-smoothness (Assumption~\ref{asm:L-smoothness}), convexity and $\mu$-strong convexity. Their proofs can be found in \cite{zhou2018fenchel, garrigos2023handbook}.

\textit{Bregman Divergence} associated with function $h$ and arbitrary $\vx$, $\vy$ is denoted as
\begin{align*}
	D_h(\vx, \vy) \coloneqq h(\vx) - h(\vy) - \inp{\nabla h(\vy)}{\vx-\vy}
\end{align*}
When the function $h$ is convex, the divergence is strictly non-negative. A more formal definition can be found in \cite{orabona2019modern}. One corollary (Chen and Teboulle, 1993) called \textit{three-point-identity} is, 
\begin{align*}
D_h(\vz, \vx) + D_h(\vx,\vy) - D_h(\vz,\vy) = \inp{\nabla h(\vy) - \nabla h(\vx)}{\vz-\vx}
\end{align*}
where $\vx, \vy, \vz$ is three points in the domain.

Let $h$ be $L$-smooth. With the definition of Bregman divergence, a useful consequence of $L$-smoothness is
\begin{equation}
	D_h(\vx, \vy) = h(\vx) - h(\vy) - \inp{\nabla h(\vy)}{\vx-\vy} \leq \frac{L}{2}\norm{\vx-\vy}^2 \label{eq:L-smoothness:1}
\end{equation}

Further, If $h$ is $L$-smooth and lower bounded by $h_\ast$, then
\begin{equation}
	\sqn{\nabla h(\vx)} \leq 2 L \left(h(\vx) - h_\ast\right).\label{eq:L-smoothness grad bound}
\end{equation}

Then if $h$ is $L$-smooth and convex (The definition of convexity can be found in \cite{boyd2004convex}), it holds that
\begin{align}
	D_h(\vx, \vy)\ge \frac{1}{2L}\norm{\nabla h(\vx) - \nabla h(\vy)}^2. \label{eq:bregman lower smooth}
\end{align}

The function $h: \R^d\to \R$ is $\mu$-\textbf{strongly convex} if and only if there exists a convex function $g: \R^d\to \R$ such that $h(\vx) = g(\vx) + \frac{\mu}{2}\norm{\vx}^2$.

If $h$ is $\mu$-strongly convexity, it holds that
\begin{align}
	\frac{\mu}{2}\norm{\vx-\vy}^2 \leq D_h(\vx, \vy)
\end{align}

\subsection{Technical lemmas}

\begin{lemma}[\cite{wang2020tackling}]\label{lem:martingale difference property}
	Suppose $\{A_k\}_{k=1}^T$ is a sequence of random matrices and $\E[A_k|A_{k-1},A_{k-2},\dots,A_1] = \bm{0},\forall k$. Then,
	\begin{align*}
		\E\left[\fronorm{\sum_{k=1}^T A_k}^2\right] = \sum_{k=1}^T \E\left[\fronorm{A_k}^2\right].\label{eq:martingale difference property}
	\end{align*}
\end{lemma}
\begin{proof}
	This is the Lemma 2 of \cite{wang2020tackling}. Similar conclusions can be also found in \cite{stich19errorfeedback, karimireddy2020scaffold}.
\end{proof}

\begin{lemma}[\cite{karimireddy2020scaffold}]
	\label{lem:perturbed strong convexity}
	The following holds for any $L$-smooth and $\mu$-strongly convex function $h$, and any $\vx, \vy, \vz$ in the domain of $h$:
	\begin{align}
		\left\langle \nabla h(\vx), \vz-\vy\right\rangle \geq h(\vz) - h(\rvy) +\frac{\mu}{4}\norm{\vy - \vz}^2  - L\norm{\vz - \vx}^2.
	\end{align}
\end{lemma}
\begin{proof}
	This is Lemma~5 (perturbed strong convexity) in \cite{karimireddy2020scaffold}.
\end{proof}

\begin{lemma}[Simple Random Sampling]\label{lem:simple random sampling}
	Let $X_1, X_2,\ldots, X_M\in \R^d$ be fixed units. The population mean and population variance are give as
	\begin{align*}
		\textstyle\overline X \coloneqq \frac{1}{M}\sum_{i=1}^M X_i 
		&&
		\textstyle\sigma^2 \coloneqq \frac{1}{M}\sum_{i=1}^M \norm{X_i-\overline X}^2.
	\end{align*}
	Sample $S\in [M]= \{1,2,\ldots,M\}$ units $\rvx_{\pi_1}, \rvx_{\pi_2},\ldots \rvx_{\pi_S}$ from the population. There are two possible ways of simple random sampling, well known as ``sampling with replacement'' and ``sampling without replacement''. For these two ways, the expectation and variance of the sample mean $\overline \rvx_\pi \coloneqq \frac{1}{S}\sum_{s=1}^S \rvx_{\pi_s}$ satisfies
	\begin{itemize}[leftmargin=1em]
		\item Sampling without replacement (SWOR).
		\begin{align}
			\E[\overline \rvx_\pi] = \overline X
			&&
			\E\left[\norm{\overline \rvx_{\pi} - \overline X}^2\right] = \frac{M-S}{S(M-1)}\sigma^2\label{eq:lem:random sampling without replacement}
		\end{align}
		\item Sampling with replacement (SWR).
		\begin{align}
			\E[\overline \rvx_\pi] = \overline X
			&&
			\E\left[\norm{\overline \rvx_{\pi} - \overline X}^2\right] = \frac{\sigma^2}{S}\label{eq:lem:random sampling with replacement}
		\end{align}
	\end{itemize}
	
\end{lemma}

\begin{proof}
	This lemma is mainly based on the Lemma 1 in \cite{mishchenko2020random} and Appendix G Extension: Incorporating Client Sampling in \cite{wang2020tackling}.
	\paragraph{Expectation of the sample mean}	
	\begin{itemize}[leftmargin=1em]
		\item (SWOR) Note $\Pr(\rvx_{\pi_s} = X_i) = \frac{1}{M}$ since every unit has the same probability to be selected.
		\begin{align*}
			\E[\overline \rvx_\pi] = \E \left[\frac{1}{S} \sum_{s=1}^S \rvx_s\right] = \frac{1}{S}\sum_{s=1}^S\E\left[\rvx_s\right] = \frac{1}{S}\sum_{s=1}^S\left[\sum_{i=1}^M \Pr(\rvx_s= X_i) X_i\right] = \frac{1}{S}\sum_{s=1}^S\left[\frac{1}{M}\sum_{i=1}^M X_i\right] = \overline X
		\end{align*}
		\item (SWR) $\E[\overline \rvx_\pi] = \overline X$, same as SRSWOR, since $\Pr(\rvx_{\pi_s} = X_i) = \frac{1}{M}$.
	\end{itemize}
	\paragraph{Variance of the sample mean}
	\begin{align*}
		\E\norm{\overline \rvx_\pi - \overline X}^2 &= \E\norm{\frac{1}{S}\sum_{s=1}^S\rvx_{\pi_s} - \overline X}^2 = \E\norm{\frac{1}{S}\sum_{s=1}^S(\rvx_{\pi_s} - \overline X)}^2 = \frac{1}{S^2}\E\norm{\sum_{s=1}^S(\rvx_{\pi_s} - \overline X)}^2\\
		&=\frac{1}{S^2}\sum_{s=1}^S\Var(\rvx_{\pi_s}) + \frac{1}{S^2}\sum_{s=1}^S\sum_{t=1, t\neq s}^S \Cov (\rvx_{\pi_s}, \rvx_{\pi_t})
	\end{align*}
	Since $\Var(\rvx_{\pi_s}) = \E\norm{\rvx_{\pi_s} - \E [\rvx_{\pi_s}]}^2 = \E\norm{\rvx_{\pi_s} - \overline X}^2$, we have
	\begin{align*}
		\frac{1}{S^2}\sum_{s=1}^S\Var(\rvx_{\pi_s}) = \frac{1}{S^2}\sum_{s=1}^S\left[\sum_{i=1}^M \Pr(\rvx_{\pi_s} = X_i) (X_i - \overline X)^2\right] = \frac{1}{S^2}\sum_{s=1}^S\left[\frac{1}{M}\sum_{i=1}^M (X_i - \overline X)^2\right] = \frac{\sigma^2}{S}
	\end{align*}
	For the covariance term, we need to consider two cases:
	\begin{itemize}[leftmargin=1em]
		\item (SWOR) For $s\neq t$, we have
		\begin{align*}
			\Cov(\rvx_{\pi_s}, \rvx_{\pi_t}) = \E \left\langle \rvx_{\pi_s} - \overline X, \rvx_{\pi_t} - \overline X\right\rangle
			=\sum_{i=1}^M\sum_{j=1, j\neq i}^M\left[\Pr(\rvx_{\pi_s}=X_i, \rvx_{\pi_t} = X_j)\left\langle X_i - \overline X, X_j - \overline X\right\rangle\right],
		\end{align*}
		Since there are $M(M-1)$ possible combinations of $(\rvx_{\pi_s},\rvx_{\pi_t})$, and each has the same probability, we get $\Pr(\rvx_{\pi_s}=X_i, \rvx_{\pi_t} = X_j) = \frac{1}{M(M-1)}$. As a consequence, we have
		\begin{align}
			\Cov(\rvx_{\pi_s}, \rvx_{\pi_t})&=\frac{1}{M(M-1)}\sum_{i=1}^M\sum_{j=1, j\neq i}^M\left[\left\langle X_i - \overline X, X_j - \overline X\right\rangle\right] \nonumber\\
			&= \frac{1}{M(M-1)}\sum_{i=1}^M\sum_{j=1}^M\left[\left\langle X_i - \overline X, X_j - \overline X\right\rangle\right] - \frac{1}{M(M-1)}\sum_{i=1}^M \norm{X_i-\overline X}^2 \nonumber\\
			&= \frac{1}{M(M-1)}\sum_{i=1}^M\left[\left\langle X_i - \overline X, \underbrace{\sum_{j=1}^M (X_j - \overline X)}_{=0}\right\rangle\right] - \frac{1}{M(M-1)}\sum_{i=1}^M \norm{X_i-\overline X}^2 \nonumber\\
			&= -\frac{\sigma^2}{M-1}\label{eq:proof:lem:random sampling without replacement:covariance}
		\end{align}
		Thus we have
		\begin{align*}
			\E\norm{\overline \rvx_\pi - \overline X}^2 = \frac{\sigma^2}{S} - \frac{S(S-1)}{S^2}\left(\frac{\sigma^2}{M-1}\right)=\frac{(M-S)\sigma^2}{S(M-1)}
		\end{align*}
		\item (SWR) For $s\neq t$, we have $\Cov(\rvx_{\pi_s}, \rvx_{\pi_t}) = 0$ since $x_s$ and $x_t$ are independent.
		Thus we get
		\begin{align*}
			\E\norm{\overline \rvx_\pi - \overline X}^2 = \frac{\sigma^2}{S}
		\end{align*}
	\end{itemize}
	Now, we complete the proof. When $M$ is infinite (or large enough), we get $\frac{(\E\norm{\overline \rvx_\pi - \overline X}^2)_\textrm{SWOR}}{(\E\norm{\overline \rvx_\pi - \overline X}^2)_\textrm{SWR}} \approx 1-\frac{S}{M}$. This constant has appeared in Lemma 7 (one round progress) in \cite{karimireddy2020scaffold} and Section 7 (Using a Subset of Machines in Each Round) in \cite{woodworth2020minibatch}.
\end{proof}


\section{Proofs of Theorem~\ref{thm:SSL convergence}}\label{sec:proof SSL}
In this section, we provide the proof of Theorem~\ref{thm:SSL convergence} for the strongly convex, general convex and non-convex cases in \ref{subsec:strongly convex ssl}, \ref{subsec:general convex ssl} and \ref{subsec:non-convex ssl}, respectively.

\subsection{Additional technical lemmas}
\begin{lemma}\label{lem:technical lemma SSL}
	Under the same conditions of Lemma~\ref{lem:simple random sampling}, use the way ``sampling without replacement'' and let $\mathcal{K}(i) \coloneqq \begin{cases}
		K-1,\; 1 \leq i \leq m-1\\
		k-1,\; i = m
	\end{cases}$ ($k$, $m$ are not constants). It holds that
	\begin{align}
		\sum_{m=1}^M\sum_{k=0}^{K-1}\E\norm{\sum_{i=1}^{m}\sum_{j=0}^{\gK(i)}\left(\rvx_{\pi_i} - \overline X\right)}^2 \leq \frac{1}{3}M^2K^3\sigma^2
	\end{align}
\end{lemma}
\begin{proof}
	Let us focus on the term in the following:
	\begin{align}
		&\E\norm{\sum_{i=1}^{m}\sum_{j=0}^{\gK(i)}\left(\rvx_{\pi_i} - \overline X\right)}^2 = \E\norm{K\sum_{i=1}^{m-1}\left(\rvx_{\pi_i} - \overline X\right) + k\left(\rvx_{\pi_m} - \overline X\right)}^2 \nonumber\\
		&= \E\norm{K\sum_{i=1}^{m-1}\left(\rvx_{\pi_i} - \overline X\right)}^2 + \E\norm{ k\left(\rvx_{\pi_m} - \overline X\right)}^2 + 2\E\inp{K\sum_{i=1}^{m-1}\left(\rvx_{\pi_i} - \overline X\right)}{k\left(\rvx_{\pi_m} - \overline X\right)}\label{eq:proof:lem:technical lemma SL:1}
	\end{align}
	For the first term on the right hand side in \eqref{eq:proof:lem:technical lemma SL:1}, using \eqref{eq:lem:random sampling without replacement}, we have
	\begin{align*}
		\E\norm{K\sum_{i=1}^{m-1}\left(\rvx_{\pi_i} - \overline X\right)}^2 = \frac{(m-1)(M-(m-1))}{M-1}K^2\sigma^2
	\end{align*}
	For the second term on the right hand side in \eqref{eq:proof:lem:technical lemma SL:1}, we have
	\begin{align*}
		\E\norm{ k\left(\rvx_{\pi_m} - \overline X\right)}^2 = k^2 \E\norm{ \rvx_{\pi_m} - \overline X}^2 = k^2\sigma^2
	\end{align*}
	For the third term on the right hand side in \eqref{eq:proof:lem:technical lemma SL:1}, we have 
	\begin{align*}
		2\E\inp{K\sum_{i=1}^{m-1}\left(\rvx_{\pi_i} - \overline X\right)}{k\left(\rvx_{\pi_m} - \overline X\right)} &= 2Kk\sum_{i=1}^{m-1} \E\inp{\rvx_{\pi_i} - \overline X}{\rvx_{\pi_m} - \overline X}\\
		&= 2Kk(m-1)\left(-\frac{\sigma^2}{M-1}\right) = \frac{2(m-1)}{M-1}Kk\sigma^2,
	\end{align*}
	where we use \eqref{eq:proof:lem:random sampling without replacement:covariance} in the last equality, since $i \in \{1,2,\ldots,m-1\} \neq m$. Combining these three preceding equations, we get
	\begin{align}
		\E\norm{\sum_{i=1}^{m}\sum_{j=0}^{\gK(i)}\left(\rvx_{\pi_i} - \overline X\right)}^2 = \frac{(m-1)(M-(m-1))}{M-1}K^2\sigma^2 + k^2\sigma^2 - \frac{2(m-1)}{M-1}Kk\sigma^2
	\end{align}
	Then summing over $m$ and $k$, we can get
	\begin{align}
		\sum_{m=1}^M\sum_{k=0}^{K-1}\E\norm{\sum_{i=1}^{m}\sum_{j=0}^{\gK(i)}\left(\rvx_{\pi_i} - \overline X\right)}^2
		&= K^3\sigma^2 \left(\frac{M}{M-1}\sum_{m=1}^M(m-1)-\frac{1}{M-1}\sum_{m=1}^M(m-1)^2\right) \nonumber\\
		&\quad+ M\sigma^2\sum_{k=0}^{K-1}k^2 +
		\frac{2K\sigma^2}{M-1}\sum_{m=1}^M (m-1)\sum_{k=0}^{K-1}k
	\end{align}
	Then applying the fact that $\sum_{k=1}^{K-1}k = \frac{(K-1)K}{2}$ and $\sum_{k=1}^{K-1}k^2 = \frac{(K-1)K(2K-1)}{6}$, we can simplify the preceding equation as
	\begin{align}
		\sum_{m=1}^M\sum_{k=0}^{K-1}\E\norm{\sum_{i=1}^{m}\sum_{j=0}^{\gK(i)}\left(\rvx_{\pi_i} - \overline X\right)}^2 = \frac{1}{6}(M^2K^3+MK) \sigma^2 \leq \frac{1}{3}M^2K^3\sigma^2,
	\end{align}
	which is the claim of this lemma.
\end{proof}

\subsection{Strongly convex case}\label{subsec:strongly convex ssl}

\subsubsection{Finding the recursion}
\begin{lemma}
	Let Assumptions~\ref{asm:L-smoothness}, \ref{asm:stochasticity}, \eqref{eq:asm:heterogeneity:optimum} of \ref{asm:heterogeneity} hold and assume that all the local objectives are $\mu$-strongly convex. If the learning rate satisfies $\eta \leq \frac{1}{6LMK}$, then it holds that
	\begin{align}
		\E\norm{\rvx^{r+1}-\rvx^\ast}^2 &\leq \left(1-\tfrac{\mu MK\eta}{2}\right)\norm{\rvx^{r}-\rvx^*}^2 - MK\eta(2-6LMK\eta)\E D_F(\rvx^r, \rvx^\ast) + 3MK\eta^2\sigma^2 \nonumber\\
		&\quad+ L\eta(2+3LMK\eta)\sum_{m=1}^M\sum_{k=0}^{K-1}\E\norm{\rvx_m^{(r,k)} - \rvx^r}^2\label{eq:lem:SSL strongly convex:the per-round recursion}
	\end{align}
\end{lemma}
\begin{proof}
	Without otherwise stated, \textbf{the expectation is conditioned on} $\rvx^r$. The update rule of SL in round $r$ can be written as \(\displaystyle \rvx^{r+1} = \rvx^{r} -\eta\sum_{m=1}^M\sum_{k=0}^{K-1}\rvg_{\pi_m}^{(r,k)}\),
	where $\rvg_{\pi_m}^{(r,k)} = \nabla f_{\pi_m}(\rvx_{m}^{(r,k)};\xi_{m}^{(r,k)})$.
	As a consequence, we can get,
	\begin{align}
		\E\norm{\rvx^{r+1}-\rvx^\ast}^2 = \norm{\rvx^{r}-\rvx^*}^2 - 2\eta\sum_{m=1}^M\sum_{k=0}^{K-1}\inp{\nabla F_{\pi_m}(\rvx_m^{(r,k)})}{\rvx^r-\rvx^\ast} + \eta^2\E\norm{\sum_{m=1}^M\sum_{k=0}^{K-1}\rvg_{\pi_m}^{(r,k)}}^2\label{eq:proof:lem:SSL strongly convex:the per-round recursion:1}
	\end{align}
	
	Using Lemma~\ref{lem:perturbed strong convexity} with $\vx=\rvx_m^{(r,k)}$, $\vy=\rvx^\ast$, $\vz=\rvx^r$ and $h = F_{\pi_m}$, we have
	\begin{align}
		&-2\eta\sum_{m=1}^M\sum_{k=0}^{K-1}\E\inp{\nabla F_{\pi_m}(\rvx_m^{(r,k)})}{\rvx^r-\rvx^\ast} \nonumber\\
		&\leq -2\eta\sum_{m=1}^M\sum_{k=0}^{K-1} \E\left[F_{\pi_m}(\rvx^{r})-F_{\pi_m}(\rvx^\ast) + \frac{\mu}{4}\norm{\rvx^r-\rvx^\ast}^2 - L\norm{\rvx_m^{(r,k)}-\rvx^{r}}^2\right] \nonumber\\
		&\leq -2MK\eta \left(F(\rvx^{r})- F(\rvx^\ast)\right) - \frac{\mu MK\eta}{2}\norm{\rvx^{r}-\rvx^\ast}^2 + 2L\eta\sum_{m=1}^M\sum_{k=0}^{K-1}\E\norm{\rvx_m^{(r,k)}-\rvx^{r}}^2\label{eq:proof:lem:SSL strongly convex:the per-round recursion:2}
	\end{align}
	
	For the third term on the right hand side in \eqref{eq:proof:lem:SSL strongly convex:the per-round recursion:1}, using Jensen's inequality, we have:
	\begin{align}
		\E\norm{\sum_{m=1}^M\sum_{k=0}^{K-1} \rvg_{\pi_m}^{(r,k)}}^2&\leq 3\E\norm{\sum_{m=1}^M\sum_{k=0}^{K-1} \rvg_{\pi_m}^{(r,k)} - \sum_{m=1}^M\sum_{k=0}^{K-1} \nabla F_{\pi_m}(\rvx_m^{(r,k)})}^2 \nonumber\\
		&+ 3\E\norm{\sum_{m=1}^M\sum_{k=0}^{K-1} \nabla F_{\pi_m}(\rvx_m^{(r,k)}) - \sum_{m=1}^M\sum_{k=0}^{K-1} \nabla F_{\pi_m}(\rvx^r)}^2 + 3\E\norm{\sum_{m=1}^M\sum_{k=0}^{K-1} \nabla F_{\pi_m}(\rvx^r)}^2\label{eq:proof:lem:SSL strongly convex:the per-round recursion:3}
	\end{align}

	For the first term on the right hand side in \eqref{eq:proof:lem:SSL strongly convex:the per-round recursion:3}, using Lemma~\ref{lem:martingale difference property}, we have
	\begin{align}
		\E\norm{\sum_{m=1}^M\sum_{k=0}^{K-1} \rvg_{\pi_m}^{(r,k)} - \sum_{m=1}^M\sum_{k=0}^{K-1} \nabla F_{\pi_m}(\rvx_m^{(r,k)})}^2 \leq \sum_{m=1}^M\sum_{k=0}^{K-1}\E\norm{\rvg_{\pi_m}^{(r,k)}-\nabla F_{\pi_m}(\rvx_m^{(r,k)})}^2 \leq MK\sigma^2
	\end{align}
	For the second term on the right hand side in \eqref{eq:proof:lem:SSL strongly convex:the per-round recursion:3}, we have
	\begin{align}
		\E\norm{\sum_{m=1}^M\sum_{k=0}^{K-1} \nabla F_{\pi_m}(\rvx_m^{(r,k)}) - \sum_{m=1}^M\sum_{k=0}^{K-1} \nabla F_{\pi_m}(\rvx^r)}^2
		&\leq MK\sum_{m=1}^M\sum_{k=0}^{K-1}\E\norm{\nabla F_{\pi_m}(\rvx_m^{(r,k)}) - \nabla F_{\pi_m}(\rvx^r)}^2 \nonumber\\
		&\overset{\text{Asm.}~\ref{asm:L-smoothness}}{\leq} L^2MK\sum_{m=1}^M\sum_{k=0}^{K-1}\E\norm{\rvx_m^{(r,k)} - \rvx^r}^2
	\end{align}
	For the third term on the right hand side in \eqref{eq:proof:lem:SSL strongly convex:the per-round recursion:3}, we have
	\begin{align}
		\E\norm{\sum_{m=1}^M\sum_{k=0}^{K-1} \nabla F_{\pi_m}(\rvx^r)}^2
		\leq M^2K^2\E\norm{\nabla F(\rvx^r)}^2 \overset{\eqref{eq:L-smoothness grad bound}}{\leq} 2LM^2K^2D_{F}(\rvx^{r}, \rvx^\ast)
	\end{align}
	As a result, the third term on the right hand side in \eqref{eq:proof:lem:SSL strongly convex:the per-round recursion:1} can be bounded as:
	\begin{align}
		\E\norm{\sum_{m=1}^M\sum_{k=0}^{K-1}\rvg_{\pi_m}^{(r,k)}}^2 &\leq 3L^2MK\sum_{m=1}^M\sum_{k=0}^{K-1}\E\norm{\rvx_m^{(r,k)} - \rvx^r}^2 + 6LM^2K^2D_{F}(\rvx^{r}, \rvx^\ast) + 3MK\sigma^2\label{eq:proof:lem:SSL strongly convex:the per-round recursion:4}
	\end{align}
	
	Then substituting \eqref{eq:proof:lem:SSL strongly convex:the per-round recursion:2} and \eqref{eq:proof:lem:SSL strongly convex:the per-round recursion:4} into \eqref{eq:proof:lem:SSL strongly convex:the per-round recursion:1}, we have
	\begin{align}
		\E\norm{\rvx^{r+1}-\rvx^\ast}^2 &\leq \left(1-\tfrac{\mu MK\eta}{2}\right)\norm{\rvx^{r}-\rvx^*}^2 - MK\eta(2-6LMK\eta)D_F(\rvx^r, \rvx^\ast) + 3MK\eta^2\sigma^2 \nonumber\\
		&\quad+ L\eta(2+3LMK\eta)\sum_{m=1}^M\sum_{k=0}^{K-1}\E\norm{\rvx_m^{(r,k)} - \rvx^r}^2
	\end{align}
	The claim of this lemma follows after taking unconditional expectation.
\end{proof}

\subsubsection{Bounding the client drift with \eqref{eq:asm:heterogeneity:optimum}}
Similar to the client drift in FL \citep{karimireddy2020scaffold}, we define the client drift in \texttt{SSL} as \(\displaystyle E_r \coloneqq \sum_{m=1}^M\sum_{k=0}^{K-1} \E\norm{\rvx_m^{(r,k)} - \rvx^r}^2 \).

\begin{lemma}\label{lem:SSL strongly convex:client drift}
	For any learning rate satisfying $\eta \leq \frac{1}{6LMK}$, the client drift caused by local updates is bounded, as given by:
	\begin{align}
		\E\left[E_r\right]
		&\leq \frac{9}{2}LM^3K^3\eta^2\E\left[D_{F}(\rvx^{r}, \rvx^\ast)\right] + \frac{9}{4}M^2K^2\eta^2\sigma^2 + \frac{3}{2}M^2K^3\eta^2\zeta_\ast^2 \label{eq:lem:SSL strongly convex:client drift}
	\end{align}
\end{lemma}
\begin{proof}
	Without otherwise stated, \textbf{the expectation is conditioned on} $\rvx^r$. Beginning with \(\E\norm{\rvx_m^{(r,k)} - \rvx^r}^2\). 
	Considering 
	\begin{align}
		\rvx_m^{(r,k)}- \rvx^{r} = \rvx_m^{(r,k)}-\rvx_m^{(r,0)} + \rvx_m^{(r,0)}-\rvx_{m-1}^{(r,0)}+\cdots+\rvx_2^{(r,0)}-\rvx_1^{(r,0)} = - \eta\sum_{i=1}^{m} \sum_{j=0}^{\gK(i)} \rvg_{\pi_i}^{(r,j)},
	\end{align}
	with $\mathcal{K}(i) \coloneqq \begin{cases}
		K-1,\; 1 \leq i \leq m-1\\
		k-1,\; i = m
	\end{cases},$
	we have
	\begin{align}
		\E\norm{\rvx_m^{(r,k)} - \rvx^r}^2 &= \eta^2\E \norm{\sum_{i=1}^{m}\sum_{j=0}^{\gK(i)} \rvg_{\pi_i}^{(r,j)}}^2
	\end{align}
	Then using the Jensen's inequality to the preceding equation, we have
	\begin{align}
		&\E\norm{\rvx_m^{(r,k)} - \rvx^r}^2 \nonumber\\
		&\leq 4\eta^2\E \norm{\sum_{i=1}^{m}\sum_{j=0}^{\gK(i)} \rvg_{\pi_i}^{(r,j)} - \sum_{i=1}^{m}\sum_{j=0}^{\gK(i)}\nabla F_{\pi_i} (\rvx_i^{(r,j)})}^2 + 4\eta^2\E\norm{\sum_{i=1}^{m}\sum_{j=0}^{\gK(i)}\nabla F_{\pi_i} (\rvx_i^{(r,j)}) - \sum_{i=1}^{m}\sum_{j=0}^{\gK(i)}\nabla F_{\pi_i} (\rvx^{r})}^2 \nonumber\\
		&\quad+ 4\eta^2\E \norm{\sum_{i=1}^{m}\sum_{j=0}^{\gK(i)}\nabla F_{\pi_i} (\rvx^{r}) - \sum_{i=1}^{m}\sum_{j=0}^{\gK(i)}\nabla F_{\pi_i} (\rvx^{\ast})}^2 + 4\eta^2\E\norm{\sum_{i=1}^{m}\sum_{j=0}^{\gK(i)}\nabla F_{\pi_i} (\rvx^{\ast})}^2\label{eq:proof:lem:SSL strongly convex:client drift:1}
	\end{align}
	Applying Lemma~\ref{lem:martingale difference property} to the first term and Jensen’s inequality to the last three terms on the right hand side in \eqref{eq:proof:lem:SSL strongly convex:client drift:1} respectively, we get
	\begin{align}
		\E\norm{\rvx_m^{(r,k)} - \rvx^r}^2
		&\leq 4\eta^2\sum_{i=1}^{m}\sum_{j=0}^{\gK(i)}\E\norm{\rvg_{\pi_i}^{(r,j)} - \nabla F_{\pi_i} (\rvx_i^{(r,j)})}^2 + 4\eta^2\gK\sum_{i=1}^{m}\sum_{j=0}^{\gK(i)}\E\norm{\nabla F_{\pi_i} (\rvx_i^{(r,j)}) - \nabla F_{\pi_i} (\rvx^{r})}^2 \nonumber\\
		&\;+ 4\eta^2\gK\sum_{i=1}^{m}\sum_{j=0}^{\gK(i)}\E\norm{\nabla F_{\pi_i} (\rvx^{r})- \nabla F_{\pi_i} (\rvx^{\ast})}^2 + 4\eta^2\E\norm{\sum_{i=1}^{m}\sum_{j=0}^{\gK(i)}\nabla F_{\pi_i} (\rvx^{\ast})}^2,\label{eq:proof:lem:SSL strongly convex:client drift:2}
	\end{align}
	where $\displaystyle \gK \coloneqq \sum_{i=1}^{m}\sum_{j=0}^{\gK(i)}1 = (m-1)K+k$.
	For the first term on the right hand side in \eqref{eq:proof:lem:SSL strongly convex:client drift:2}, we have
	\[
	\E\norm{\rvg_{\pi_i}^{(r,j)} - \nabla F_{\pi_i} (\rvx_i^{(r,j)})}^2 \overset{\eqref{eq:asm:stochasticity}}{\leq} \sigma^2
	\]
	For the second term on the right hand side in \eqref{eq:proof:lem:SSL strongly convex:client drift:2}, we have
	\[
	\E\norm{\nabla F_{\pi_i} (\rvx_m^{(r,j)}) - \nabla F_{\pi_i} (\rvx^{r})}^2 \overset{\text{Asm.}~\ref{asm:L-smoothness}}{\leq} L^2 \E\norm{\rvx_i^{(r,j)} - \rvx^{r}}^2
	\]
	For the third term on the right hand side in \eqref{eq:proof:lem:SSL strongly convex:client drift:2}, we have
	\begin{align}
	\E \norm{\nabla F_{\pi_i} (\rvx^{r}) - \nabla F_{\pi_i} (\rvx^{\ast})}^2 \overset{\eqref{eq:bregman lower smooth}}{\leq} 2L D_{F_{\pi_i}}(\rvx^{r},\rvx^\ast)\label{eq:proof:lem:SSL strongly convex:client drift:4}
	\end{align}
	As a result, we can get
	\begin{align}
		\E\norm{\rvx_m^{(r,k)} - \rvx^r}^2 &\leq 4L^2\eta^2\gK\sum_{i=1}^{m}\sum_{j=0}^{\gK(i)}\E\norm{\rvx_i^{(r,j)}-x^r}^2 +8L\eta^2\gK\sum_{i=1}^{m}\sum_{j=0}^{\gK(i)}D_{F_{\pi_i}}(\rvx^{r},\rvx^\ast) \nonumber\\ 
		&\quad+ 4\gK\eta^2\sigma^2 + 4\eta^2\E\norm{\sum_{i=1}^{m}\sum_{j=0}^{\gK(i)}\nabla F_{\pi_i} (\rvx^{\ast})}^2
	\end{align}
	
	Returning to $\displaystyle E_r \coloneqq \sum_{m=1}^M\sum_{k=0}^{K-1} \E\norm{\rvx_m^{(r,k)} - \rvx^r}^2$, we have
	\begin{align}
		E_r &\leq 4L^2\eta^2\sum_{m=1}^M\sum_{k=0}^{K-1}\gK\sum_{i=1}^{M}\sum_{j=0}^{K-1}\E\norm{\rvx_i^{(r,j)}-x^r}^2 +8L\eta^2\sum_{m=1}^M\sum_{k=0}^{K-1}\gK\sum_{i=1}^{M}\sum_{j=0}^{K-1}D_{F_{\pi_i}}(\rvx^{r},\rvx^\ast) \pfcomment{$0 \leq j \leq \gK(i) \leq K-1,\forall i\in [m]$ and $1 \leq i \leq m \leq M$}\nonumber\\ 
		&\quad+ 4\sum_{m=1}^M\sum_{k=0}^{K-1}\gK\eta^2\sigma^2 + 4\eta^2\sum_{m=1}^M\sum_{k=0}^{K-1}\E\norm{\sum_{i=1}^{m}\sum_{j=0}^{\gK(i)}\nabla F_{\pi_i} (\rvx^{\ast})}^2\label{eq:proof:the per-round client drift of SL:3}
	\end{align}
	Then using
	\[
	\sum_{m=1}^M\sum_{k=0}^{K-1}\gK = \sum_{m=1}^M(m-1)K^2 + \sum_{m=1}^M\sum_{k=0}^{K-1}k = \frac{1}{2}M(M-1)K^2 + \frac{1}{2}MK(K-1) \leq \frac{1}{2}M^2K^2
	\]
	and
	\[
	\sum_{m=1}^M\sum_{k=0}^{K-1}\E\norm{\sum_{i=1}^{m}\sum_{j=0}^{\gK(i)}\nabla F_{\pi_i} (\rvx^{\ast})}^2 \overset{\eqref{eq:asm:heterogeneity:optimum}}{\leq} \frac{1}{3}M^2K^3\zeta_\ast^2
	\]
	where we use Lemma~\ref{lem:technical lemma SSL} with $\rvx_{\pi_i} = \nabla F_{\pi_i}(\rvx^\ast)$ and $\overline X = \frac{1}{M}\sum_{m=1}^M F_{\pi_i}(\rvx^\ast) = F(\rvx^\ast) = 0$, we have
	\begin{align}
		\E\left[E_r\right] &\leq 2L^2M^2K^2\eta^2\E\left[E_r\right] + 4LM^3K^3\eta^2D_{F}(\rvx^{r},\rvx^\ast) + 2M^2K^2\eta^2\sigma^2 + \frac{4}{3}M^2K^3\eta^2\zeta_\ast^2
	\end{align}
	After rearranging the preceding inequality, we get
	\begin{align}
		(1-2L^2M^2K^2\eta^2)\E\left[E_r\right] \leq 4LM^3K^3\eta^2D_{F}(\rvx^{r},\rvx^\ast) + 2M^2K^2\eta^2\sigma^2 + \frac{4}{3}M^2K^3\eta^2\zeta_\ast^2
	\end{align}
	Finally, using the choice of $\eta$, $LMK\eta \leq \frac{1}{6}$, which implies $1-2L^2M^2K^2\eta^2 \geq \frac{8}{9}$, we have
	\begin{align}
		\E\left[E_r\right]
		&\leq \frac{9}{2}LM^3K^3\eta^2D_{F}(\rvx^{r}, \rvx^\ast) + \frac{9}{4}M^2K^2\eta^2\sigma^2 + \frac{3}{2}M^2K^3\eta^2\zeta_\ast^2.
	\end{align}
	The claim of this lemma follows after taking unconditional expectations.
\end{proof}

\subsubsection{Tuning the learning rate}
Here we make a clear version of Lemma 1 in \cite{karimireddy2020scaffold} based on the works \cite{stich2019unified, stich19errorfeedback}.

\begin{lemma}[\cite{karimireddy2020scaffold}]\label{lem:stongly convex:tuning learning rate}
	Two non-negative sequences $\{r_t\}_{t\geq 0}$, $\{s_t\}_{t\geq 0}$, which satisfies the relation
	\begin{align}
		r_{t+1} \leq (1-a\gamma_t ) r_t - b\gamma_t s_t +  c \gamma_t^2,\label{eq:lem:stongly convex:tuning learning rate:problem}
	\end{align}
	for all $t \geq 0$ and 
	for parameters $b > 0$, $a, c \geq 0$ and non-negative learning rates $\{\gamma_t\}_{t \geq 0}$ with $\gamma_t \leq \frac{1}{d}$, $\forall t \geq 0$, for a parameter $d \geq a$, $d > 0$.
	
	\textbf{Selection of weights for average.}
	Then there exists a constant learning rate $\gamma_t =\gamma \leq \frac{1}{d}$ and the weights $w_t := (1-a\gamma)^{-(t+1)}$ and $W_T := \sum_{t=0}^T w_t$, making it hold that:
	\begin{align}
		\Psi_T = \frac{b}{W_T} \sum_{t=0}^{T}s_t w_t \leq 3ar_0(1-a\gamma)^{(T+1)} + c \gamma \leq 3ar_0 \exp \left[-a\gamma (T+1)\right] +c\gamma.\label{eq:lem:stongly convex:tuning learning rate:result1}
	\end{align}
	
	\textbf{Tuning the learning rate carefully.} By tuning the learning rate in \eqref{eq:lem:stongly convex:tuning learning rate:result1}, for $(T+1)\geq \frac{1}{2a\gamma}$, we have
	\begin{align}
		\Psi_T = \tilde \gO \left(  a r_0 \exp \left[-\frac{aT}{d} \right] + \frac{c}{aT} \right).\label{eq:lem:stongly convex:tuning learning rate:result2}
	\end{align}
\end{lemma}
\begin{proof}
	We start by rearranging~\eqref{eq:lem:stongly convex:tuning learning rate:problem} and multiplying both sides with $w_t$:
	\begin{align*}
		b s_t w_t \leq \frac{w_t (1-a\gamma) r_t}{\gamma} - \frac{w_t r_{t+1}}{\gamma}  + c \gamma w_t = \frac{w_{t-1} r_t}{\gamma} - \frac{w_t r_{t+1}}{\gamma}  + c \gamma w_t \,.
	\end{align*} 
	By summing from $t=0$ to $t=T$, we obtain a telescoping sum:
	\begin{align*}
		\frac{b}{W_T} \sum_{t=0}^{T} s_t w_t \leq \frac{1}{\gamma W_T} \left(w_0 (1-a\gamma)r_0 - w_{T} r_{T+1}\right) + c \gamma\,, 
	\end{align*}
	and hence
	\begin{align}
		\Psi_T = \frac{b}{W_T} \sum_{t=0}^{T}s_t w_t \leq \frac{b}{W_T} \sum_{t=0}^{T}s_t w_t + \frac{w_{T} r_{T+1}}{\gamma W_T} \leq \frac{r_0}{\gamma W_{T}} + c \gamma\label{eq:proof:lem:stongly convex:tuning learning rate}
	\end{align}
	Note that in the proof of Lemma 2 in \cite{stich2019unified}, they use $W_T \geq w_T = (1-a\gamma)^{-(T+1)}$ to estimate $W_T$. It is reasonable given that $w_T$ is extremely larger than all the terms $w_t$ ($t < T$) when $T$ is large. Yet \cite{karimireddy2020scaffold} goes further, showing that $W_T$ can be estimated more precisely:
		\[
		W_T = \sum_0^T w_t = (1-a\gamma)^{-(T+1)}\sum_{t=0}^T (1-a\gamma)^t = (1-a\gamma)^{-(T+1)}\left(\frac{1-(1-a\gamma)^{T+1}}{a\gamma}\right)
		\]
		When $(T+1)\geq \frac{1}{2a\gamma}$, $(1-a\gamma)^{T+1}\leq \exp(-a\gamma(T+1)) \leq e^{-\frac{1}{2}} \leq \frac{2}{3}$, so it follows that
		\[
		W_T = (1-a\gamma)^{-(T+1)}\left(\frac{1-(1-a\gamma)^{T+1}}{a\gamma}\right) \geq \frac{(1-a\gamma)^{-(T+1)}}{3a\gamma}
		\]
	With the estimates
	\begin{itemize}[leftmargin=2em]
		\item $W_T = (1-a\gamma)^{-(T+1)}\sum_{t=0}^T (1-a \gamma)^t \leq \frac{w_{T}}{a \gamma}$ (here we leverage $a\gamma \leq \frac{a}{d} \leq 1$),
		\item and $W_T \geq \frac{(1-a\gamma)^{-(T+1)}}{3a\gamma}$,
	\end{itemize}
	we can further simplified the terms on both sides in \eqref{eq:proof:lem:stongly convex:tuning learning rate}:
	\begin{align*}
		\Psi_T \leq 3ar_0(1-a\gamma)^{(T+1)} + c \gamma \leq 3ar_0 \exp \left[-a\gamma (T+1)\right] +c\gamma
	\end{align*}
	which is the first result of this lemma.
	
	Now the lemma follows by carefully tuning $\gamma$ in \eqref{eq:lem:stongly convex:tuning learning rate:result1}. Consider the two cases:
	\begin{itemize}[leftmargin=2em]
		\item If $\frac{1}{d} > \frac{\ln(\max\{2,a^2 r_0 T /c\})}{a T}$ then we choose $\gamma = \frac{\ln(\max\{2,a^2 r_0 T /c\})}{a T}$ and get that
		\begin{align*}
			\tilde \gO \left( 3a r_0 \exp[-\ln(\max\{2,a^2 r_0 T /c\})  ] \right)  + \tilde \gO \left( \frac{c}{aT} \right) = \tilde \gO \left( \frac{c}{aT} \right)\,,
		\end{align*}  
		as in case $2 \geq a^2 r_0 T/c$ it holds $a r_0 \leq \frac{2c}{aT}$.
		\item If otherwise $\frac{1}{2a(T+1)} \leq \frac{1}{d} \leq \frac{\ln(\max\{2,a^2 r_0 T /c\})}{a T}$ (Note $\frac{1}{2a(T+1)} \leq \frac{\ln(2)}{a T} \leq \frac{\ln(\max\{2,a^2 r_0 T /c\})}{a T}$) then we pick $\gamma = \frac{1}{d}$ and get that
		\begin{align*}
			3 ar_0 \exp \left[- \frac{aT}{d} \right] + \frac{c}{d} \leq  3 a r_0 \exp \left[- \frac{aT}{d} \right] + \frac{c \ln(\max\{2,a^2 r_0 T /c\})}{aT} = \tilde \gO \left(  a r_0 \exp \left[- \frac{aT}{d} \right] + \frac{c}{aT} \right)
		\end{align*}
	\end{itemize}
	Combining these two cases, we get
	\begin{align*}
		\Psi_T = \tilde \gO \left(  a r_0 \exp \left[- \frac{aT}{d} \right] + \frac{c}{aT} \right)
	\end{align*}
	
	Note that this lemma holds when $(T+1)\geq \frac{1}{2a\gamma}$, so it restricts the value of $T$ (Lemma 2 in \cite{stich2019unified} without this constraint).
\end{proof}

\subsubsection{Proof of strongly convex case of Theorem~\ref{thm:SSL convergence} and Corollary~\ref{cor:SSL convergence}}
\begin{proof}[Proof of strongly convex case of Theorem~\ref{thm:SSL convergence}]
Substituting \eqref{eq:lem:SSL strongly convex:client drift} into \eqref{eq:lem:SSL strongly convex:the per-round recursion} and using $LK\eta \leq \frac{1}{6}$, we can simplify the recursion as, 
\begin{align}
	\E\norm{\rvx^{r+1}-\rvx^*}^2
	&\leq \left(1-\tfrac{\mu MK\eta}{2}\right)\E\norm{\rvx^{r}- \rvx^\ast}^2 - \frac{1}{2}MK\eta\E\left[D_F(\rvx^r, \rvx^\ast)\right] + 3MK\eta^2\sigma^2 \nonumber\\
	&\quad+ \frac{45}{8}LM^2K^2\eta^3\sigma^2 + \frac{15}{4}LM^2K^3\eta^3\zeta_\ast^2
\end{align}
Let $\tilde{\eta} = MK\eta$, we have
\begin{align}
	\E\norm{\rvx^{r+1}-\rvx^*}^2 \leq \left(1-\frac{\mu \tilde{\eta}}{2}\right)\E\norm{\rvx^{r}- \rvx^\ast}^2 - \frac{\tilde{\eta}}{2}\E\left[D_F(\rvx^r, \rvx^\ast)\right] + \frac{3\tilde{\eta}^2\sigma^2}{MK} + \frac{45L\tilde{\eta}^3\sigma^2}{8MK} + \frac{15L \tilde{\eta}^3\zeta_\ast^2}{4M}\label{eq:thm:proof:SSL strongly convex:simplified per-round recursion:effective learning rate}
\end{align}

Applying Lemma~\ref{lem:stongly convex:tuning learning rate} with $t=r$ ($T=R$), $\gamma=\tilde\eta$, $r_{t} = \E\norm{\rvx^{r}- \rvx^*}^2$, $a = \frac{\mu}{2}$, $b=\frac{1}{2}$, $s_t = \E\left[D_F(\rvx^r, \rvx^\ast)\right] = \E\left[F(\rvx^{r}) - F(\rvx^*)\right]$, $w_t=(1-a\tilde\eta)^{-(r+1)}$, $c_1 = \frac{3\sigma^2}{MK}$, $c_2 = \frac{45L\sigma^2}{8MK} + \frac{15L\zeta_\ast^2}{4M}$ and $\frac{1}{d}=\frac{1}{6L}$ ($\tilde\eta=MK\eta\leq\frac{1}{6L}$).

Note that in Lemma~\ref{lem:stongly convex:tuning learning rate}, there are no terms containing $\gamma^3$. As the terms containing $\gamma^3$ is not the determining factor for the convergence rate, Lemma~\ref{lem:stongly convex:tuning learning rate} can also be applied to this case \citep{karimireddy2020scaffold,koloskova2020unified}.
\begin{align}
	F(\bar\rvx^R)-F(\rvx^\ast) \leq 3\mu\norm{\rvx^{0}- \rvx^*}^2 \exp\left[-\frac{1}{2}\mu\tilde\eta(R+1)\right] + \frac{6\tilde\eta\sigma^2}{MK} + \frac{45L\tilde\eta^2\sigma^2}{4MK} + \frac{15L\tilde\eta^2\zeta_\ast^2}{2M}
\end{align}
where $\bar\rvx^R = \frac{1}{W_R}\sum_{r=0}^Rw_r\rvx^r$. Here we have $F(\bar\rvx^R)-F(\rvx^\ast) \leq \frac{1}{W_R}\sum_{r=0}^Rw_r\E\left[F(\rvx^{r}) - F(\rvx^*)\right]$, where the convexity of $F$ and Jensen's inequality are used.

By tuning the learning rate carefully, we get
\begin{align}
	F(\bar\rvx^R)-F(\rvx^\ast) = \tilde\gO\left(\mu D^2 \exp\left(-\frac{\mu}{12L}R\right) + \frac{\sigma^2}{\mu MKR} + \frac{L\sigma^2}{\mu^2MKR^2} + \frac{L\zeta_\ast^2}{\mu^2MR^2}\right)
\end{align}
with $D=\norm{\rvx^{0}- \rvx^\ast}$.
\end{proof}

\subsection{General convex case}\label{subsec:general convex ssl}

\subsubsection{Tuning the learning rate}

\begin{lemma}[\cite{koloskova2020unified}]\label{lem:general convex:tuning learning rate}
	Two non-negative sequences $\{r_t\}_{t\geq 0}$, $\{s_t\}_{t\geq 0}$, which satisfies the relation
	\begin{align*}
		r_{t+1} \leq r_t - b \gamma_t s_t + c_1\gamma_t^2 + c_2\gamma_t^3 
	\end{align*}
	for all $t\geq 0$ and for parameters $b>0$, $c_1, c_2\geq 0$ and non-negative learning rates $\{\gamma_t\}_{t\geq 0}$ with $\gamma_t\leq \frac{1}{d}$, $\forall t\geq 0$, for a parameter $d>0$.
	
	\textbf{Selection of weights for average.} Then there exists a constant learning rate $\gamma = \gamma_t \leq \frac{1}{d}$ and the weights $w_t=1$ and $W_T = \sum_{t=0}^Tw_t$, making it hod that:
	\begin{align}
		\Psi_T  \coloneqq \frac{b}{T+1} \sum_{t=0}^T s_t \leq \frac{r_0}{\gamma(T+1)} + c_1 \gamma + c_2 \gamma^2\label{eq:lem:general convex:tuning learning rate:result1}
	\end{align}
	
	\textbf{Tuning the learning rate carefully.} By tuning the learning rate carefully in \eqref{eq:lem:general convex:tuning learning rate:result1}, we have
	\begin{align}
		\Psi_T \leq 2c_1^\frac{1}{2}\left(\frac{r_0}{T+1}\right)^\frac{1}{2} + 2c_2^\frac{1}{3}\left(\frac{r_0}{T+1}\right)^\frac{2}{3} + \frac{dr_0}{T+1}.
	\end{align}
\end{lemma}
\begin{proof} For constant learning rates $\gamma_t = \gamma$ we can derive the estimate
	\begin{align*}
		\Psi_T = \frac{1}{\gamma (T+1)} \sum_{t=0}^T \left( r_t - r_{t+1} \right) + c_1 \gamma + c_2 \gamma^2  \leq \frac{r_0}{\gamma(T+1)} + c_1 \gamma + c_2 \gamma^2,
	\end{align*}
	which is the first result \eqref{eq:lem:general convex:tuning learning rate:result1} of this lemma.
	Let $\frac{r_0}{\gamma (T+1)} = c_1\gamma$ and $\frac{r_0}{\gamma (T+1)} = c_2\gamma^2$, yielding two choices of $\gamma$, $\gamma = \left(\frac{r_0}{c_1(T+1)}\right)^\frac{1}{2}$ and $\gamma = \left(\frac{r_0}{c_2(T+1)}\right)^\frac{1}{3}$. Then choosing $\gamma = \min\left\{\left(\frac{r_0}{c_1(T+1)}\right)^\frac{1}{2}, \left(\frac{r_0}{c_2(T+1)}\right)^\frac{1}{3}, \frac{1}{d}\right\} \leq \frac{1}{d}$, there are three cases:
	\begin{itemize}[leftmargin=2em]
		\item If $\gamma = \frac{1}{d}$, which implies that $\gamma=\frac{1}{d}\leq \left(\frac{r_0}{c_1(T+1)}\right)^\frac{1}{2}$ and $\gamma=\frac{1}{d}\leq \left(\frac{r_0}{c_2(T+1)}\right)^\frac{1}{3}$, then
		\[
		\Psi_T \leq \frac{dr_0}{T+1} + \frac{c_1}{d} + \frac{c_2}{d^2} \leq \frac{dr_0}{T+1} + c_1^\frac{1}{2}\left(\frac{r_0}{T+1}\right)^\frac{1}{2} + c_2^\frac{1}{3}\left(\frac{r_0}{T+1}\right)^\frac{2}{3}
		\]
		\item If $\gamma = \left(\frac{r_0}{c_1(T+1)}\right)^\frac{1}{2}$, which implies that $\gamma = \left(\frac{r_0}{c_1(T+1)}\right)^\frac{1}{2} \leq \left(\frac{r_0}{c_2(T+1)}\right)^\frac{1}{3}$, then
		\[
		\Psi_T \leq 2c_1 \left(\frac{r_0}{c_1(T+1)}\right)^\frac{1}{2} + c_2 \left(\frac{r_0}{c_1(T+1)}\right) \leq 2c_1^\frac{1}{2}\left(\frac{r_0}{T+1}\right)^\frac{1}{2} + c_2^\frac{1}{3}\left(\frac{r_0}{T+1}\right)^\frac{2}{3}
		\]
		\item If $\gamma = \left(\frac{r_0}{c_2(T+1)}\right)^\frac{1}{3}$, which implies that $\gamma = \left(\frac{r_0}{c_2(T+1)}\right)^\frac{1}{3} \leq \left(\frac{r_0}{c_1(T+1)}\right)^\frac{1}{2}$, then
		\[
		\Psi_T \leq c_1 \left(\frac{r_0}{c_2(T+1)}\right)^\frac{1}{3} + 2c_2^\frac{1}{3}\left(\frac{r_0}{T+1}\right)^\frac{2}{3} \leq c_1^\frac{1}{2}\left(\frac{r_0}{T+1}\right)^\frac{1}{2} + 2c_2^\frac{1}{3}\left(\frac{r_0}{T+1}\right)^\frac{2}{3}
		\]
	\end{itemize}
	Combining these three cases, we get the second result of this lemma.
\end{proof}

\subsubsection{Proof of general convex case of Theorem~\ref{thm:SSL convergence} and Corollary~\ref{cor:SSL convergence}}
\begin{proof}[Proof of general convex case of Theorem~\ref{thm:SSL convergence}]
Let $\mu=0$ in \eqref{eq:thm:proof:SSL strongly convex:simplified per-round recursion:effective learning rate}, we get the simplified recursion of general convex case,
\begin{align}
	\E\norm{\rvx^{r+1}-\rvx^*}^2 \leq \E\norm{\rvx^{r}- \rvx^\ast}^2 - \frac{\tilde{\eta}}{2}\E\left[D_F(\rvx^r, \rvx^\ast)\right] + \frac{3\tilde{\eta}^2\sigma^2}{MK} + \frac{45L\tilde{\eta}^3\sigma^2}{8MK} + \frac{15L \tilde{\eta}^3\zeta_\ast^2}{4M}
\end{align}
Then Applying Lemma~\ref{lem:general convex:tuning learning rate} with $t=r$ ($T=R$), $\gamma=\tilde\eta$, $r_{t} = \E\norm{\rvx^{r}- \rvx^*}^2$, $b=\frac{1}{2}$, $s_t = \E\left[D_F(\rvx^r, \rvx^\ast)\right] = \E\left[F(\rvx^{r}) - F(\rvx^*)\right]$, $w_t=1$, $c_1 = \frac{3\sigma^2}{MK}$, $c_2 = \frac{45L\sigma^2}{8MK} + \frac{15L\zeta_\ast^2}{4M}$ and $\frac{1}{d}=\frac{1}{6L}$ ($\tilde\eta=MK\eta\leq\frac{1}{6L}$), we can have
\begin{align}
	F(\bar\rvx^R)-F(\rvx^\ast) \leq \frac{2\norm{\rvx^{0}- \rvx^\ast}^2}{\tilde\eta R} + \frac{6\tilde\eta\sigma^2}{MK} + \frac{45L\tilde\eta^2\sigma^2}{4MK} + \frac{15L\tilde\eta^2\zeta_\ast^2}{2M},
\end{align}
where $\bar\rvx^R = \frac{1}{W_R}\sum_{r=0}^Rw_r\rvx^r$. Here we have
$F(\bar\rvx^R)-F(\rvx^\ast) \leq \frac{1}{W_R}\sum_{r=0}^Rw_r\E\left[F(\rvx^{r}) - F(\rvx^*)\right]$,
with the convexity of $F$ and Jensen's inequality. By tuning the learning rate carefully, we get
\begin{align}
	F(\bar\rvx^R)-F(\rvx^\ast) = \gO\left(\frac{\sigma D}{\sqrt{MKR}} + \frac{\left(L\sigma^2D^4\right)^{1/3}}{(MK)^{1/3}R^{2/3}} + \frac{\left(L\zeta_\ast^2D^4\right)^{1/3}}{M^{1/3}R^{2/3}} + \frac{LD^2}{R}\right),
\end{align}
where $D = \norm{\rvx^{0}- \rvx^\ast}$.
\end{proof}

\subsection{Non-convex case}\label{subsec:non-convex ssl}
\begin{lemma}\label{lem:ssl non-convex:per-round recursion}
	Under the same assumptions as Theorem~\ref{thm:SSL convergence}, we can find the per-round recursion as
	\begin{align}
		\E [F(\rvx^{r+1}) - F(\rvx^{r})] \leq -\frac{MK\eta}{2}\norm{\nabla F(\rvx^{r})}^2 +  LMK\eta^2\sigma^2 + \frac{L^2\eta}{2}\sum_{m=1}^M\sum_{k=0}^{K-1}\E\norm{\rvx_m^{(r,k)}-\rvx^r}^2\label{eq:lem:ssl non-convex:per-round recursion}
	\end{align}
\end{lemma}
\begin{proof}
	Without otherwise stated, \textbf{the expectation is conditioned on} $\rvx^r$. Beginning from $L$-smoothness, 
	\[
	D_F(\rvx^{r+1}, \rvx^r) = F(\rvx^{r+1}) - F(\rvx^r) - \inp{\nabla F(\rvx^r)}{\rvx^{r+1}-\rvx^r} \leq \frac{L}{2}\norm{\rvx^{r+1} - \rvx^r}^2,
	\] and the update rule of \texttt{SSL}, 
	\[
	\displaystyle \rvx^{r+1} = \rvx^r - \eta\sum_{m=1}^M\sum_{k=0}^{K-1}\rvg_{\pi_m}^{(r,k)},
	\]
	where $\rvg_m^{(r,k)} = \nabla f_m(\rvx_m^{(r,k)}; \xi_m^{(r,k)})$, we have
	\begin{align}
		&\E [F(\rvx^{r+1})] - F(\rvx^{r})\leq - M\eta \sum_{k=0}^{K-1}\E\inp{\nabla F(\rvx^{r})}{\frac{1}{M}\sum_{m=1}^M\nabla F_{\pi_m}(\rvx_m^{(r,k)})} + \frac{L}{2}\E\norm{ \frac{\eta}{M}\sum_{m=1}^{M}\sum_{k=0}^{K-1}\rvg_{\pi_m}^{r,k} }^2\label{eq:proof:lem:ssl non-convex:per-round recursion:1}
	\end{align}
	For the first term on the right hand side in \eqref{eq:proof:lem:ssl non-convex:per-round recursion:1}, using $\inp{a}{b} = \frac{1}{2}\norm{a}^2 + \frac{1}{2}\norm{b}^2 - \frac{1}{2}\norm{a-b}^2$ with $a = \nabla F(\rvx^{r})$ and $b = \frac{1}{M}\sum_{m=1}^M\nabla F_{\pi_m}(\rvx_m^{(r,k)})$, we have
	\begin{align}
		&- M\eta \sum_{k=0}^{K-1}\E\inp{\nabla F(\rvx^{r})}{\frac{1}{M}\sum_{m=1}^M\nabla F_{\pi_m}(\rvx_m^{(r,k)})} \nonumber\\
		&\leq - \frac{M\eta}{2} \sum_{k=0}^{K-1}\E\left[\norm{\nabla F(\rvx^{r})}^2 + \norm{\frac{1}{M}\sum_{m=1}^M\nabla F_{\pi_m}(\rvx_m^{(r,k)})}^2 - \norm{\frac{1}{M}\sum_{m=1}^M\nabla F_{\pi_m}(\rvx_m^{(r,k)}) - \nabla F(\rvx^{r})}^2\right] \nonumber\\
		&\overset{\eqref{eq:jensen norm},\text{Asm.~}\ref{asm:L-smoothness}}{\leq} -\frac{MK\eta}{2}\norm{\nabla F(\rvx^{r})}^2 - \frac{M\eta}{2}\sum_{k=0}^{K-1}\E\norm{\frac{1}{M}\sum_{m=1}^M\nabla F_{\pi_m}(\rvx_m^{(r,k)})}^2 + \frac{L^2\eta}{2}\sum_{m=1}^M\sum_{k=0}^{K-1}\E\norm{\rvx_m^{(r,k)}-\rvx^r}^2\label{eq:proof:lem:ssl non-convex:per-round recursion:2}
	\end{align}
	For the second term on the right hand side in \eqref{eq:proof:lem:ssl non-convex:per-round recursion:1}, we have
	\begin{align}
		\frac{L}{2}\E\norm{ \eta\sum_{m=1}^{M}\sum_{k=0}^{K-1}\rvg_{\pi_m}^{r,k} }^2 &\leq L\eta^2\E\norm{\sum_{m=1}^{M}\sum_{k=0}^{K-1}\rvg_{\pi_m}^{r,k}-\sum_{m=1}^{M}\sum_{k=0}^{K-1} \nabla F_{\pi_m}(\rvx_m^{(r,k)})}^2 + L\eta^2\E\norm{\sum_{m=1}^{M}\sum_{k=0}^{K-1} \nabla F_{\pi_m}(\rvx_m^{(r,k)})}^2\nonumber\\
		&\leq LMK\eta^2\sigma^2 + L\eta^2\E\norm{\sum_{m=1}^{M}\sum_{k=0}^{K-1} \nabla F_{\pi_m}(\rvx_m^{(r,k)})}^2\label{eq:proof:lem:ssl non-convex:per-round recursion:3}
	\end{align}
	
	Substituting \eqref{eq:proof:lem:ssl non-convex:per-round recursion:2} and \eqref{eq:proof:lem:ssl non-convex:per-round recursion:3} into \eqref{eq:proof:lem:ssl non-convex:per-round recursion:1}, we have
	\begin{align}
		\E [F(\rvx^{r+1})] - F(\rvx^{r}) &\overset{\eqref{eq:jensen norm}}{\leq} -\frac{MK\eta}{2}\norm{\nabla F(\rvx^{r})}^2 + \frac{L^2\eta}{2}\sum_{m=1}^M\sum_{k=0}^{K-1}\E\norm{\rvx_m^{(r,k)}-\rvx^r}^2
		+ LMK\eta^2\sigma^2 \nonumber\\
		&\quad-\frac{\eta}{2}(1-2LMK\eta)\frac{1}{M}\sum_{m=1}^M\sum_{k=0}^{K-1} \norm{\nabla F_{\pi_m}(\rvx_m^{(r,k)})}^2
	\end{align}
	Considering $MK\eta\leq \frac{1}{6L}$, the last term on the right hand side in the preceding inequality is negative. As a consequence, we have
	\begin{align*}
		\E [F(\rvx^{r+1})] - F(\rvx^{r}) &\overset{\eqref{eq:jensen norm}}{\leq} -\frac{MK\eta}{2}\norm{\nabla F(\rvx^{r})}^2 +  LMK\eta^2\sigma^2 + \frac{L^2\eta}{2}\sum_{m=1}^M\sum_{k=0}^{K-1}\E\norm{\rvx_m^{(r,k)}-\rvx^r}^2
	\end{align*}
	The claim follows after taking unconditional expectation.
\end{proof}

\subsubsection{Bounding the client drift with \eqref{eq:asm:heterogeneity:everywhere}}

\begin{lemma}\label{lem:ssl non-convex:per-round client drift}
	Under the same assumptions as Theorem~\ref{thm:fedavg convergence}, for any learning rate satisfying $\eta \leq \frac{1}{6LK}$, the client drift $E_r$, defined as
	\begin{align*}
		E_r \coloneqq \sum_{m=1}^M\sum_{k=0}^{K-1} \E\norm{\rvx_m^{(r,k)} - \rvx^r}^2,
	\end{align*}
	is bounded, as given by:
	\begin{align}
		\E\left[E_r\right]
		&\leq \frac{3}{2}M^3K^3\eta^2\E\norm{\nabla F(\rvx^r)}^2 +  \frac{3}{2}M^2K^3B^2\eta^2\E\norm{\nabla F(\rvx^r)}^2 + \frac{9}{4}M^2K^2\eta^2\sigma^2 + \frac{3}{2}M^2K^3\eta^2\zeta^2\label{eq:lem:ssl non-convex:per-round client drift}
	\end{align}
\end{lemma}
\begin{proof}
	Without otherwise stated, \textbf{the expectation is conditioned on} $\rvx^r$. Beginning with \(\E\norm{\rvx_m^{(r,k)} - \rvx^r}^2\). Considering 
	\begin{align}
		\rvx_m^{(r,k)}- \rvx^{r} = \rvx_m^{(r,k)}-\rvx_m^{(r,0)} + \rvx_m^{(r,0)}-\rvx_{m-1}^{(r,0)}+\cdots+\rvx_2^{(r,0)}-\rvx_1^{(r,0)} = - \eta\sum_{i=1}^{m} \sum_{j=0}^{\gK(i)} \rvg_{\pi_i}^{(r,j)},
	\end{align}
	with $\mathcal{K}(i) \coloneqq \begin{cases}
		K-1,\; 1 \leq i \leq m-1\\
		k-1,\; i = m
	\end{cases},$
	we have
	\begin{align}
		\E\norm{\rvx_m^{(r,k)} - \rvx^r}^2 &= \eta^2\E \norm{\sum_{i=1}^{m}\sum_{j=0}^{\gK(i)} \rvg_{\pi_i}^{(r,j)}}^2
	\end{align}
	Then using the Jensen's inequality to the preceding equation, we have
	\begin{align}
		&\E\norm{\rvx_m^{(r,k)} - \rvx^r}^2 \nonumber\\
		&\leq 4\eta^2\E \norm{\sum_{i=1}^{m}\sum_{j=0}^{\gK(i)} \rvg_{\pi_i}^{(r,j)} - \sum_{i=1}^{m}\sum_{j=0}^{\gK(i)}\nabla F_{\pi_i} (\rvx_i^{(r,j)})}^2 + 4\eta^2\E\norm{\sum_{i=1}^{m}\sum_{j=0}^{\gK(i)}\nabla F_{\pi_i} (\rvx_i^{(r,j)}) - \sum_{i=1}^{m}\sum_{j=0}^{\gK(i)}\nabla F_{\pi_i} (\rvx^{r})}^2 \nonumber\\
		&\quad+ 4\eta^2\E \norm{\sum_{i=1}^{m}\sum_{j=0}^{\gK(i)}\nabla F_{\pi_i} (\rvx^{r}) - \sum_{i=1}^{m}\sum_{j=0}^{\gK(i)}\nabla F(\rvx^{r})}^2 + 4\eta^2\E\norm{\sum_{i=1}^{m}\sum_{j=0}^{\gK(i)}\nabla F (\rvx^{r})}^2\label{eq:proof:lem:ssl non-convex:client drift:1}
	\end{align}
	Applying Lemma~\ref{lem:martingale difference property} to the first term and Jensen’s inequality to the last three terms on the right hand side in \eqref{eq:proof:lem:ssl non-convex:client drift:1} respectively, we get
	\begin{align}
		\E\norm{\rvx_m^{(r,k)} - \rvx^r}^2
		&\leq 4\eta^2\sum_{i=1}^{m}\sum_{j=0}^{\gK(i)}\E\norm{\rvg_{\pi_i}^{(r,j)} - \nabla F_{\pi_i} (\rvx_i^{(r,j)})}^2 + 4\eta^2\gK\sum_{i=1}^{m}\sum_{j=0}^{\gK(i)}\E\norm{\nabla F_{\pi_i} (\rvx_i^{(r,j)}) - \nabla F_{\pi_i} (\rvx^{r})}^2 \nonumber\\
		&+ 4\eta^2\gA + 4\eta^2\gK^2\E\norm{\nabla F (\rvx^{r})}^2\label{eq:proof:lem:ssl non-convex:client drift:2}
	\end{align}
	where $\displaystyle \gA=\E \norm{\sum_{i=1}^{m}\sum_{j=0}^{\gK(i)}\nabla F_{\pi_i} (\rvx^{r}) - \sum_{i=1}^{m}\sum_{j=0}^{\gK(i)}\nabla F(\rvx^{r})}^2$ and $\displaystyle \gK \coloneqq \sum_{i=1}^{m}\sum_{j=0}^{\gK(i)}1 = (m-1)K+k$.
	For the first term on the right hand side in \eqref{eq:proof:lem:SSL strongly convex:client drift:2}, we have
	\[
	\E\norm{\rvg_{\pi_i}^{(r,j)} - \nabla F_{\pi_i} (\rvx_i^{(r,j)})}^2 \overset{\eqref{eq:asm:stochasticity}}{\leq} \sigma^2
	\]
	For the second term on the right hand side in \eqref{eq:proof:lem:SSL strongly convex:client drift:2}, we have
	\[
	\E\norm{\nabla F_{\pi_i} (\rvx_m^{(r,j)}) - \nabla F_{\pi_i} (\rvx^{r})}^2 \overset{\text{Asm.}~\ref{asm:L-smoothness}}{\leq} L^2 \E\norm{\rvx_i^{(r,j)} - \rvx^{r}}^2
	\]
	As a result, we can get
	\begin{align*}
		\E\norm{\rvx_m^{(r,k)} - \rvx^r}^2 &\leq 4L^2\eta^2\gK\sum_{i=1}^{m}\sum_{j=0}^{\gK(i)}\E\norm{\rvx_i^{(r,j)}-x^r}^2 +4\eta^2\gK^2\E\norm{\nabla F (\rvx^{r})}^2 + 4\gK\eta^2\sigma^2 + 4\eta^2\gA
	\end{align*}
	
	Returning to $\displaystyle E_r \coloneqq \sum_{m=1}^M\sum_{k=0}^{K-1} \E\norm{\rvx_m^{(r,k)} - \rvx^r}^2$, we have
	\begin{align}
		\E[E_r] &\leq 4L^2\eta^2\sum_{m=1}^M\sum_{k=0}^{K-1}\gK\sum_{i=1}^{M}\sum_{j=0}^{K-1}\E\norm{\rvx_i^{(r,j)}-x^r}^2 +4\eta^2\E\norm{\nabla F (\rvx^{r})}^2\sum_{m=1}^M\sum_{k=0}^{K-1}\gK^2 \pfcomment{$0 \leq j \leq \gK(i) \leq K-1,\forall i\in [m]$ and $1 \leq i \leq m \leq M$}\nonumber\\ 
		&\quad+ 4\sum_{m=1}^M\sum_{k=0}^{K-1}\gK\eta^2\sigma^2 + 4\eta^2\sum_{m=1}^M\sum_{k=0}^{K-1}\gA\label{eq:proof:lem:ssl non-convex:client drift:3}
	\end{align}
	Here we have
	$\displaystyle \sum_{m=1}^M\sum_{k=0}^{K-1}\gK \leq \frac{1}{2}M^2K^2$ and $\displaystyle \sum_{m=1}^M\sum_{k=0}^{K-1}\gK^2 \leq \frac{M^3K^3}{3}$. For the last term on the right hand side in \eqref{eq:proof:lem:ssl non-convex:client drift:3}, we have
	\begin{align}
		\sum_{m=1}^M\sum_{k=0}^{K-1}\gA &= \sum_{m=1}^M\sum_{k=0}^{K-1}\E \norm{\sum_{i=1}^{m}\sum_{j=0}^{\gK(i)}\left[\nabla F_{\pi_i} (\rvx^{r}) - \nabla F(\rvx^{r})\right]}^2 \nonumber\\
		&\overset{\text{Lem.}~\ref{lem:technical lemma SSL}}{\leq} \frac{1}{3}M^2K^3 \frac{1}{M}\sum_{i=1}^M \norm{\nabla F_{\pi_i}(\rvx^r) - \nabla F(\rvx^r)}^2\nonumber\\
		&\overset{\eqref{eq:asm:heterogeneity:everywhere}}{\leq} \frac{1}{3}M^2K^3 (B^2\norm{\nabla F(\rvx^r)}^2+\zeta^2)
	\end{align}
	where we use Lemma~\ref{lem:technical lemma SSL} with $\rvx_{\pi_i} = \nabla F_{\pi_i}(\rvx^r)$ and $\overline X = \frac{1}{M}\sum_{m=1}^M F_{\pi_i}(\rvx^r) = F(\rvx^r)$. Then we get
	\begin{align}
		\E\left[E_r\right] &\leq 2L^2M^2K^2\eta^2\E\left[E_r\right] + \frac{4}{3}M^3K^3\eta^2\E\norm{\nabla F(\rvx^r)}^2 +  \frac{4}{3}M^2K^3B^2\eta^2\E\norm{\nabla F(\rvx^r)}^2\nonumber\\
		&\quad+ 2M^2K^2\eta^2\sigma^2 + \frac{4}{3}M^2K^3\eta^2\zeta^2 
	\end{align}
	After rearranging the preceding inequality, we get
	\begin{align}
		(1-2L^2M^2K^2\eta^2)\E\left[E_r\right] 
		&\leq \frac{4}{3}M^3K^3\eta^2\E\norm{\nabla F(\rvx^r)}^2 +  \frac{4}{3}M^2K^3B^2\eta^2\E\norm{\nabla F(\rvx^r)}^2\nonumber\\
		&\quad+ 2M^2K^2\eta^2\sigma^2 + \frac{4}{3}M^2K^3\eta^2\zeta^2 
	\end{align}
	Finally, using the choice of $\eta$, $LMK\eta \leq \frac{1}{6}$, which implies $1-2L^2M^2K^2\eta^2 \geq \frac{8}{9}$, we have
	\begin{align*}
		\E\left[E_r\right]
		&\leq \frac{3}{2}M^3K^3\eta^2\E\norm{\nabla F(\rvx^r)}^2 +  \frac{3}{2}M^2K^3B^2\eta^2\E\norm{\nabla F(\rvx^r)}^2 + \frac{9}{4}M^2K^2\eta^2\sigma^2 + \frac{3}{2}M^2K^3\eta^2\zeta^2
	\end{align*}
\end{proof}

\subsubsection{Proof of non-convex case of Theorem~\ref{thm:SSL convergence} and Corollary~\ref{cor:SSL convergence}}

\begin{proof}[Proof of non-convex case of Theorem~\ref{thm:SSL convergence}]
	Substituting \eqref{eq:lem:ssl non-convex:per-round client drift} (Lemma~\ref{lem:ssl non-convex:per-round client drift}) into \eqref{eq:lem:ssl non-convex:per-round recursion} (Lemma~\ref{lem:ssl non-convex:per-round recursion}), and using $MK\eta \leq \frac{1}{6L(B^2+1)}$we have
	\begin{align*}
		\E [F(\rvx^{r+1})] - F(\rvx^{r}) 
		&\leq -MK\eta\left[\frac{1}{2}-\frac{3}{4}L^2M^2K^2\eta^2-\frac{3}{4}L^2MK^2\eta^2\right]\E\norm{\nabla F(\rvx^{r})}^2 \nonumber\\
		&\quad+ LMK\eta^2\sigma^2 + \frac{9}{8}L^2M^2K^2\eta^3\sigma^2 + \frac{3}{4}L^2M^2K^3\eta^3\zeta^2 \nonumber\\
		&\leq -\frac{MK\eta}{3}\E\norm{\nabla F(\rvx)}^2 +LMK\eta^2\sigma^2 + \frac{9}{8}L^2M^2K^2\eta^3\sigma^2 + \frac{3}{4}L^2M^2K^3\eta^3\zeta^2
	\end{align*}
	Letting $\tilde \eta \coloneqq MK\eta$ and minus both sides by $F^\ast$, we have
	\begin{align}
		\E [F(\rvx^{r+1})- F^\ast] \leq F(\rvx^{r}) - F^\ast - \frac{\tilde\eta}{3}\E\norm{\nabla F(\rvx)}^2 + \frac{L\tilde\eta^2\sigma^2}{MK} + \frac{9L^2\tilde\eta^3\sigma^2}{8MK} + \frac{3L^2\tilde\eta^3\zeta^2}{4M}
	\end{align}
	Then applying Lemma~\ref{lem:general convex:tuning learning rate} with $t=r$ ($T=R$), $\gamma=\tilde\eta$, $r_{t} = \E[F(\rvx^r)-F^\ast]$, $b=\frac{1}{3}$, $s_t = \E\norm{\nabla F(\rvx)}^2$, $w_t=1$, $c_1 = \frac{L\sigma^2}{MK}$, $c_2 = \frac{9L^2\sigma^2}{8MK} + \frac{3L^2\zeta^2}{4M}$ and $\frac{1}{d}=\frac{1}{6L(B^2+1)}$ ($\tilde\eta=MK\eta\leq\frac{1}{6L(B^2+1)}$), we have
	\begin{align}
		\frac{1}{R+1}\sum_{r=0}^{R}\E\norm{\nabla F(\rvx)}^2 \leq \frac{3\E [F(\rvx^{0}) - F^\ast]}{\tilde\eta R} + \frac{3L\tilde\eta\sigma^2}{MK} + \frac{27L^2\tilde\eta^2\sigma^2}{8MK} + \frac{9L^2\tilde\eta^2\zeta^2}{4M}
	\end{align}
	By tuning the learning rate carefully, we get
	\begin{align}
		\min_{0\leq r\leq R}\E\norm{\nabla F(\rvx)}^2 = \gO\left(\frac{ \left(\sigma^2LA\right)^{1/2}}{\sqrt{MKR}} + \frac{\left(L^2\sigma^2A^2\right)^{1/3}}{(MK)^{1/3}R^{2/3}} + \frac{\left(L^2\zeta^2A^2\right)^{1/3}}{M^{1/3}R^{2/3}} + \frac{LB^2A}{R}\right),
	\end{align}
	where $A \coloneqq F(\rvx^{0})- F^\ast$ and $\min_{0\leq r\leq R} \E\norm{\nabla F(\rvx)}^2 \leq  \frac{1}{R+1}\sum_{r=0}^{R}\E\norm{\nabla F(\rvx)}^2$.
\end{proof}
\section{Proofs of Theorem~\ref{thm:fedavg convergence}}\label{sec:proof FedAvg}
Here we slightly improve the convergence guarantee for the strongly convex case by combining the work of \cite{karimireddy2020scaffold,koloskova2020unified}. Moreover, we reproduce the guarantees for the general convex and non-convex cases based on \cite{karimireddy2020scaffold} for completeness. The results are given in Theorem~\ref{thm:fedavg convergence}.

We provide the proof of Theorem~\ref{thm:fedavg convergence} for the strongly convex, general convex and non-convex cases in \ref{subsec:strongly convex fedavg}, \ref{subsec:general convex fedavg} and \ref{subsec:non-convex fedavg}, respectively.


\begin{theorem}\label{thm:fedavg convergence}
	For \texttt{FedAvg}, there exists a constant effective learning rate $\tilde\eta \coloneqq K\eta$, making the weighted average of the model parameter $\bar{\rvx}^R\coloneqq \frac{1}{R}\sum_{r=0}^{R}w_r\rvx^r$ satisfy the following upper bounds:
	\begin{itemize}[leftmargin=2em]
		\item \textbf{Strongly convex}: Under Assumptions~\ref{asm:L-smoothness}, \ref{asm:stochasticity}, \eqref{eq:asm:heterogeneity:optimum} of \ref{asm:heterogeneity}, there exist one constant effective learning rate $\frac{1}{\mu R}\leq \tilde\eta \leq \frac{1}{6L}$ and weights $w_r=(1-\frac{\mu\tilde\eta}{2})^{-(r+1)}$, making it hold that
		\begin{flalign}
			F(\bar\rvx^R)-F(\rvx^\ast) \leq 2\mu\norm{\rvx^{0}- \rvx^*}^2 \exp\left(-\frac{1}{2}\mu\tilde\eta R\right) + \frac{4\tilde\eta\sigma^2}{MK} + \frac{15L\tilde\eta^2\sigma^2}{2K} + 5L\tilde\eta^2\zeta_\ast^2\label{eq:thm1:strongly convex} &&
		\end{flalign}
		\item \textbf{General convex}: Under Assumptions~\ref{asm:L-smoothness}, \ref{asm:stochasticity}, \eqref{eq:asm:heterogeneity:optimum} of \ref{asm:heterogeneity}, there exist one constant effective learning rate $\tilde\eta \leq \frac{1}{6L}$ and weights $w_r=1$, making it hold that
		\begin{flalign}
			F(\bar\rvx^R)-F(\rvx^\ast) \leq \frac{4\norm{\rvx^{0}- \rvx^\ast}^2}{3\tilde\eta R} + \frac{4\tilde\eta\sigma^2}{MK} + \frac{15L\tilde\eta^2\sigma^2}{2K} + 5L\tilde\eta^2\zeta_\ast^2 &&
		\end{flalign}
		\item \textbf{Non-convex}: Under Assumptions~\ref{asm:L-smoothness}, \ref{asm:stochasticity}, \eqref{eq:asm:heterogeneity:everywhere} of \ref{asm:heterogeneity}, there exist one constant effective learning rate $\tilde\eta \leq \frac{1}{6L(B^2+1)}$ and weights $w_r=1$, making it hold that
		\begin{flalign}
			\min_{0\leq r\leq R}\E\norm{\nabla F(\rvx)}^2 \leq \frac{3\E [F(\rvx^{0}) - F^\ast]}{\tilde\eta R} + \frac{3L\tilde\eta\sigma^2}{MK} + \frac{27L^2\tilde\eta^2\sigma^2}{8K} + \frac{9}{4}L^2\tilde\eta^2\zeta^2 &&
		\end{flalign}
	\end{itemize}
\end{theorem}

\subsection{Strongly convex case}\label{subsec:strongly convex fedavg}

\subsubsection{Find the per-round recursion}
\begin{lemma}\label{lem:fedavg strongly convex:per-round recursion}
Under the same assumptions as Theorem~\ref{thm:fedavg convergence}, we can find the per-round recursion as
\begin{align}
\E\norm{\rvx^{r+1}-\rvx^*}^2
&\leq \left(1-\tfrac{\mu K\eta}{2}\right)\E\norm{\rvx^{r}- \rvx^*}^2 - K\eta(2-6LK\eta)\E[D_F(\rvx^r, \rvx^*)] + \frac{3K \eta^2\sigma^2}{M}\nonumber\\
&\quad\quad + L\eta(2+3LK\eta) \frac{1}{M}\sum_{m=1}^M\sum_{k=0}^{K-1} \E\norm{\rvx_m^{(r,k)} - \rvx^r}^2.\label{eq:lem:fedavg strongly convex:per-round recursion}
\end{align}
\end{lemma}
\begin{proof}
Without otherwise stated, \textbf{the expectation is conditioned on} $\rvx^r$.
Recalling the update rule of FL, 
$\displaystyle
\displaystyle \rvx^{r+1} = \rvx^r - \frac{\eta}{M}\sum_{m=1}^M\sum_{k=0}^{K-1} \rvg_m^{(r,k)},
$
where $\rvg_m^{(r,k)} = \nabla f_m(\rvx_m^{(r,k)}; \xi_m^{(r,k)})$, we have
\begin{align}
&\E\norm{\rvx^{r+1}-\rvx^\ast}^2 = \E\norm{\rvx^{r}- \frac{\eta}{M}\sum_{m=1}^M\sum_{k=0}^{K-1} \rvg_m^{(r,k)} - \rvx^*}^2 \nonumber\\
&= \E\norm{\rvx^{r}- \rvx^\ast}^2 - 2\E\inp{\rvx^r-\rvx^\ast}{\frac{\eta}{M}\sum_{m=1}^M\sum_{k=0}^{K-1} \nabla F_m(\rvx_m^{(r,k)})} + \E\norm{ \frac{\eta}{M}\sum_{m=1}^M\sum_{k=0}^{K-1} \rvg_m^{(r,k)}}^2\label{eq:proof:lem:fedavg strongly convex:per-round recursion:1},
\end{align}
where we use $\E\left[\rvg_m^{(r,k)}\right] = \nabla F_m(\rvx_m^{(r,k)})$ for all $r, m, k$ in the second equality.

Using Lemma~\ref{lem:perturbed strong convexity} with $\vx=\rvx_m^{(r,k)}$, $\vy=\rvx^*$, $\vz=\rvx^r$ and $h = F_m$, we can bound the second term on the right hand side in \eqref{eq:proof:lem:fedavg strongly convex:per-round recursion:1}:
\begin{align}
&- 2\E\inp{\rvx^r-\rvx^\ast}{\frac{\eta}{M}\sum_{m=1}^M\sum_{k=0}^{K-1} \nabla F_m(\rvx_m^{(r,k)})}\nonumber\\
&=- 2\frac{\eta}{M}\sum_{m=1}^M\sum_{k=0}^{K-1}\E\left\langle \rvx^{r}- \rvx^*, \nabla F_m(\rvx_m^{(r,k)})\right\rangle\nonumber\\
&\leq- 2\frac{\eta}{M}\sum_{m=1}^M\sum_{k=0}^{K-1}\E\left( F_m(\rvx^{r}) - F_m(\rvx^*) + \frac{\mu}{4} \norm{\rvx^r - \rvx^*}^2 - L \norm{\rvx_m^{(r,k)} - \rvx^r}^2\right) \nonumber\\
&\leq- 2K\eta\left( F(\rvx^{r}) - F(\rvx^*)\right) - \frac{\mu K\eta}{2}\norm{\rvx^r - \rvx^*}^2 + 2L \frac{\eta}{M}\sum_{m=1}^M\sum_{k=0}^{K-1} \E\norm{\rvx_m^{(r,k)} - \rvx^r}^2\label{eq:proof:lem:fedavg strongly convex:per-round recursion:2}
\end{align}

For the first term on the right hand side in \eqref{eq:proof:lem:fedavg strongly convex:per-round recursion:1}, using Jensen's inequality, we have
\begin{align}
&\E\norm{\frac{\eta}{M}\sum_{m=1}^M\sum_{k=0}^{K-1} \rvg_m^{(r,k)}}^2 \nonumber\\
&\leq 3\E\norm{\frac{\eta}{M}\sum_{m=1}^M\sum_{k=0}^{K-1} \rvg_m^{(r,k)} - \frac{\eta}{M}\sum_{m=1}^M\sum_{k=0}^{K-1} \nabla F_m(\rvx_m^{(r,k)})}^2 \nonumber\\
&\quad+ 3\E\norm{\frac{\eta}{M}\sum_{m=1}^M\sum_{k=0}^{K-1} \nabla F_m(\rvx_m^{(r,k)}) - \frac{\eta}{M}\sum_{m=1}^M\sum_{k=0}^{K-1} \nabla F_m(\rvx^r)}^2 + 3\E\norm{\frac{\eta}{M}\sum_{m=1}^M\sum_{k=0}^{K-1} \nabla F_m(\rvx^r)}^2\label{eq:proof:lem:fedavg strongly convex:per-round recursion:3}
\end{align}
To bound the first term on the right hand side in \eqref{eq:proof:lem:fedavg strongly convex:per-round recursion:3}, we have,
\begin{align}
\E\norm{\frac{\eta}{M}\sum_{m=1}^M\sum_{k=0}^{K-1} \rvg_m^{(r,k)} - \frac{\eta}{M}\sum_{m=1}^M\sum_{k=0}^{K-1} \nabla F_m(\rvx_m^{(r,k)})}^2 &= \frac{\eta^2}{M^2}\sum_{m=1}^M\E\norm{\sum_{k=0}^{K-1} \rvg_m^{(r,k)} - \sum_{k=0}^{K-1} \nabla F_m(\rvx_m^{(r,k)})}^2 \nonumber\\
&\overset{\text{Lem.~}\ref{lem:martingale difference property}}{=} \frac{\eta^2}{M^2}\sum_{m=1}^M\sum_{k=0}^{K-1}\E\norm{ \rvg_m^{(r,k)} - \nabla F_m(\rvx_m^{(r,k)})}^2 \nonumber\\
&\overset{\eqref{eq:asm:stochasticity}}{\leq} \frac{K \eta^2\sigma^2}{M}\label{eq:the per-round recursion proof:3:1},
\end{align}
where we use the fact that clients are independent to each other in the first equality and Lemma~\ref{lem:martingale difference property} in the second equality. For the second term on the right hand side in \eqref{eq:proof:lem:fedavg strongly convex:per-round recursion:3}, using Jensen's inequality, we have
\begin{align}
\E\norm{\frac{\eta}{M}\sum_{m=1}^M\sum_{k=0}^{K-1} \nabla F_m(\rvx_m^{(r,k)}) - \frac{\eta}{M}\sum_{m=1}^M\sum_{k=0}^{K-1} \nabla F_m(\rvx^r)}^2
&\leq \frac{K\eta^2}{M}\sum_{m=1}^M\sum_{k=0}^{K-1}\E\norm{\nabla F_m(\rvx_m^{(r,k)}) - \nabla F_m(\rvx^r)}^2 \nonumber\\
&\overset{\text{Asm.~}\ref{asm:L-smoothness}}{\leq} \frac{L^2K\eta^2}{M}\sum_{m=1}^M\sum_{k=0}^{K-1}\E\norm{\rvx_m^{(r,k)} - \rvx^r}^2
\end{align}
For the third term on the right hand side in \eqref{eq:proof:lem:fedavg strongly convex:per-round recursion:3}, we have
\begin{align}
\E\norm{\frac{\eta}{M}\sum_{m=1}^M\sum_{k=0}^{K-1} \nabla F_m(\rvx^r)}^2
\leq K^2 \eta^2\E\norm{\nabla F(\rvx^r)}^2
\overset{\eqref{eq:L-smoothness grad bound}}{\leq} 2LK^2 \eta^2D_F(\rvx^r, \rvx^\ast)
\end{align}
Thus, substituting the preceding three inequalities into \eqref{eq:proof:lem:fedavg strongly convex:per-round recursion:3}, we have
\begin{align}
\E\norm{\frac{\eta}{M}\sum_{m=1}^M\sum_{k=0}^{K-1} \rvg_m^{(r,k)}}^2
\leq \frac{3K \eta^2\sigma^2}{M} +  \frac{3L^2K\eta^2}{M}\sum_{m=1}^M\sum_{k=0}^{K-1}\E\norm{\rvx_m^{(r,k)} - \rvx^r}^2 + 6LK^2 \eta^2D_F(\rvx^r, \rvx^\ast)\label{eq:proof:lem:fedavg strongly convex:per-round recursion:4}
\end{align}

So, substituting \eqref{eq:proof:lem:fedavg strongly convex:per-round recursion:2} and \eqref{eq:proof:lem:fedavg strongly convex:per-round recursion:4} into \eqref{eq:proof:lem:fedavg strongly convex:per-round recursion:1}, we have
\begin{align}
\E\norm{\rvx^{r+1}-\rvx^*}^2
&\leq \left(1-\tfrac{\mu K\eta}{2}\right)\E\norm{\rvx^{r}- \rvx^*}^2 - K\eta(2-6LK\eta)D_F(\rvx^r, \rvx^*) + \frac{3K \eta^2\sigma^2}{M}\nonumber\\
&\quad\quad + L\eta(2+3LK\eta) \frac{1}{M}\sum_{m=1}^M\sum_{k=0}^{K-1} \E\norm{\rvx_m^{(r,k)} - \rvx^r}^2
\end{align}
The claim of this lemma follows after taking unconditional expectations.
\end{proof}

\subsubsection{Bounding the client drift with \eqref{eq:asm:heterogeneity:optimum}}
\begin{lemma}\label{lem:fedavg strongly convex:per-round client drift}
Under the same assumptions as Theorem~\ref{thm:fedavg convergence}, for any learning rate satisfying $\eta \leq \frac{1}{6LK}$, the client drift $\gE_r$, defined as
\begin{align}
\gE_r \coloneqq \frac{1}{M}\sum_{m=1}^M\sum_{k=0}^{K-1} \E\norm{\rvx_m^{(r,k)} - \rvx^r}^2,\label{eq:def:fedavg strongly convex:per-round client drift}
\end{align}
is bounded, as given by:
\begin{align}
\E\left[\gE_r\right]
&\leq 3LK^3\eta^2\E \left[D_{F}(\rvx^{r}, \rvx^\ast)\right] + \frac{9}{4}K^2\eta^2\sigma^2 + \frac{3}{2}K^3\eta^2\zeta_\ast^2\label{eq:lem:fedavg strongly convex:per-round client drift}
\end{align}
\end{lemma}

\begin{proof}
Without otherwise stated, \textbf{the expectation is conditioned on} $\rvx^r$. Beginning with \(\E\norm{\rvx_m^{(r,k)} - \rvx^r}^2\). Considering \( \rvx_m^{(r,k)} = \rvx^{r} - \eta\sum_{j=0}^{k-1} \rvg_m^{(r,j)} \), we have
\begin{align}
\E\norm{\rvx_m^{(r,k)} - \rvx^r}^2 &= \eta^2\E \norm{\sum_{j=0}^{k-1} \rvg_m^{(r,j)}}^2
\end{align}
Then using the Jensen's inequality to the preceding equation, we have
\begin{align}
\E\norm{\rvx_m^{(r,k)} - \rvx^r}^2 &\leq 4\eta^2\E \norm{\sum_{j=0}^{k-1}\rvg_m^{(r,j)} - \sum_{j=0}^{k-1}\nabla F_m (\rvx_m^{(r,j)})}^2 + 4\eta^2\E\norm{\sum_{j=0}^{k-1}\nabla F_m (\rvx_m^{(r,j)}) - \sum_{j=0}^{k-1}\nabla F_m (\rvx^{r})}^2 \nonumber\\
&\quad+ 4\eta^2\E \norm{\sum_{j=0}^{k-1}\nabla F_m (\rvx^{r}) - \sum_{j=0}^{k-1}\nabla F_m (\rvx^{\ast})}^2 + 4\eta^2\E\norm{\sum_{j=0}^{k-1}\nabla F_m (\rvx^{\ast})}^2\label{eq:proof:lem:fedavg strongly convex:per-round client drift:1}
\end{align}
Applying Lemma~\ref{lem:martingale difference property} to the first term and Jensen’s inequality to the last three terms on the right hand side in \eqref{eq:proof:lem:fedavg strongly convex:per-round client drift:1} respectively, we get
\begin{align}
\E\norm{\rvx_m^{(r,k)} - \rvx^r}^2
&\leq 4\sum_{j=0}^{k-1}\eta^2\E\norm{\rvg_m^{(r,j)} - \nabla F_m (\rvx_m^{(r,j)})}^2 + 4k\sum_{j=0}^{k-1}\eta^2\E\norm{\nabla F_m (\rvx_m^{(r,j)}) - \nabla F_m (\rvx^{r})}^2 \nonumber\\
&\quad+ 4k^2\eta^2\E \norm{\nabla F_m (\rvx^{r})- \nabla F_m (\rvx^{\ast})}^2 + 4k^2\eta^2\E\norm{\nabla F_m (\rvx^{\ast})}^2\label{eq:proof:lem:fedavg strongly convex:per-round client drift:2}
\end{align}
For the first term on the right hand side in \eqref{eq:proof:lem:fedavg strongly convex:per-round client drift:2}, we have
\[
\E\norm{\rvg_m^{(r,j)} - \nabla F_m (\rvx_m^{(r,j)})}^2 \overset{\eqref{eq:asm:stochasticity}}{\leq} \sigma^2
\]
For the second term on the right hand side in \eqref{eq:proof:lem:fedavg strongly convex:per-round client drift:2}, we have
\[
\E\norm{\nabla F_m (\rvx_m^{(r,j)}) - \nabla F_m (\rvx^{r})}^2 \overset{\text{Asm.~}\ref{asm:L-smoothness}}{\leq} L^2 \E\norm{\rvx_m^{(r,j)} - \rvx^{r}}^2
\]
For the third term on the right hand side in \eqref{eq:proof:lem:fedavg strongly convex:per-round client drift:2}, we have
\begin{align}
\E \norm{\nabla F_m (\rvx^{r}) - \nabla F_m (\rvx^{\ast})}^2 \overset{\eqref{eq:bregman lower smooth}}{\leq} 2L D_{F_m}(\rvx^{r}, \rvx^\ast)\label{eq:proof:lem:fedavg strongly convex:per-round client drift:4}
\end{align}
So we get
\begin{align}
\E\norm{\rvx_m^{(r,k)} - \rvx^r}^2 &\leq 4k\eta^2\sigma^2 + 4L^2k\eta^2\sum_{j=0}^{k-1}\E\norm{\rvx_m^{(r,j)}-x^r}^2 \nonumber\\
&\quad+8Lk^2\eta^2D_{F_m}(\rvx^{r}, \rvx^\ast) + 4k^2\eta^2\norm{\nabla F_m (\rvx^{\ast})}^2\label{eq:proof:lem:fedavg strongly convex:per-round client drift:3}
\end{align}

Since $\E\norm{\rvx_m^{(r,k)} - \rvx^r}^2 = 0$ when $k=0$, now we have
\begin{align}
\E\left[\gE_r\right] = \frac{1}{M}\sum_{m=1}^M\sum_{k=0}^{K-1} \left[\E\norm{\rvx_m^{(r,k)} - \rvx^r}^2\right]
= \frac{1}{M}\sum_{m=1}^M\sum_{k=1}^{K-1} \left[\E\norm{\rvx_m^{(r,k)} - \rvx^r}^2\right]
\end{align}
Substituting \eqref{eq:proof:lem:fedavg strongly convex:per-round client drift:3} into the preceding equation, we have
\begin{align}
\E\left[\gE_r\right]
&\leq4L^2\eta^2\frac{1}{M}\sum_{m=1}^M\sum_{j=0}^{K-1}\E\norm{\rvx_m^{(r,j)}-x^r}^2\sum_{k=1}^{K-1}k \pfcomment{$0\leq j \leq k-1 \leq K-1$} \nonumber\\
&\quad+ 8L\eta^2D_{F}(\rvx^{r}, \rvx^\ast)\sum_{k=1}^{K-1}k^2 + 4\eta^2\sigma^2\sum_{k=1}^{K-1}k + 4\eta^2\zeta_\ast^2\sum_{k=1}^{K-1}k^2
\end{align}
Then using \( \sum_{k=1}^{K-1}k = \frac{(K-1)K}{2} \leq \frac{K^2}{2}\) for the first and third terms, $\sum_{k=1}^{K-1}k^2 = \frac{(K-1)K(2K-1)}{6} \leq \frac{K^3}{3}$ for the second and forth terms, we get
\begin{align}
\E\left[\gE_r\right]
&\leq 2L^2K^2\eta^2\frac{1}{M}\sum_{m=1}^M\sum_{j=0}^{K-1}\E\norm{\rvx_m^{(r,j)}-x^r}^2 + \frac{8}{3}LK^3\eta^2D_F(\rvx^r, \rvx^\ast) + 2K^2\eta^2\sigma^2 + \frac{4}{3}K^3\eta^2\zeta_\ast^2 \nonumber\\
&= 2L^2K^2\eta^2\E\left[\gE_r\right] + \frac{8}{3}LK^3\eta^2D_{F}(\rvx^{r}, \rvx^\ast) + 2K^2\eta^2\sigma^2 + \frac{4}{3}K^3\eta^2\zeta_\ast^2,
\end{align}
where we notice that $\displaystyle \frac{1}{M}\sum_{m=1}^M\sum_{j=0}^{K-1}\E\norm{\rvx_m^{(r,j)}-x^r}^2 = \E\left[\gE_r\right]$ in the first equality. Then we can rearrange the preceding inequality as follows:
\begin{align}
(1-2L^2K^2\eta^2)\E\left[\gE_r\right]
&\leq \frac{8}{3}LK^3\eta^2D_{F}(\rvx^{r}, \rvx^\ast) + 2K^2\eta^2\sigma^2 + \frac{4}{3}K^3\eta^2\zeta_\ast^2
\end{align}
Finally, using the choice of $\eta$, $LK\eta \leq \frac{1}{6}$, which implies $1-2L^2K^2\eta^2 \geq \frac{8}{9}$, we have
\begin{align}
\E\left[\gE_r\right]
&\leq 3LK^3\eta^2D_{F}(\rvx^{r}, \rvx^\ast) + \frac{9}{4}K^2\eta^2\sigma^2 + \frac{3}{2}K^3\eta^2\zeta_\ast^2.
\end{align}
The claim of this lemma follows after taking unconditional expectations.
\end{proof}

\subsubsection{Proof of strongly convex case of Theorem~\ref{thm:fedavg convergence}}
\begin{proof}[Proof of strongly convex case of Theorem~\ref{thm:fedavg convergence}]
Substituting the result of Lemma~\ref{lem:fedavg strongly convex:per-round client drift} (i.e., Inequality~\eqref{eq:lem:fedavg strongly convex:per-round client drift}) into the result of Lemma~\ref{lem:fedavg strongly convex:per-round recursion} (i.e., Inequality~\eqref{eq:lem:fedavg strongly convex:per-round recursion}) and using $LK\eta \leq \frac{1}{6}$, we can simplify the recursion as, 
\begin{align}
\E\norm{\rvx^{r+1}-\rvx^*}^2
&\leq \left(1-\tfrac{\mu K\eta}{2}\right)\E\norm{\rvx^{r}- \rvx^\ast}^2 - \frac{3}{4}K\eta\E\left[D_F(\rvx^r, \rvx^\ast)\right] + \frac{3K\eta^2\sigma^2}{M} \nonumber\\
&\quad+ \frac{45}{8}LK^2\eta^3\sigma^2 + \frac{15}{4}LK^3\eta^3\zeta_\ast^2
\end{align}
Let $\tilde{\eta} = K\eta$, we have
\begin{align}
\E\norm{\rvx^{r+1}-\rvx^*}^2 \leq \left(1-\frac{\mu \tilde{\eta}}{2}\right)\E\norm{\rvx^{r}- \rvx^\ast}^2 - \frac{3\tilde{\eta}}{4}\E\left[D_F(\rvx^r, \rvx^\ast)\right] + \frac{3\tilde{\eta}^2\sigma^2}{MK} + \frac{45L\tilde{\eta}^3\sigma^2}{8K} + \frac{15L \tilde{\eta}^3\zeta_\ast^2}{4}\label{eq:thm:proof:fedavg strongly convex:simplified per-round recursion:effective learning rate}
\end{align}

Applying Lemma~\ref{lem:stongly convex:tuning learning rate} with $t=r$ ($T=R$), $\gamma=\tilde\eta$, $r_{t} = \E\norm{\rvx^{r}- \rvx^*}^2$, $a = \frac{\mu}{2}$, $b=\frac{3}{4}$, $s_t = \E\left[D_F(\rvx^r, \rvx^\ast)\right] = \E\left[F(\rvx^{r}) - F(\rvx^*)\right]$, $w_t=(1-a\tilde\eta)^{-(r+1)}$, $c_1 = \frac{3\sigma^2}{MK}$, $c_2 = \frac{45L\sigma^2}{8K} + \frac{15L\zeta_\ast^2}{4}$ and $\frac{1}{d}=\frac{1}{6L}$ ($\tilde\eta=K\eta\leq\frac{1}{6L}$).

Note that in Lemma~\ref{lem:stongly convex:tuning learning rate}, there are no terms containing $\gamma^3$. As the terms containing $\gamma^3$ is not the determining factor for the convergence rate, Lemma~\ref{lem:stongly convex:tuning learning rate} can also be applied to this case \citep{karimireddy2020scaffold,koloskova2020unified}.
\begin{align}
F(\bar\rvx^R)-F(\rvx^\ast) \leq 2\mu\norm{\rvx^{0}- \rvx^*}^2 \exp\left[-\frac{1}{2}\mu\tilde\eta(R+1)\right] + \frac{4\tilde\eta\sigma^2}{MK} + \frac{15L\tilde\eta^2\sigma^2}{2K} + 5L\tilde\eta^2\zeta_\ast^2
\end{align}
where $\bar\rvx^R = \frac{1}{W_R}\sum_{r=0}^Rw_r\rvx^r$. Here we use
$F(\bar\rvx^R)-F(\rvx^\ast) \leq \frac{1}{W_R}\sum_{r=0}^Rw_r\E\left[F(\rvx^{r}) - F(\rvx^*)\right]$, 
since the convexity of $F$ and Jensen's inequality.

By tuning the learning rate carefully, we get
\begin{align}
F(\bar\rvx^R)-F(\rvx^\ast) = \tilde\gO\left(\mu D^2 \exp\left(-\frac{\mu}{12L}R\right) + \frac{\sigma^2}{\mu MKR} + \frac{L\sigma^2}{\mu^2KR^2} + \frac{L\zeta_\ast^2}{\mu^2 R^2}\right)
\end{align}
with $D=\norm{\rvx^{0}- \rvx^\ast}$.
\end{proof}

\subsubsection{Partial participation (Corollary~\ref{cor:without replacement:fedavg strongly convex} and Corollary~\ref{cor:with replacement:fedavg strongly convex})}
The proofs below are based on the work \cite{karimireddy2020scaffold, wang2020tackling}.

\begin{corollary}[Sampling without replacement]\label{cor:without replacement:fedavg strongly convex}
Under the same assumptions as Theorem~\ref{thm:fedavg convergence}, and assuming that at each round, the server randomly selects $S$ ($1\leq S\leq M$) clients without replacement to participate in training, we have
\begin{align}
F(\bar\rvx^R)-F(\rvx^\ast) \leq \frac{27}{8}\mu \norm{\rvx^{0}- \rvx^\ast}^2 \exp\left[-\frac{1}{2}\mu\tilde\eta(R+1)\right] + \left[\frac{4\sigma^2}{SK} + 4\zeta_\ast^2\frac{(M-S)}{S(M-1)}\right]\tilde{\eta} + \left[\frac{6L\sigma^2}{K} + 4L \zeta_\ast^2\right]\tilde{\eta}^2\label{eq:cor:without replacement:fedavg strongly convex:1}
\end{align}
After tuning the learning rate carefully, it follows that
\begin{align}
F(\bar\rvx^R)-F(\rvx^\ast) = \tilde\gO\left(\mu \norm{\rvx^{0}- \rvx^\ast}^2 \exp\left(-\frac{\mu}{12L}R\right) + \frac{\sigma^2}{\mu SKR} + \frac{\zeta_\ast^2}{\mu R}\frac{M-S}{S(M-1)} + \frac{L\sigma^2}{\mu^2KR^2} + \frac{L\zeta_\ast^2}{\mu^2 R^2}\right)\label{eq:cor:without replacement:fedavg strongly convex:2}
\end{align}
\end{corollary}
\begin{proof}
Since only a subset of clients participate the training, the update rule of FL is 
\[
\displaystyle \rvx^{r+1} = \rvx^r - \frac{\eta}{S}\sum_{m=1}^S\sum_{k=0}^{K-1} \rvg_m^{(r,k)}.
\]
Considering the sample mean is an unbiased estimator of the population mean (Lemma~\ref{lem:simple random sampling}), we have
\[
\E_{S,\xi}\left[\frac{1}{S}\sum_{m=1}^S\sum_{k=0}^{K-1} \rvg_m^{(r,k)}\right] = \E_\xi\left[\frac{1}{M}\sum_{m=1}^M\sum_{k=0}^{K-1} \rvg_m^{(r,k)}\right] = \E_\xi\left[\frac{1}{M}\sum_{m=1}^M\sum_{k=0}^{K-1} \nabla F_m(\rvx_m^{(r,k)})\right].
\]
The proof of this corollary is almost the same as the proof of Theorem~\ref{thm:fedavg convergence}, except the differences picked below. Without otherwise stated, the expectation is conditioned on $\rvx^r$.

\paragraph{Differences from Lemma~\ref{lem:fedavg strongly convex:per-round recursion}.}
\begin{align}
&\E\norm{\rvx^{r+1}-\rvx^\ast}^2 = \E\norm{\rvx^{r}- \frac{\eta}{S}\sum_{m=1}^S\sum_{k=0}^{K-1} \rvg_m^{(r,k)} - \rvx^*}^2 \nonumber\\
&= \E\norm{\rvx^{r}- \rvx^\ast}^2 - 2\E\inp{\rvx^r-\rvx^\ast}{\frac{\eta}{M}\sum_{m=1}^M\sum_{k=0}^{K-1} \nabla F_m(\rvx_m^{(r,k)})} + \E\norm{ \frac{\eta}{M}\sum_{m=1}^M\sum_{k=0}^{K-1} \rvg_m^{(r,k)}}^2\label{eq:proof:cor:without replacment:fedavg strongly convex:1},
\end{align}
where we use $\displaystyle \E_{S}\left[\frac{1}{S}\sum_{m=1}^S\sum_{k=0}^{K-1} \rvg_m^{(r,k)}\right] = \frac{1}{M}\sum_{m=1}^M\sum_{k=0}^{K-1} \rvg_m^{(r,k)}$ and $\E_{\xi}\left[\rvg_m^{(r,k)}\right] = \nabla F_m(\rvx_m^{(r,k)})$ for all $r, m, k$ in the second equality.

The second term on the right hand side in \eqref{eq:proof:cor:without replacment:fedavg strongly convex:1} by Lemma~\ref{lem:perturbed strong convexity} can be bounded as same as \eqref{eq:proof:lem:fedavg strongly convex:per-round recursion:2}.

For the third term on the right hand side in \eqref{eq:proof:cor:without replacment:fedavg strongly convex:1}, using Jensen's inequality, we have
\begin{align}
&\E\norm{\frac{\eta}{S}\sum_{m=1}^S\sum_{k=0}^{K-1} \rvg_m^{(r,k)}}^2 \nonumber\\
&\leq 4\E\norm{\frac{\eta}{S}\sum_{m=1}^S\sum_{k=0}^{K-1} \rvg_m^{(r,k)} - \frac{\eta}{S}\sum_{m=1}^S\sum_{k=0}^{K-1} \nabla F_m(\rvx_m^{(r,k)})}^2 + 4\E\norm{\frac{\eta}{S}\sum_{m=1}^S\sum_{k=0}^{K-1} \nabla F_m(\rvx_m^{(r,k)}) - \frac{\eta}{S}\sum_{m=1}^S\sum_{k=0}^{K-1} \nabla F_m(\rvx^r)}^2 \nonumber\\
&\quad+ 4\E\norm{\frac{\eta}{S}\sum_{m=1}^S\sum_{k=0}^{K-1} \nabla F_m(\rvx^r) - \frac{\eta}{S}\sum_{m=1}^S\sum_{k=0}^{K-1} \nabla F_m(\rvx^\ast)}^2 + 4\E\norm{\frac{\eta}{S}\sum_{m=1}^S\sum_{k=0}^{K-1} \nabla F_m(\rvx^\ast)}^2\label{eq:proof:cor:without replacment:fedavg strongly convex:2}
\end{align}
For the first term on the right hand side in \eqref{eq:proof:cor:without replacment:fedavg strongly convex:2}, we have
\begin{align}
\E\norm{\frac{\eta}{S}\sum_{m=1}^S\sum_{k=0}^{K-1} \rvg_m^{(r,k)} - \frac{\eta}{S}\sum_{m=1}^S\sum_{k=0}^{K-1} \nabla F_m(\rvx_m^{(r,k)})}^2 
&= \frac{\eta^2}{S}\E\left[\frac{1}{S}\sum_{m=1}^S\norm{\sum_{k=0}^{K-1} \rvg_m^{(r,k)} - \sum_{k=0}^{K-1} \nabla F_m(\rvx_m^{(r,k)})}^2\right] \nonumber\\
&= \frac{\eta^2}{SM}\sum_{m=1}^M\E\left[\norm{\sum_{k=0}^{K-1} \rvg_m^{(r,k)} - \sum_{k=0}^{K-1} \nabla F_m(\rvx_m^{(r,k)})}^2\right] \nonumber\\
&= \frac{\eta^2}{SM}\sum_{m=1}^M\sum_{k=0}^{K-1}\E\norm{ \rvg_m^{(r,k)} - \nabla F_m(\rvx_m^{(r,k)})}^2 \nonumber\\
&\overset{\eqref{eq:asm:stochasticity}}{\leq} \frac{K \eta^2\sigma^2}{S}\label{eq:proof:cor:without replacment:fedavg strongly convex:3}	
\end{align}
For the second term on the right hand side in \eqref{eq:proof:cor:without replacment:fedavg strongly convex:2}, we have
\begin{align}
\E\norm{\frac{\eta}{M}\sum_{m=1}^M\sum_{k=0}^{K-1} \nabla F_m(\rvx_m^{(r,k)}) - \frac{\eta}{M}\sum_{m=1}^M\sum_{k=0}^{K-1} \nabla F_m(\rvx^r)}^2
&\leq \eta^2\E\left[\frac{K}{S}\sum_{m=1}^S\sum_{k=0}^{K-1}\norm{\nabla F_m(\rvx_m^{(r,k)}) - \nabla F_m(\rvx^r)}^2\right] \nonumber\\
&\leq \frac{K\eta^2}{M}\sum_{m=1}^M\sum_{k=0}^{K-1}\E\norm{\nabla F_m(\rvx_m^{(r,k)}) - \nabla F_m(\rvx^r)}^2 \nonumber\\
&\leq \frac{L^2K\eta^2}{M}\sum_{m=1}^M\sum_{k=0}^{K-1}\E\norm{\rvx_m^{(r,k)} - \rvx^r}^2\label{eq:proof:cor:without replacment:fedavg strongly convex:4}
\end{align}
For the third term on the right hand side in \eqref{eq:proof:cor:without replacment:fedavg strongly convex:2}, we have
\begin{align}
\E\norm{\frac{\eta}{S}\sum_{m=1}^S\sum_{k=0}^{K-1} \nabla F_m(\rvx^r) - \frac{\eta}{S}\sum_{m=1}^S\sum_{k=0}^{K-1} \nabla F_m(\rvx^\ast)}^2
&\leq \eta^2\E\left[\frac{K}{S}\sum_{m=1}^S\sum_{k=0}^{K-1}\norm{\nabla F_m(\rvx^{r}) - \nabla F_m(\rvx^\ast)}^2\right] \nonumber\\
&\leq \frac{K\eta^2}{M}\sum_{m=1}^M\sum_{k=0}^{K-1}\E\norm{\nabla F_m(\rvx^{r}) - \nabla F_m(\rvx^\ast)}^2 \nonumber\\		
&\leq 2LK^2 \eta^2D_F(\rvx^r, \rvx^\ast)\label{eq:proof:cor:without replacment:fedavg strongly convex:5}
\end{align}
For the forth term on the right hand side in \eqref{eq:proof:cor:without replacment:fedavg strongly convex:2}, we have
\begin{align}
\E\norm{\frac{\eta}{S}\sum_{m=1}^S\sum_{k=0}^{K-1} \nabla F_m(\rvx^\ast)}^2 \leq K^2\eta^2\zeta_\ast^2\frac{(M-S)}{S(M-1)}\label{eq:proof:cor:without replacment:fedavg strongly convex:6}
\end{align}

Thus, substituting \eqref{eq:proof:cor:without replacment:fedavg strongly convex:3}, \eqref{eq:proof:cor:without replacment:fedavg strongly convex:4}, \eqref{eq:proof:cor:without replacment:fedavg strongly convex:5}, and \eqref{eq:proof:cor:without replacment:fedavg strongly convex:6} into \eqref{eq:proof:cor:without replacment:fedavg strongly convex:2}, we have
\begin{align}
\E\norm{\frac{\eta}{M}\sum_{m=1}^M\sum_{k=0}^{K-1} \rvg_m^{(r,k)}}^2
&\leq \frac{4K \eta^2\sigma^2}{S} + 4K^2\eta^2\zeta_\ast^2\frac{(M-S)}{S(M-1)}+ 8LK^2\eta^2D_F(\rvx^r, \rvx^\ast) \nonumber\\ 
&\quad+\frac{4L^2K\eta^2}{M}\sum_{m=1}^M\sum_{k=0}^{K-1}\E\norm{\rvx_m^{(r,k)} - \rvx^r}^2\label{eq:proof:cor:without replacment:fedavg strongly convex:7}
\end{align}

So, substituting \eqref{eq:proof:cor:without replacment:fedavg strongly convex:2} and \eqref{eq:proof:cor:without replacment:fedavg strongly convex:7} into \eqref{eq:proof:cor:without replacment:fedavg strongly convex:1}, we have
\begin{align}
\E\norm{\rvx^{r+1}-\rvx^*}^2
&\leq \left(1-\tfrac{\mu K\eta}{2}\right)\E\norm{\rvx^{r}- \rvx^*}^2 - K\eta(2-8LK\eta)D_F(\rvx^r, \rvx^*) + \frac{3K \eta^2\sigma^2}{M} \nonumber\\
&\quad\quad + 4K^2\eta^2\zeta_\ast^2\frac{(M-S)}{S(M-1)} + L\eta(2+3LK\eta) \frac{1}{M}\sum_{m=1}^M\sum_{k=0}^{K-1} \E\norm{\rvx_m^{(r,k)} - \rvx^r}^2\label{eq:proof:cor:without replacment:fedavg strongly convex:8}
\end{align}

\paragraph{Differences from Theorem~\ref{thm:fedavg convergence}.} Substituting the result of Lemma~\ref{lem:fedavg strongly convex:per-round recursion} (i.e., Inequality~\eqref{eq:lem:fedavg strongly convex:per-round recursion}) into \eqref{eq:proof:cor:without replacment:fedavg strongly convex:8} and using $LK\eta \leq \frac{1}{6}$, we can simplify the recursion as,
\begin{align}
\E\norm{\rvx^{r+1}-\rvx^*}^2
&\leq \left(1-\tfrac{\mu K\eta}{2}\right)\E\norm{\rvx^{r}- \rvx^\ast}^2 - \frac{4}{9}K\eta\E\left[D_F(\rvx^r, \rvx^\ast)\right] \nonumber\\
&\quad+ \frac{4K\eta^2\sigma^2}{S} + 6LK^2\eta^3\sigma^2 +4K^2\eta^2\zeta_\ast^2\frac{(M-S)}{S(M-1)} + 4LK^3\eta^3\zeta_\ast^2
\end{align}
Let $\tilde{\eta} = K\eta$, we have
\begin{align}
\E\norm{\rvx^{r+1}-\rvx^*}^2 \leq \left(1-\tfrac{\mu \tilde{\eta}}{2}\right)\E\norm{\rvx^{r}- \rvx^\ast}^2 - \frac{4\tilde{\eta}}{9}\E\left[D_F(\rvx^r, \rvx^\ast)\right] + \frac{4\tilde{\eta}^2\sigma^2}{SK} + \frac{6L\tilde{\eta}^3\sigma^2}{K} +4\tilde{\eta}^2\zeta_\ast^2\frac{(M-S)}{S(M-1)} + 4L \tilde{\eta}^3\zeta_\ast^2 \nonumber
\end{align}

Applying Lemma~\ref{lem:stongly convex:tuning learning rate} with $t=r$ ($T=R$), $\gamma=\tilde\eta$, $r_{t} = \E\norm{\rvx^{r}- \rvx^*}^2$, $a = \frac{\mu}{2}$, $b=\frac{4}{9}$, $s_t = \E\left[D_F(\rvx^r, \rvx^\ast)\right] = \E\left[F(\rvx^{r}) - F(\rvx^*)\right]$, $w_t=(1-a\tilde\eta)^{-(r+1)}$, $c_1 = \frac{4\sigma^2}{SK} + 4\zeta_\ast^2\frac{(M-S)}{S(M-1)}$, $c_2 = \frac{6L\sigma^2}{K} + 4L \zeta_\ast^2$ and $\frac{1}{d}=\frac{1}{6L}$ ($\tilde\eta=K\eta\leq\frac{1}{6L}$), we have
\begin{align}
F(\bar\rvx^R)-F(\rvx^\ast) \leq \frac{27}{8}\mu\norm{\rvx^{0}- \rvx^*}^2 \exp\left[-\frac{1}{2}\mu\tilde\eta(R+1)\right] + \left[\frac{4\sigma^2}{SK} + 4\zeta_\ast^2\frac{(M-S)}{S(M-1)}\right]\tilde{\eta} + \left[\frac{6L\sigma^2}{K} + 4L \zeta_\ast^2\right]\tilde{\eta}^2\nonumber
\end{align}
where $\bar\rvx^R = \frac{1}{W_R}\sum_{r=0}^Rw_r\rvx^r$. The preceding inequality is the first claim of Corollary~\ref{cor:without replacement:fedavg strongly convex}.

By tuning the learning rate carefully, we get the second claim of Corollary~\ref{cor:without replacement:fedavg strongly convex},
\begin{align}
F(\bar\rvx^R)-F(\rvx^\ast) = \tilde\gO\left(\mu D^2 \exp\left(-\frac{\mu}{12L}R\right) + \frac{\sigma^2}{\mu SKR}+ \frac{\zeta_\ast^2}{\mu R}\frac{M-S}{S(M-1)} + \frac{L\sigma^2}{\mu^2KR^2} + \frac{L\zeta_\ast^2}{\mu^2 R^2}\right)\nonumber
\end{align}
with $D=\norm{\rvx^{0}- \rvx^\ast}$.
When $M$ is large enough, we have $\frac{(M-S)}{S(M-1)} \approx (1-\frac{S}{M})\frac{1}{S}$. This is the constant appearing in \cite{karimireddy2020scaffold,woodworth2020minibatch}.
\end{proof}

\begin{corollary}[Sampling with replacement]\label{cor:with replacement:fedavg strongly convex}
Under the same assumptions as Theorem~\ref{thm:fedavg convergence}, and assuming that at each round, the server randomly selects $S$ ($1\leq S\leq M$) clients without replacement to participate in training, we have
\begin{align}
F(\bar\rvx^R)-F(\rvx^\ast) \leq \frac{27}{8}\mu\norm{\rvx^{0}- \rvx^*}^2 \exp\left[-\frac{1}{2}\mu\tilde\eta(R+1)\right] + \left[\frac{4\sigma^2}{SK} + \frac{4\zeta_\ast^2}{S}\right]\tilde{\eta} + \left[\frac{6L\sigma^2}{K} + 4L \zeta_\ast^2\right]\tilde{\eta}^2\label{eq:cor:with replacement:fedavg strongly convex:1}
\end{align}
After tuning the learning rate carefully, it follows that
\begin{align}
F(\bar\rvx^R)-F(\rvx^\ast) = \tilde\gO\left(\mu\norm{\rvx^{0}- \rvx^\ast}^2 \exp\left(-\frac{\mu}{12L}R\right) + \frac{\sigma^2}{\mu SKR}+ \frac{\zeta_\ast^2}{\mu SR} + \frac{L\sigma^2}{\mu^2KR^2} + \frac{L\zeta_\ast^2}{\mu^2 R^2}\right)\label{eq:cor:with replacement:fedavg strongly convex:2}
\end{align}
\end{corollary}
\begin{proof}
The proof of Corollary~\ref{cor:with replacement:fedavg strongly convex} is almost the same as the proof of Corollary~\ref{cor:without replacement:fedavg strongly convex} except the forth term on the right hand side in \eqref{eq:proof:cor:without replacment:fedavg strongly convex:2} (i.e., \eqref{eq:proof:cor:without replacment:fedavg strongly convex:6}).
\begin{align}
\E\norm{\frac{\eta}{S}\sum_{m=1}^S\sum_{k=0}^{K-1} \nabla F_m(\rvx^\ast)}^2 \leq \frac{K^2\eta^2\zeta_\ast^2}{S}
\end{align}
where we use \eqref{eq:lem:random sampling with replacement} of Lemma~\ref{lem:simple random sampling}.
\end{proof}

\subsection{General convex case}\label{subsec:general convex fedavg}

\subsubsection{Proof of general convex case of Theorem~\ref{thm:fedavg convergence}}
\begin{proof}[Proof of general convex case of Theorem~\ref{thm:fedavg convergence}]
Let $\mu=0$ in \eqref{eq:thm:proof:fedavg strongly convex:simplified per-round recursion:effective learning rate}, we get the simplified per-round recursion of general convex case,
\begin{align}
	\E\norm{\rvx^{r+1}-\rvx^*}^2 \leq \E\norm{\rvx^{r}- \rvx^\ast}^2 - \frac{3\tilde{\eta}}{4}\E\left[D_F(\rvx^r, \rvx^\ast)\right] + \frac{3\tilde{\eta}^2\sigma^2}{MK} + \frac{45L\tilde{\eta}^3\sigma^2}{8K} + \frac{15L \tilde{\eta}^3\zeta_\ast^2}{4}
\end{align}
Then applying Lemma~\ref{lem:general convex:tuning learning rate} with $t=r$ ($T=R$), $\gamma=\tilde\eta$, $r_{t} = \E\norm{\rvx^{r}- \rvx^*}^2$, $b=\frac{3}{4}$, $s_t = \E\left[D_F(\rvx^r, \rvx^\ast)\right] = \E\left[F(\rvx^{r}) - F(\rvx^*)\right]$, $w_t=1$, $c_1 = \frac{3\sigma^2}{MK}$, $c_2 = \frac{45L\sigma^2}{8K} + \frac{15L\zeta_\ast^2}{4}$ and $\frac{1}{d}=\frac{1}{6L}$ ($\tilde\eta=K\eta\leq\frac{1}{6L}$), we can have
\begin{align}
	F(\bar\rvx^R)-F(\rvx^\ast) \leq \frac{4\norm{\rvx^{0}- \rvx^\ast}^2}{3\tilde\eta R} + \frac{4\tilde\eta\sigma^2}{MK} + \frac{15L\tilde\eta^2\sigma^2}{2K} + 5L\tilde\eta^2\zeta_\ast^2
\end{align}
where $\bar\rvx^R = \frac{1}{W_R}\sum_{r=0}^Rw_r\rvx^r$. Here we have
$F(\bar\rvx^R)-F(\rvx^\ast) \leq \frac{1}{W_R}\sum_{r=0}^Rw_r\E\left[F(\rvx^{r}) - F(\rvx^*)\right]$,
with the convexity of $F$ and Jensen's inequality. By tuning the learning rate carefully, we get
\begin{align}
	F(\bar\rvx^R)-F(\rvx^\ast) = \gO\left(\frac{\sigma D}{\sqrt{MKR}} + \frac{\left(L\sigma^2D^4\right)^{1/3}}{K^{1/3}R^{2/3}} + \frac{\left(L\zeta_\ast^2D^4\right)^{1/3}}{R^{2/3}} + \frac{LD^2}{R}\right),
\end{align}
where $D \coloneqq \norm{\rvx^{0}- \rvx^\ast}$.
\end{proof}

\subsection{Non-convex case}\label{subsec:non-convex fedavg}
\begin{lemma}\label{lem:fedavg non-convex:per-round recursion}
	Under the same assumptions as Theorem~\ref{thm:fedavg convergence}, we can find the per-round recursion as
	\begin{align}
		\E [F(\rvx^{r+1})] - F(\rvx^{r}) \leq -\frac{K\eta}{2}\norm{\nabla F(\rvx^{r})}^2 + \frac{LK\eta^2\sigma^2}{M} + \frac{L^2\eta}{2M}\sum_{m=1}^M\sum_{k=0}^{K-1}\E\norm{\rvx_m^{(r,k)}-\rvx^r}^2\label{eq:lem:fedavg non-convex:per-round recursion}
	\end{align}
\end{lemma}
\begin{proof}
	Without otherwise stated, \textbf{the expectation is conditioned on} $\rvx^r$. Beginning from $L$-smoothness, 
	\[
	D_F(\rvx^{r+1}, \rvx^r) = F(\rvx^{r+1}) - F(\rvx^r) - \inp{\nabla F(\rvx^r)}{\rvx^{r+1}-\rvx^r} \leq \frac{L}{2}\norm{\rvx^{r+1} - \rvx^r}^2,
	\] and the update rule of FL, 
	\[
	\displaystyle \rvx^{r+1} = \rvx^r - \frac{\eta}{M}\sum_{m=1}^M\sum_{k=0}^{K-1} \rvg_m^{(r,k)},
	\]
	where $\rvg_m^{(r,k)} = \nabla f_m(\rvx_m^{(r,k)}; \xi_m^{(r,k)})$, we have
	\begin{align}
		&\E [F(\rvx^{r+1})] - F(\rvx^{r})\leq - \eta \sum_{k=0}^{K-1}\E\inp{\nabla F(\rvx^{r})}{\frac{1}{M}\sum_{m=1}^M\nabla F_m(\rvx_m^{(r,k)})} + \frac{L}{2}\E\norm{ \frac{\eta}{M}\sum_{m=1}^{M}\sum_{k=0}^{K-1}\rvg_m^{r,k} }^2\label{eq:proof:lem:fedavg non-convex:per-round recursion:1}
	\end{align}
	For the first term on the right hand side in \eqref{eq:proof:lem:fedavg non-convex:per-round recursion:1}, using $\inp{a}{b} = \frac{1}{2}\norm{a}^2 + \frac{1}{2}\norm{b}^2 - \frac{1}{2}\norm{a-b}^2$ with $a = \nabla F(\rvx^{r})$ and $b = \frac{1}{M}\sum_{m=1}^M\nabla F_m(\rvx_m^{(r,k)})$, we have
	\begin{align}
		&- \eta \sum_{k=0}^{K-1}\E\inp{\nabla F(\rvx^{r})}{\frac{1}{M}\sum_{m=1}^M\nabla F_m(\rvx_m^{(r,k)})} \nonumber\\
		&\leq - \frac{\eta}{2} \sum_{k=0}^{K-1}\E\left[\norm{\nabla F(\rvx^{r})}^2 + \norm{\frac{1}{M}\sum_{m=1}^M\nabla F_m(\rvx_m^{(r,k)})}^2 - \norm{\frac{1}{M}\sum_{m=1}^M\nabla F_m(\rvx_m^{(r,k)}) - \nabla F(\rvx^{r})}^2\right]\\
		&\overset{\eqref{eq:jensen norm},\text{Asm.~}\ref{asm:L-smoothness}}{\leq} -\frac{\eta}{2}\sum_{k=0}^{K-1}\norm{\nabla F(\rvx^{r})}^2 - \frac{\eta}{2}\sum_{k=0}^{K-1}\E\norm{\frac{1}{M}\sum_{m=1}^M\nabla F_m(\rvx_m^{(r,k)})}^2 + \frac{L^2\eta}{2M}\sum_{m=1}^M\sum_{k=0}^{K-1}\E\norm{\rvx_m^{(r,k)}-\rvx^r}^2\label{eq:proof:lem:fedavg non-convex:per-round recursion:2}
	\end{align}
	For the second term on the right hand side in \eqref{eq:proof:lem:fedavg non-convex:per-round recursion:1}, we have
	\begin{align}
		\frac{L}{2}\E\norm{ \frac{\eta}{M}\sum_{m=1}^{M}\sum_{k=0}^{K-1}\rvg_m^{r,k} }^2 &\leq \frac{L\eta^2}{M^2}\E\norm{\sum_{m=1}^{M}\sum_{k=0}^{K-1}\rvg_m^{r,k}-\sum_{m=1}^{M}\sum_{k=0}^{K-1} \nabla F_m(\rvx_m^{(r,k)})}^2 + \frac{L\eta^2}{M^2}\E\norm{\sum_{m=1}^{M}\sum_{k=0}^{K-1} \nabla F_m(\rvx_m^{(r,k)})}^2\nonumber\\
		&\leq \frac{LK\eta^2\sigma^2}{M} + \frac{L\eta^2}{M^2}\E\norm{\sum_{m=1}^{M}\sum_{k=0}^{K-1} \nabla F_m(\rvx_m^{(r,k)})}^2\label{eq:proof:lem:fedavg non-convex:per-round recursion:3}
	\end{align}
	
	Substituting \eqref{eq:proof:lem:fedavg non-convex:per-round recursion:2} and \eqref{eq:proof:lem:fedavg non-convex:per-round recursion:3} into \eqref{eq:proof:lem:fedavg non-convex:per-round recursion:1}, we have
	\begin{align}
		\E [F(\rvx^{r+1})] - F(\rvx^{r}) &\overset{\eqref{eq:jensen norm}}{\leq} -\frac{K\eta}{2}\norm{\nabla F(\rvx^{r})}^2 + \frac{L^2\eta}{2M}\sum_{m=1}^M\sum_{k=0}^{K-1}\E\norm{\rvx_m^{(r,k)}-\rvx^r}^2 
		+\frac{LK\eta^2\sigma^2}{M} \nonumber\\
		&\quad-\frac{\eta}{2M}(1-2LK\eta)\frac{1}{M}\sum_{m=1}^M\sum_{k=0}^{K-1} \norm{\nabla F_m(\rvx_m^{(r,k)})}^2
	\end{align}
	Considering $K\eta\leq \frac{1}{6L}$, the last term on the right hand side in the preceding inequality is negative. As a consequence, we have
	\begin{align*}
		\E [F(\rvx^{r+1})] - F(\rvx^{r}) &\overset{\eqref{eq:jensen norm}}{\leq} -\frac{K\eta}{2}\norm{\nabla F(\rvx^{r})}^2 + \frac{LK\eta^2\sigma^2}{M} + \frac{L^2\eta}{2M}\sum_{m=1}^M\sum_{k=0}^{K-1}\E\norm{\rvx_m^{(r,k)}-\rvx^r}^2 
	\end{align*}
\end{proof}

\subsubsection{Bounding the client drift with \eqref{eq:asm:heterogeneity:everywhere}}

\begin{lemma}\label{lem:fedavg non-convex:per-round client drift}
	Under the same assumptions as Theorem~\ref{thm:fedavg convergence}, for any learning rate satisfying $\eta \leq \frac{1}{6LK}$, the client drift $\gE_r$, defined as
	\begin{align*}
		\gE_r \coloneqq \frac{1}{M}\sum_{m=1}^M\sum_{k=0}^{K-1} \E\norm{\rvx_m^{(r,k)} - \rvx^r}^2,
	\end{align*}
	is bounded, as given by:
	\begin{align}
		\E\left[\gE_r\right]
		&\leq \frac{3}{2}K^3\eta^2(B^2+1)\norm{\nabla F (\rvx^{r})}^2 + \frac{9}{4}K^2\eta^2\sigma^2 + \frac{3}{2}K^3\eta^2\zeta^2\label{eq:lem:fedavg non-convex:per-round client drift}
	\end{align}
\end{lemma}
\begin{proof}
	Without otherwise stated, \textbf{the expectation is conditioned on} $\rvx^r$. Beginning with \(\E\norm{\rvx_m^{(r,k)} - \rvx^r}^2\). Considering \( \rvx_m^{(r,k)} = \rvx^{r} - \eta\sum_{j=0}^{k-1} \rvg_m^{(r,j)} \), we have
	\begin{align}
		\E\norm{\rvx_m^{(r,k)} - \rvx^r}^2 &= \eta^2\E \norm{\sum_{j=0}^{k-1} \rvg_m^{(r,j)}}^2
	\end{align}
	Then using the Jensen's inequality to the preceding equation, we have
	\begin{align}
		\E\norm{\rvx_m^{(r,k)} - \rvx^r}^2 &\leq 4\eta^2\E \norm{\sum_{j=0}^{k-1}\rvg_m^{(r,j)} - \sum_{j=0}^{k-1}\nabla F_m (\rvx_m^{(r,j)})}^2 + 4\eta^2\E\norm{\sum_{j=0}^{k-1}\nabla F_m (\rvx_m^{(r,j)}) - \sum_{j=0}^{k-1}\nabla F_m (\rvx^{r})}^2 \nonumber\\
		&\quad+ 4\eta^2\E \norm{\sum_{j=0}^{k-1}\nabla F_m (\rvx^{r}) - \sum_{j=0}^{k-1}\nabla F (\rvx^{r})}^2 + 4\eta^2\E\norm{\sum_{j=0}^{k-1}\nabla F (\rvx^{r})}^2\label{eq:proof:lem:fedavg non-convex:per-round client drift:1}
	\end{align}
	Applying Lemma~\ref{lem:martingale difference property} to the first term and Jensen’s inequality to the last three terms on the right hand side in \eqref{eq:proof:lem:fedavg strongly convex:per-round client drift:1} respectively, we get
	\begin{align}
		\E\norm{\rvx_m^{(r,k)} - \rvx^r}^2
		&\leq 4\sum_{j=0}^{k-1}\eta^2\E\norm{\rvg_m^{(r,j)} - \nabla F_m (\rvx_m^{(r,j)})}^2 + 4k\sum_{j=0}^{k-1}\eta^2\E\norm{\nabla F_m (\rvx_m^{(r,j)}) - \nabla F_m (\rvx^{r})}^2 \nonumber\\
		&\quad+ 4k^2\eta^2\E \norm{\nabla F_m (\rvx^{r})- \nabla F (\rvx^{r})}^2 + 4k^2\eta^2\E\norm{\nabla F (\rvx^{r})}^2\label{eq:proof:lem:fedavg non-convex:per-round client drift:2}
	\end{align}
	For the first term on the right hand side in \eqref{eq:proof:lem:fedavg non-convex:per-round client drift:2}, we have
	\[
	\E\norm{\rvg_m^{(r,j)} - \nabla F_m (\rvx_m^{(r,j)})}^2 \overset{\eqref{eq:asm:stochasticity}}{\leq} \sigma^2
	\]
	For the second term on the right hand side in \eqref{eq:proof:lem:fedavg non-convex:per-round client drift:2}, we have
	\[
	\E\norm{\nabla F_m (\rvx_m^{(r,j)}) - \nabla F_m (\rvx^{r})}^2 \overset{\text{Asm.~}\ref{asm:L-smoothness}}{\leq} L^2 \E\norm{\rvx_m^{(r,j)} - \rvx^{r}}^2
	\]
	So we get
	\begin{align}
		\E\norm{\rvx_m^{(r,k)} - \rvx^r}^2 &\leq 4k\eta^2\sigma^2 + 4L^2k\eta^2\sum_{j=0}^{k-1}\E\norm{\rvx_m^{(r,j)}-x^r}^2 \nonumber\\
		&\quad+ 4k^2\eta^2\E \norm{\nabla F_m (\rvx^{r})- \nabla F (\rvx^{r})}^2 + 4k^2\eta^2\norm{\nabla F (\rvx^{r})}^2\label{eq:proof:lem:fedavg non-convex:per-round client drift:3}
	\end{align}
	
	Since $\E\norm{\rvx_m^{(r,k)} - \rvx^r}^2 = 0$ when $k=0$, now we have
	\begin{align}
		\E\left[\gE_r\right] = \frac{1}{M}\sum_{m=1}^M\sum_{k=0}^{K-1} \left[\E\norm{\rvx_m^{(r,k)} - \rvx^r}^2\right]
		= \frac{1}{M}\sum_{m=1}^M\sum_{k=1}^{K-1} \left[\E\norm{\rvx_m^{(r,k)} - \rvx^r}^2\right]
	\end{align}
	Substituting \eqref{eq:proof:lem:fedavg non-convex:per-round client drift:3} into the preceding equation, we have
	\begin{align}
		\E\left[\gE_r\right]
		&\leq4L^2\eta^2\frac{1}{M}\sum_{m=1}^M\sum_{j=0}^{K-1}\E\norm{\rvx_m^{(r,j)}-x^r}^2\sum_{k=1}^{K-1}k \pfcomment{$0\leq j \leq k-1 \leq K-1$} \nonumber\\
		&\quad+ 4\eta^2\frac{1}{M}\sum_{m=1}^M\E\norm{\nabla F_m (\rvx^{r})- \nabla F (\rvx^{r})}^2\sum_{k=1}^{K-1}k^2 + 4\eta^2\sigma^2\sum_{k=1}^{K-1}k + 4\eta^2\norm{\nabla F (\rvx^{r})}^2\sum_{k=1}^{K-1}k^2
	\end{align}
	Then using \( \sum_{k=1}^{K-1}k = \frac{(K-1)K}{2} \leq \frac{K^2}{2}\) for the first and third terms, $\sum_{k=1}^{K-1}k^2 = \frac{(K-1)K(2K-1)}{6} \leq \frac{K^3}{3}$ for the second and forth terms, we get
	\begin{align}
		\E\left[\gE_r\right]
		&\leq 2L^2K^2\eta^2\frac{1}{M}\sum_{m=1}^M\sum_{j=0}^{K-1}\E\norm{\rvx_m^{(r,j)}-x^r}^2 + \frac{4}{3}K^3\eta^2\gA + 2K^2\eta^2\sigma^2 + \frac{4}{3}K^3\eta^2\norm{\nabla F (\rvx^{r})}^2 \nonumber\\
		&= 2L^2K^2\eta^2\E\left[\gE_r\right] + \frac{4}{3}K^3\eta^2\gA + 2K^2\eta^2\sigma^2 + \frac{4}{3}K^3\eta^2\norm{\nabla F (\rvx^{r})}^2,
	\end{align}
	where we notice that $\displaystyle \frac{1}{M}\sum_{m=1}^M\sum_{j=0}^{K-1}\E\norm{\rvx_m^{(r,j)}-x^r}^2 = \E\left[\gE_r\right]$ in the first equality and use $\displaystyle \gA = \frac{1}{M}\sum_{m=1}^M\E\norm{\nabla F_m (\rvx^{r})- \nabla F (\rvx^{r})}^2$ for simplicity. Then we rearrange the preceding inequality as:
	\begin{align}
		(1-2L^2K^2\eta^2)\E\left[\gE_r\right]
		&\leq \frac{4}{3}K^3\eta^2\gA + 2K^2\eta^2\sigma^2 + \frac{4}{3}K^3\eta^2\norm{\nabla F (\rvx^{r})}^2
	\end{align}
	Using the choice of $\eta$, $LK\eta \leq \frac{1}{6}$, which implies $1-2L^2K^2\eta^2 \geq \frac{8}{9}$, we have
	\begin{align}
		\E\left[\gE_r\right]
		&\leq \frac{3}{2}K^3\eta^2\gA + \frac{9}{4}K^2\eta^2\sigma^2 + \frac{3}{2}K^3\eta^2\norm{\nabla F (\rvx^{r})}^2
	\end{align}
	Finally, by \eqref{eq:asm:heterogeneity:everywhere} in Assumption~\ref{asm:heterogeneity}, we have
	\begin{align}
		\E\left[\gE_r\right]
		&\leq \frac{3}{2}K^3\eta^2(B^2+1)\norm{\nabla F (\rvx^{r})}^2 + \frac{9}{4}K^2\eta^2\sigma^2 + \frac{3}{2}K^3\eta^2\zeta^2
	\end{align}
	The claim of this lemma follows after taking unconditional expectations.
\end{proof}

\subsubsection{Proof of non-convex case of Theorem~\ref{thm:fedavg convergence}}

\begin{proof}[Proof of non-convex case of Theorem~\ref{thm:fedavg convergence}]
	Substituting \eqref{eq:lem:fedavg non-convex:per-round client drift} (Lemma~\ref{lem:fedavg non-convex:per-round client drift}) into \eqref{eq:lem:fedavg non-convex:per-round recursion} (Lemma~\ref{lem:fedavg non-convex:per-round recursion}), and using $K\eta \leq \frac{1}{6L(B^2+1)}$we have
	\begin{align}
		\E [F(\rvx^{r+1})] - F(\rvx^{r}) 
		&\leq -K\eta\left[\frac{1}{2}-\frac{3}{4}L^2K^2\eta^2(B^2+1)\right]\E\norm{\nabla F(\rvx^{r})}^2 + \frac{LK\eta^2\sigma^2}{M} + \frac{9}{8}L^2K^2\eta^3\sigma^2 + \frac{3}{4}L^2K^3\eta^3\zeta^2 \nonumber\\
		&\leq -\frac{K\eta}{3}\E\norm{\nabla F(\rvx)}^2 + \frac{LK\eta^2\sigma^2}{M} + \frac{9}{8}L^2K^2\eta^3\sigma^2 + \frac{3}{4}L^2K^3\eta^3\zeta^2
	\end{align}
	Letting $\tilde \eta \coloneqq K\eta$, we have
	\begin{align}
		\E [F(\rvx^{r+1})-F^\ast] \leq F(\rvx^{r}) - F^\ast - \frac{\tilde\eta}{3}\E\norm{\nabla F(\rvx)}^2 + \frac{L\tilde\eta^2\sigma^2}{MK} + \frac{9L^2\tilde\eta^3\sigma^2}{8K} + \frac{3}{4}L^2\tilde\eta^3\zeta^2
	\end{align}
	Then applying Lemma~\ref{lem:general convex:tuning learning rate} with $t=r$ ($T=R$), $\gamma=\tilde\eta$, $r_{t} = \E[F(\rvx^r)-F(\rvx^\ast)]$, $b=\frac{1}{3}$, $s_t = \E\norm{\nabla F(\rvx)}^2$, $w_t=1$, $c_1 = \frac{L\sigma^2}{MK}$, $c_2 = \frac{9L^2\sigma^2}{8K} + \frac{3}{4}L^2\zeta^2$ and $\frac{1}{d}=\frac{1}{6L(B^2+1)}$ ($\tilde\eta=K\eta\leq\frac{1}{6L(B^2+1)}$), we have
	\begin{align}
		\frac{1}{R+1}\sum_{r=0}^{R}\E\norm{\nabla F(\rvx)}^2 \leq \frac{3\E [F(\rvx^{0}) - F^\ast]}{\tilde\eta R} + \frac{3L\tilde\eta\sigma^2}{MK} + \frac{27L^2\tilde\eta^2\sigma^2}{8K} + \frac{9}{4}L^2\tilde\eta^2\zeta^2
	\end{align}
	By tuning the learning rate carefully, we get
	\begin{align}
		\min_{0\leq r\leq R}\E\norm{\nabla F(\rvx)}^2 = \gO\left(\frac{ \left(\sigma^2LA\right)^{1/2}}{\sqrt{MKR}} + \frac{\left(L^2\sigma^2A^2\right)^{1/3}}{K^{1/3}R^{2/3}} + \frac{\left(L^2\zeta^2A^2\right)^{1/3}}{R^{2/3}} + \frac{LB^2A}{R}\right),
	\end{align}
	where $A \coloneqq F(\rvx^{0})- F^\ast$ and $\min_{0\leq r\leq R}\E\norm{\nabla F(\rvx)}^2 \leq  \frac{1}{R+1}\sum_{r=0}^{R}\E\norm{\nabla F(\rvx)}^2$.
\end{proof}
\section{Details about \texttt{SSL}}\label{sec:algs}
\begin{figure}[htbp]
	\centering
	\includegraphics[width=0.6\linewidth]{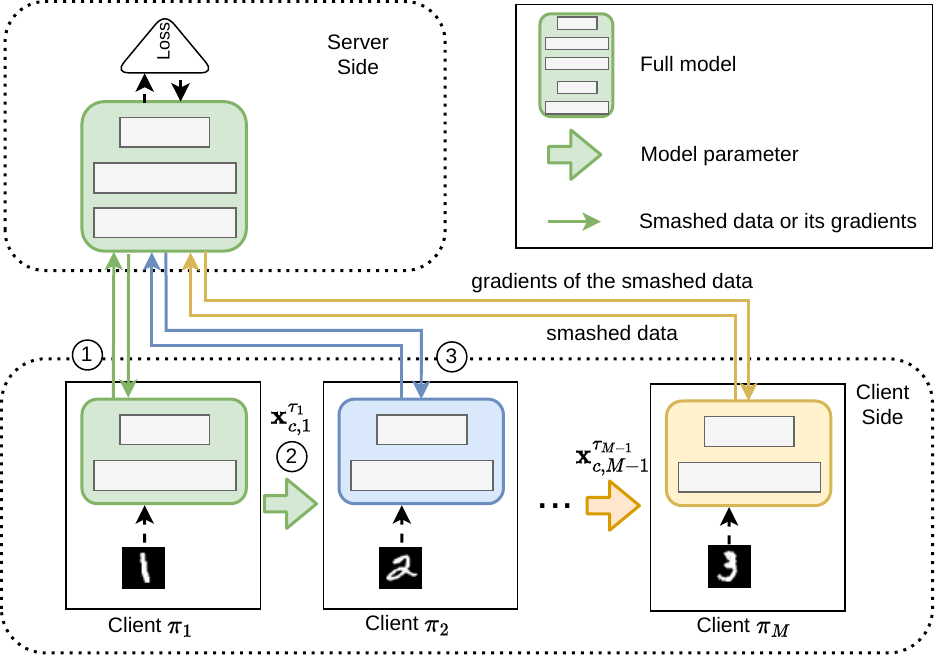}
	\caption{Overview of \texttt{SSL} in the peer-to-peer mode. The first three steps of the workflow: \ding{172} local updates with client $\pi_1$ (lines 8-15 of Alg.~\ref{alg:detailed SSL}, lines 4-10 of Alg.~\ref{alg:detailed SSL 2}); \ding{173} client-side model synchronization between clients $\pi_1$ and $\pi_2$, i.e., client $\pi_1$ sends the client-side local model parameters to client $\pi_2$; \ding{174} local updates with client~$\pi_2$.}
	\label{fig:overview of SSL}
\end{figure}
In SL, the full model is split to be trained at the clients and server, where the portion at the client side is called \textit{the client-side model} and the portion at the server is called  \textit{the server-side model}. In this section, we provide more operation details about \texttt{SSL}. All settings are the same as that in the main body except that varying number of local steps per client is adopted.

\paragraph{Operation details of \texttt{SSL}.} Figure~\ref{fig:overview of SSL} provides a overview of \texttt{SSL}. At the beginning of each training round, the indices $\pi_1, \pi_2, \ldots, \pi_M$ are sampled without replacement from $\{1,2,\ldots,M\}$ randomly as clients' training order. The training process for any client $\tau_m$ can be divided into three parts.

\textit{Initialization.} In each round, client $\tau_m$ and the server initialize their model parameters with the latest model parameters (line 3 of Alg.~\ref{alg:detailed SSL 2}, line 6 of Alg.~\ref{alg:detailed SSL}). Note that the server-side model is latest in practice, we specify here for clarity (to be consistent with the client side). 

\textit{Local updates.} The client and server perform multiple steps of local updates collaboratively (lines 4-10 of Alg.~\ref{alg:detailed SSL 2}, lines 8-15 of Alg.~\ref{alg:detailed SSL}). The local update process in \texttt{SSL} includes: (i) the client executes the forward pass with its local data samples. It sends the smashed data and labels to the server, where the smashed data is the activation of the \textit{cut layer} (the last layer of the client-side model). (ii) After receiving the smashed data and labels, the server executes the forward pass on the server-side model with the smashed data as the input, computes the loss with the output of the server-side model and labels, executes the backward pass, sends the gradients of the smashed data to the client and updates its server-side model parameters. (iii) After receiving the gradients, the client continues to execute the backward pass and updates its client-side model parameters.

\textit{Client-side model synchronization.} Client-side model parameters can be synchronized in two modes, the \textit{peer-to-peer mode} and \textit{centralized mode} \citep{gupta2018distributed}. In the peer-to-peer mode, the updated client-side model parameters will be sent to the next client directly, as shown in Figure~\ref{fig:overview of SSL}. In the centralized mode, the updated parameters will be sent to the next client with the server as a medium. We mark the differences of these two modes with pink and gray color box in Algs.~\ref{alg:detailed SSL}, \ref{alg:detailed SSL 2}.

This process continues until all the clients finish their local updates. Details are given in Algorithms~\ref{alg:detailed SSL}, \ref{alg:detailed SSL 2}. Additional notations required for operations of \texttt{SSL} are given in Table~\ref{table:apx:additional notations}.


%

\begin{table}[ht]
	\renewcommand{\arraystretch}{1.2}
	\centering
	\caption{Additional notations for Section~\ref{sec:algs}.}
	\label{table:apx:additional notations}
	\begin{tabular}{cl}
		\toprule
		Symbol &Description\\ \midrule
		$\tau_m, k$ &number, index of local update steps (when training) with client $\pi_m$\\
		$\rvx_{m}^{(r,k)}$/$\rvx_{c,m}^{(r,k)}$/$\rvx_{s,m}^{(r,k)}$ &\begin{tabular}[c]{@{}l@{}}full/client-side/server-side local model parameters ($\rvx_{m}^{(r,k)} = [\rvx_{c,m}^{(r,k)};\rvx_{s,m}^{(r,k)}]$)\\ \quad after $k$ local updates with client $\pi_m$ in the $r$-th round \\
		\end{tabular} \\
		$\mX_m^{(r,k)}$/$\mY_m^{(r,k)}$/$\hat\mY_m^{(r,k)}$ &\begin{tabular}[c]{@{}l@{}}features/labels/predictors \\
			\quad after $k$ local updates with client $\pi_m$ in the $r$-th round\end{tabular} \\
		$\rvx^{r}$/$\rvx_{c}^{r}$/$\rvx_{s}^{r}$ &full/client-side/server-side global model parameters in the $r$-th round \\
		$\mA_m^{(r,k)}$ &\begin{tabular}[c]{@{}l@{}}smashed data (activation of the cut layer) \\
			\quad after $k$ local updates with client $\pi_m$ in the $r$-th round \end{tabular}\\
		$\ell_{\pi_m}$ &loss function with client $\pi_m$\\
		$\nabla \ell_{\pi_m} (\rvx_{s,m}^{(r,k)}; \mA_m^{(r,k)})$ &gradients of the loss regarding $\rvx_{s,m}^{(r,k)}$ on input $\mA_m^{(r,k)}$\\
		$\nabla \ell_{\pi_m} (\mA_m^{(r,k)};\rvx_{s,m}^{(r,k)})$ &gradients of the loss regarding $\mA_m^{(r,k)}$ on parameters $\rvx_{s,m}^{(r,k)}$ \\	
		$\nabla \ell_{\pi_m} (\rvx_{c,m}^{(r,k)}; \mX_m^{(r,k)})$ &gradients of the loss regarding $\rvx_{c,m}^{(r,k)}$ on input $\mX_m^{(r,k)}$\\
		\bottomrule
	\end{tabular}
\end{table}

\begin{algorithm}[ht]
	\caption{\texttt{SSL} (Server-side operations) \colorbox{gray!50}{Peer-to-peer mode}, \colorbox{pink!50}{Centralized mode}}\label{alg:detailed SSL}
	\begin{algorithmic}[1]
		\Ensure server-side global model parameters $\rvx_s^{r}= \rvx_{s,M}^{(r-1,\tau_M)}$.
		\Procedure{\textbf{Main Server} executes}{}
		\State Initialize server-side global parameters $\rvx_s^{0}$
		\For{round $r = 0,\ldots, R-1$}
		\State Sample a permutation $\pi_1, \pi_2, \ldots, \pi_{M}$ of $\{1,2,\ldots,M\}$ as clients' update order
		\For{$m = 1,2,\ldots,M$ \textbf{\textcolor{winered}{in sequence}}}
		\State Initialize server-side local parameters $\rvx_{s,m}^{(r,0)} \gets
		\begin{cases}
			\rvx_s^{r}, m=1\\
			\rvx_{s,m-1}^{(r,\tau_{m-1})}, m>1
		\end{cases}$
		\State \colorbox{pink!50}{Send $\rvx_{c,m-1}^{(r,\tau_{m-1})}$ to client $\tau_m$ (Centralized mode)} \Comment{Com.}
		\For{local update step $k = 0,\ldots, \tau_m-1$}
		\State Receive ($\mA_m^{(r,k)}$, $\mY_m^{(r,k)}$) from client $m$ \Comment{Com.}
		\State Execute forward passes with smashed data $\mA_m^{(r,k)}$
		\State Calculate the loss with ($\hat\mY_m^{(r,k)}$, $\mY_m^{(r,k)}$)
		\State Execute backward passes and compute $\nabla \ell_{\pi_m} (\rvx_{s,m}^{(r,k)}; \mA_m^{(r,k)})$
		\State Send $\nabla \ell_{\pi_m} (\mA_m^{(r,k)};\rvx_{s,m}^{(r,k)})$ to client $m$  \Comment{Com.}
		\State Update server-side local parameters:
		$\rvx_{s,m}^{(r,k+1)} \gets \rvx_{s,m}^{(r,k)} - \eta \nabla \ell_{\pi_m} (\rvx_{s,m}^{(r,k)}; \mA_m^{(r,k)})$
		\EndFor
		\State \colorbox{pink!50}{Receive $\rvx_{c,m}^{(r,\tau_m)}$ from client $\tau_m$ (Centralized mode)} \Comment{Com.}
		\EndFor
		\EndFor
		\EndProcedure		
	\end{algorithmic}
\end{algorithm}

\begin{algorithm}[ht]
	\caption{\texttt{SSL} (Client-side operations) \colorbox{gray!50}{Peer-to-peer mode}, \colorbox{pink!50}{Centralized mode}}\label{alg:detailed SSL 2}
	\begin{algorithmic}[1]
		\Ensure client-side global model parameters $\rvx_c^{r}= \rvx_{c,M}^{(r-1,\tau_M)}$
		\Procedure{\textbf{Client $\pi_m$} executes}{}
		\State Request the lasted client-side parameters from \colorbox{pink!50}{the server}, or \colorbox{gray!50}{the previous client} \Comment{Com.}
		\State Initialize client-side local parameters
		$\rvx_{c,m}^{(r,0)} \gets
		\begin{cases}
			\rvx_c^{r}, m=1\\
			\rvx_{c,m-1}^{(r,\tau_{m-1})}, m>1
		\end{cases}$
		\For{local update step $k = 0,\ldots, \tau_m-1$}
			\State Execute forward passes with data features $\mX_m^{(r,k)}$
			\State Send ($\mA_m^{(r,k)}$, $\mY_m^{(r,k)}$) to the server \Comment{Com.}
			\State Receive $\nabla \ell_{\pi_m} (\mA_m^{(r,k)};\rvx_{s,m}^{(r,k)})$ \Comment{Com.}
			\State Execute backward passes with $\nabla \ell_{\pi_m} (\mA_m^{(r,k)};\rvx_{s,m}^{(r,k)})$ and compute $\nabla \ell_{\pi_m} (\rvx_{c,m}^{(r,k)}; \mX_m^{(r,k)})$
			\State Update client-side local parameters:
		$\rvx_{c,m}^{(r,k+1)} \gets \rvx_{c,m}^{(r,k)} - \eta \nabla \ell_{\pi_m} (\rvx_{c,m}^{(r,k)}; \mX_m^{(r,k)})$
		\EndFor
		\State Send the updated client-side parameters $\rvx_{c,m}^{(r,\tau_m)}$ to \colorbox{pink!50}{the server}, or  \colorbox{gray!50}{the next client} \Comment{Com.}
		\EndProcedure
	\end{algorithmic}
\end{algorithm}

\clearpage
\section{Related work and limitations}\label{sec:related work}
\paragraph{Variants in SL.} SL is deemed as one promising paradigm for distributed learning at resource-constrained devices, given its computational efficiency on the client side. Most existing works focus on reducing the training delay arising from the relay-based training manner of vanilla SL in the multi-user scenario. \texttt{SFLV1} \citep{thapa2020splitfed} is one popular model parallel algorithm that combines the strengths of FL and SL, where each client has one corresponding instance of server-side model in the \textit{main server} to form a pair. Each pair constitutes a full model and conducts the local updates in parallel. After each training round, the \textit{fed server} collects and averages the updated clien-side model parameters. The aggregated client-side model will be disseminated to all the clients before next round. The \textit{main server} does the same operations to the instances of the server-side model (i.e., average the sever-side model parameters). \texttt{SFLV2} \citep{thapa2020splitfed}, \texttt{SFLV3} \citep{gawali2020comparison}, \texttt{SFLG} \citep{gao2021evaluation} and \texttt{FedSeq} \citep{zaccone2022speeding} are the variants of \texttt{SFLV1}.



\paragraph{Convergence analyses in SL.} The algorithm in \cite{han2021accelerating} reduces the latency and downlink communication on \texttt{SFLV1} by adding auxiliary networks at client-side for quick model updates. Their convergence analysis combines the theories of \cite{belilovsky2020decoupled} and \texttt{FedAvg}. \cite{wang2022fedlite} proposed \texttt{FedLite} to reduce the uplink communication overhead by compressing activations with product quantization and provided the convergence analysis of \texttt{FedLite}. However, their convergence recovers that of Minibatch SGD when there is no quantization. SGD with biased gradients \citep{ajalloeian2020convergence} is also related. However, it only converges to a neighborhood of the solution. As a result, the convergence of \texttt{SSL} on heterogeneous data is still lacking.

\paragraph{Limitations.} The limitations of our theory appearing in this paper are summarized: (i) As discussed in Section~\ref{sec:feasibility analysis}, the counterintuitive result that \texttt{SSL} outperforms \texttt{FedAvg} only hold in extremely heterogeneous settings (see Table~\ref{table:summary for strongly convex case}), as \texttt{FedAvg} exhibits a better performance than existing theories suggest in moderately heterogeneous settings \citep{wang2022unreasonable}. (ii) Some variants like \texttt{SCAFFOLD} \citep{karimireddy2020scaffold} show much better than \texttt{FedAvg}, which will be our future work. (iii) Other limitations will be same as the previous work in FL \citep{koloskova2020unified, karimireddy2020scaffold} as standard assumptions are used for our theory.

Moreover, we discuss the limitations of \texttt{SSL} itself. One primary issue of \texttt{SSL} due to the sequential manner is the training time overhead \citep{thapa2020splitfed}. Besides, \texttt{SSL} suffers from some privacy issues like label leakage \citep{li2021label}, which we believe is beyond the scope of this
work.

\section{More experimental details}\label{sec:apx exps}
In this section, we provide more details about the experiments. 

\subsection{More simulations}

\paragraph{Simulation validation on quadratic functions.} We adopt the one-dimensional quadratic functions as in \cite{karimireddy2020scaffold} to validate the analysis above. To extend the simulated experiments in Section~\ref{sec:feasibility analysis}, 9 groups of experiments with the same global objective $F(x)=\frac{1}{2}x^2$ are considered. The detailed settings are in Table~\ref{table:apx:simulation settings}.

Notably, we use $\delta = \norm{\frac{1}{M}\sum_{m=1}^Mx_m^\ast-x^\ast}$ and $\zeta_\ast^2 = \frac{1}{M}\sum_{m=1}^M\norm{\nabla F_m(x^\ast)}^2$ to measure heterogeneity, with a larger value meaning higher heterogeneity for both $\delta$ and $\zeta_\ast^2$, where $x_m^\ast$ is the local optimum of $F_m$ and $x^\ast$ is the global optimum.

\begin{figure}[htbp]
	\begin{minipage}[l]{0.05\linewidth}$\delta$=$0$\end{minipage}
	\centering
	\begin{subfigure}{0.3\linewidth}
		\centering
		$\zeta_\ast=1$
		\includegraphics[width=1\linewidth]{figs/quadratic_F10.50-1.00_F20.50-1.00_M2_R500.pdf}
	\end{subfigure}
	\begin{subfigure}{0.3\linewidth}
		\centering
		$\zeta_\ast=10$
		\includegraphics[width=1\linewidth]{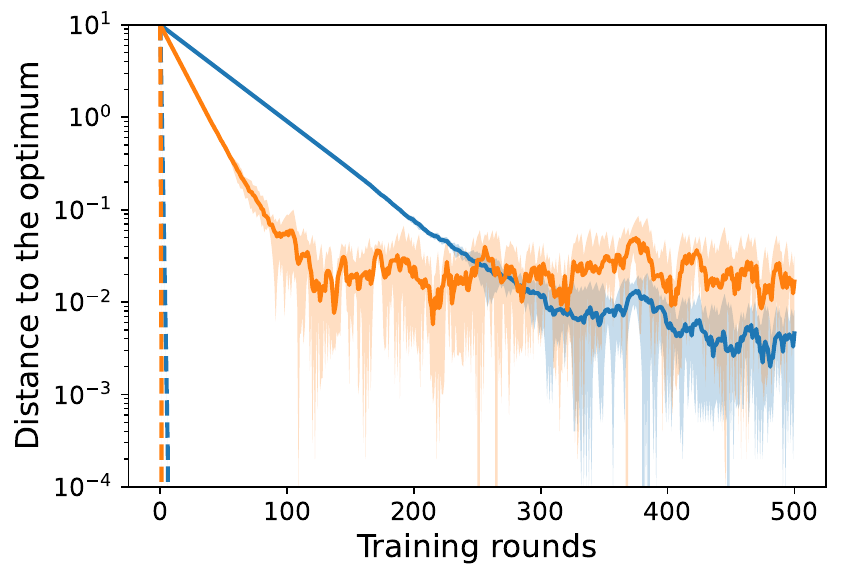}
	\end{subfigure}
	\begin{subfigure}{0.3\linewidth}
		\centering
		$\zeta_\ast=100$
		\includegraphics[width=1\linewidth]{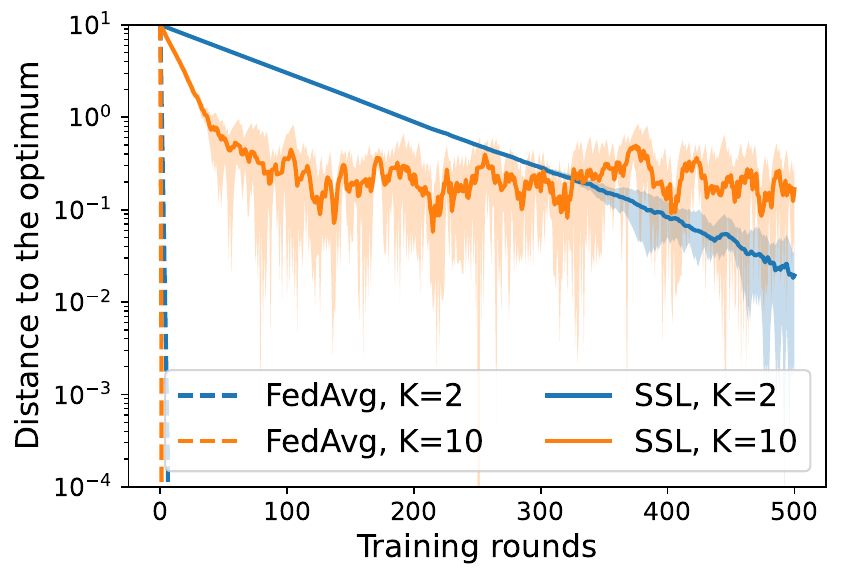}
	\end{subfigure}
	\begin{minipage}[l]{0.05\linewidth}$\delta$=$\frac{3}{8}$\end{minipage}
	\centering
	\begin{subfigure}{0.3\linewidth}
		\centering
		\includegraphics[width=1\linewidth]{figs/quadratic_F10.67-1.00_F20.33-1.00_M2_R500.pdf}
	\end{subfigure}
	\begin{subfigure}{0.3\linewidth}
		\centering
		\includegraphics[width=1\linewidth]{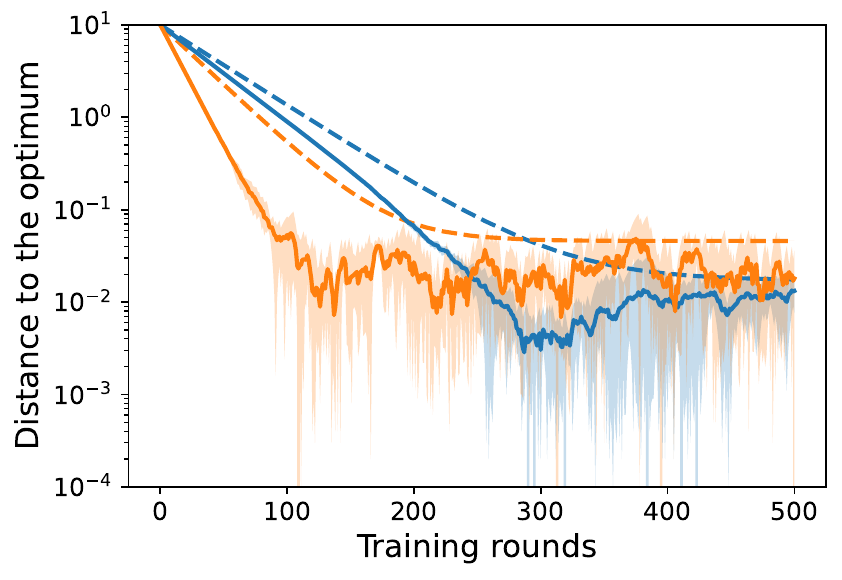}
	\end{subfigure}
	\begin{subfigure}{0.3\linewidth}
		\centering
		\includegraphics[width=1\linewidth]{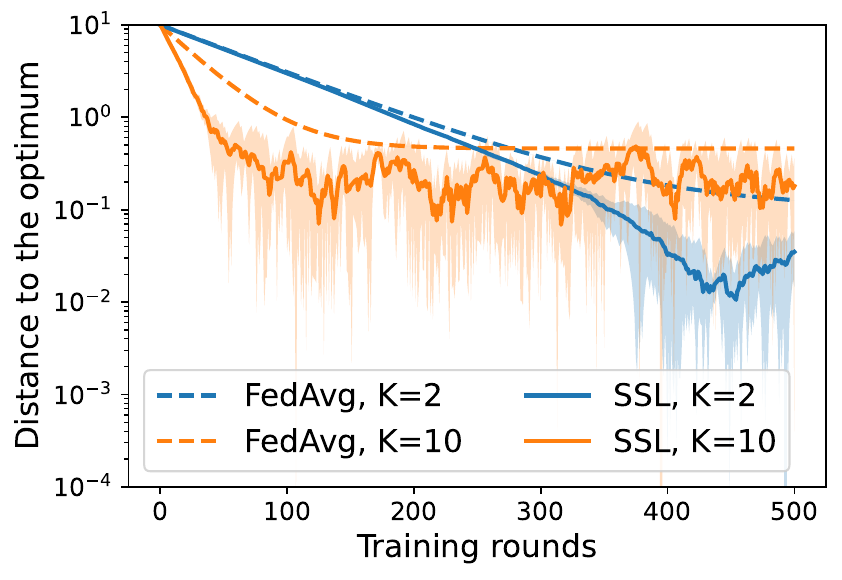}
	\end{subfigure}
	\begin{minipage}[l]{0.05\linewidth}$\delta$=$\infty$\end{minipage}
	\centering
	\begin{subfigure}{0.3\linewidth}
		\centering
		\includegraphics[width=1\linewidth]{figs/quadratic_F11.00-1.00_F20.00-1.00_M2_R500.pdf}
	\end{subfigure}
	\begin{subfigure}{0.3\linewidth}
		\centering
		\includegraphics[width=1\linewidth]{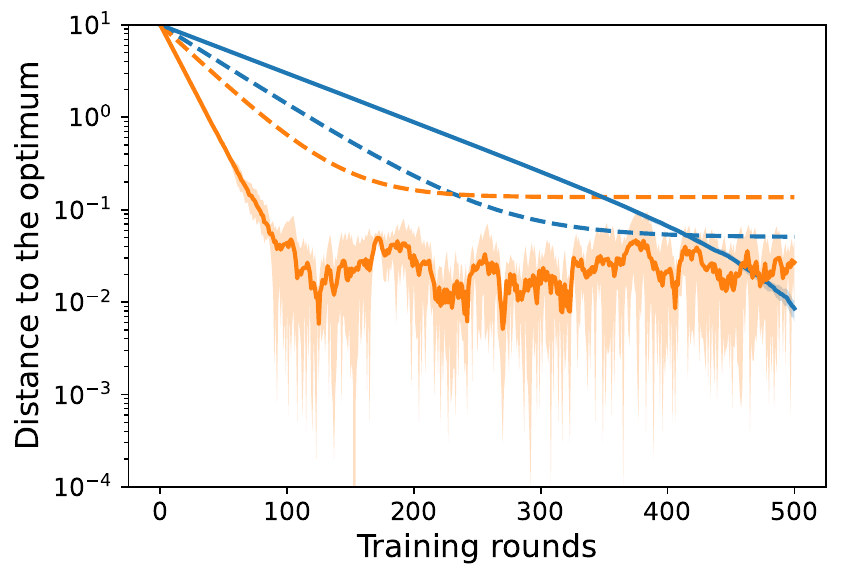}
	\end{subfigure}
	\begin{subfigure}{0.3\linewidth}
		\centering
		\includegraphics[width=1\linewidth]{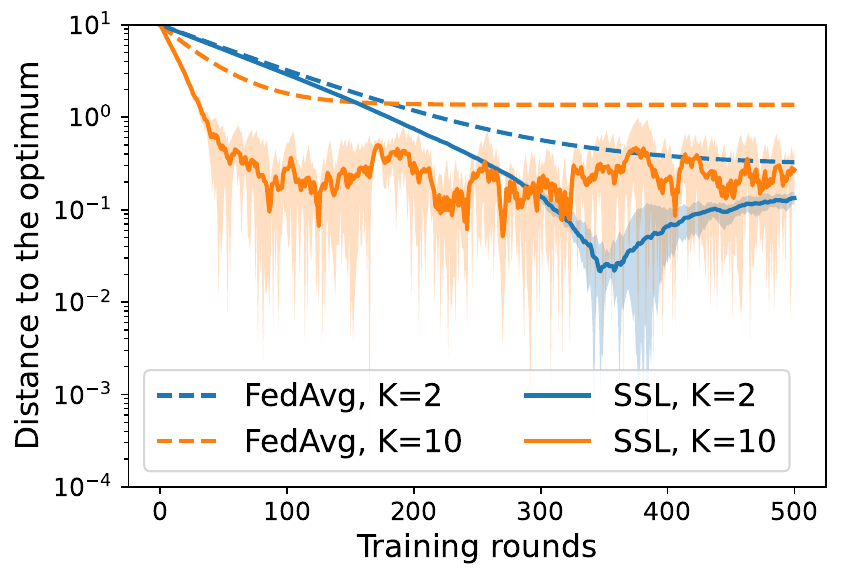}
	\end{subfigure}
	\caption{Simulations on quadratic functions. The first row shows the results of Groups 1-3 in Table~\ref{table:apx:simulation settings}. The second row shows the results of Groups 4-6. The third row shows the results of Groups 7-9. The best learning rates are chosen from [0.003, 0.006, 0.01, 0.03, 0.06, 0.1, 0.3, 0.6] with grid search. We run each experiments for 5 random seeds. Shaded areas show the min-max values.}
	\label{fig:apx:simulation}
\end{figure}

Figure~\ref{fig:apx:simulation} plots the results of \texttt{FedAvg} and \texttt{SSL} with different values of $\delta$ and $\zeta_\ast$. The results validate that \texttt{FedAvg} performs much better than \texttt{SSL} (also better than Theorem~\ref{thm:fedavg convergence} suggests) in moderately heterogeneous settings (see Groups 1-3, the first row in Figure~\ref{fig:apx:simulation}) and  worse than \texttt{SSL} in extremely heterogeneous settings (see Groups 4-9, the second and third rows in Figure~\ref{fig:apx:simulation}).
 
\begin{table}[h!]
	\renewcommand{\arraystretch}{1}
	\centering
	\caption{Settings of simulated experiments. Each group has two local objectives (i.e., $M=2$). We use $\delta = \norm{\frac{1}{M}\sum_{m=1}^Mx_m^\ast-x^\ast}$ and $\zeta_\ast^2 = \frac{1}{M}\sum_{m=1}^M\norm{\nabla F_m(x^\ast)}^2$ to measure heterogeneity, with a larger value meaning higher heterogeneity for both $\delta$ and $\zeta_\ast^2$, where $x_m^\ast$ is the local optimum of $F_m$ and $x^\ast$ is the global optimum.}
	\label{table:apx:simulation settings}
	\normalsize{
		\begin{tabular}{cllcc}
			\toprule
			Settings &$F_1$ &$F_2$ &$\delta$ &$\zeta_\ast$ \\
			\midrule
			Group 1 &$F_1(x) =\frac{1}{2}x^2 + x$ &$F_2(x) =\frac{1}{2}x^2 - x$ &$0$ &$1$ \\[1ex]
			Group 2 &$F_1(x) =\frac{1}{2}x^2 + 10x$ &$F_2(x) =\frac{1}{2}x^2 - 10x$ &$0$ &$10$ \\[1ex]
			Group 3 &$F_1(x) =\frac{1}{2}x^2 + 100x$ &$F_2(x) =\frac{1}{2}x^2 - 100x$ &$0$ &$100$ \\[1ex]
			\midrule
			Group 4 &$F_1(x) =\frac{2}{3}x^2 + x$ &$F_2(x) =\frac{1}{3}x^2 - x$ &$\frac{3}{8}$ &$1$ \\[1ex]
			Group 5 &$F_1(x) =\frac{2}{3}x^2 + 10x$ &$F_2(x) =\frac{1}{3}x^2 - 10x$ &$\frac{3}{8}$ &$10$ \\[1ex]
			Group 6 &$F_1(x) =\frac{2}{3}x^2 + 100x$ &$F_2(x) =\frac{1}{3}x^2 - 100x$ &$\frac{3}{8}$ &$100$ \\[1ex]
			\midrule
			Group 7 &$F_1(x) =x^2 + x$ &$F_2(x) = - x$ &$+\infty$ &$1$ \\[1ex]
			Group 8 &$F_1(x) =x^2 + 10x$ &$F_2(x) = - 10x$ &$+\infty$ &$10$ \\[1ex]
			Group 9 &$F_1(x) =x^2 + 100x$ &$F_2(x) = - 100x$ &$+\infty$ &$100$ \\[1ex]
			\bottomrule
	\end{tabular}}
\end{table}

\clearpage
\subsection{Extended Dirichlet partition}\label{subsec:extented dirichlet}

\paragraph{Baseline.} The partition strategy based on the Dirichlet distribution has been widely used in the FL literature \citep{yurochkin2019bayesian, hsu2019measuring, zhu2021data, wang2020tackling, jhunjhunwala2023fedexp}. The initial implementation, to the best of our knowledge, comes from \cite{yurochkin2019bayesian}. They partition the dataset into $M$ clients. They simulate a heterogeneous partition by drawing $\vp_c \sim \text{Dir}_M (\alpha) = \text{Dir}(\alpha \vq)$ and allocating a $\vp_{c,m}$ proportion of the samples of class $k$ to client $m$. Here $\vq$ is the prior distribution over $M$ clients, which is set as $\1 \in \R^{M}$.

In the FL literature, there is another common partition strategy, where each client only owns samples from $C$ classes ($C\leq C_{\text{total}}$ and $C_{\text{total}}$ classes in total) \citep{mcmahan2017communication, li2022federated}. This partition strategy is pathological, which is used commonly in the personalized FL.

\paragraph{Extended Dirichlet strategy.} These two strategies prompt us to come up with a new two-level strategy, which combing the properties of both, \textit{to generate arbitrarily heterogeneous data}. 


The difference is to add a step of allocating classes to determine the number of classes per client (denoted by $C$) before allocating samples via Dirichlet distribution (with concentrate parameter $\alpha$). Thus, the extended strategy can be denoted by $\text{ExDir}(C,\alpha)$. Suppose that there are $M$ clients. The implementation is as follows:
\begin{itemize}[leftmargin=1em]
	\item \textit{Allocating classes}. Select $C$ classes for each client until each class is allocated to at least one client. Then we can obtain the prior distribution $\vq_c \in \R^M$ over $M$ clients for any class $c$.
	\item \textit{Allocating samples}. For any class $c$, we draw $\vp_c\sim \text{Dir}(\alpha \vq_c)$ and then allocate a $\vp_{c,m}$ proportion of the samples of class $c$ to client $m$. For example, $\vq_c = [1, 1, 0, 0, \ldots,]$ means that the samples of class $c$ are only allocated to the first 2 clients.
\end{itemize}

This strategy have two levels, the first level to allocate classes and the second level to allocate samples. \textit{Note that $C$ decides the prior distribution, yet it does not mean that every client must own samples from $C$ classes}, which has shown in the following experiments. We note that \cite{reddi2020adaptive} use a two-level partition strategy to partition the CIFAR-100 dataset. They draw a multinomial distribution from the Dirichlet prior at the root ($\text{Dir}(\alpha)$) and a multinomial distribution from the Dirichlet prior at each coarse label ($\text{Dir}(\beta)$).

\begin{figure}[htbp]
	\centering
	\begin{subfigure}{0.325\linewidth}
		\centering
		\includegraphics[width=1\linewidth]{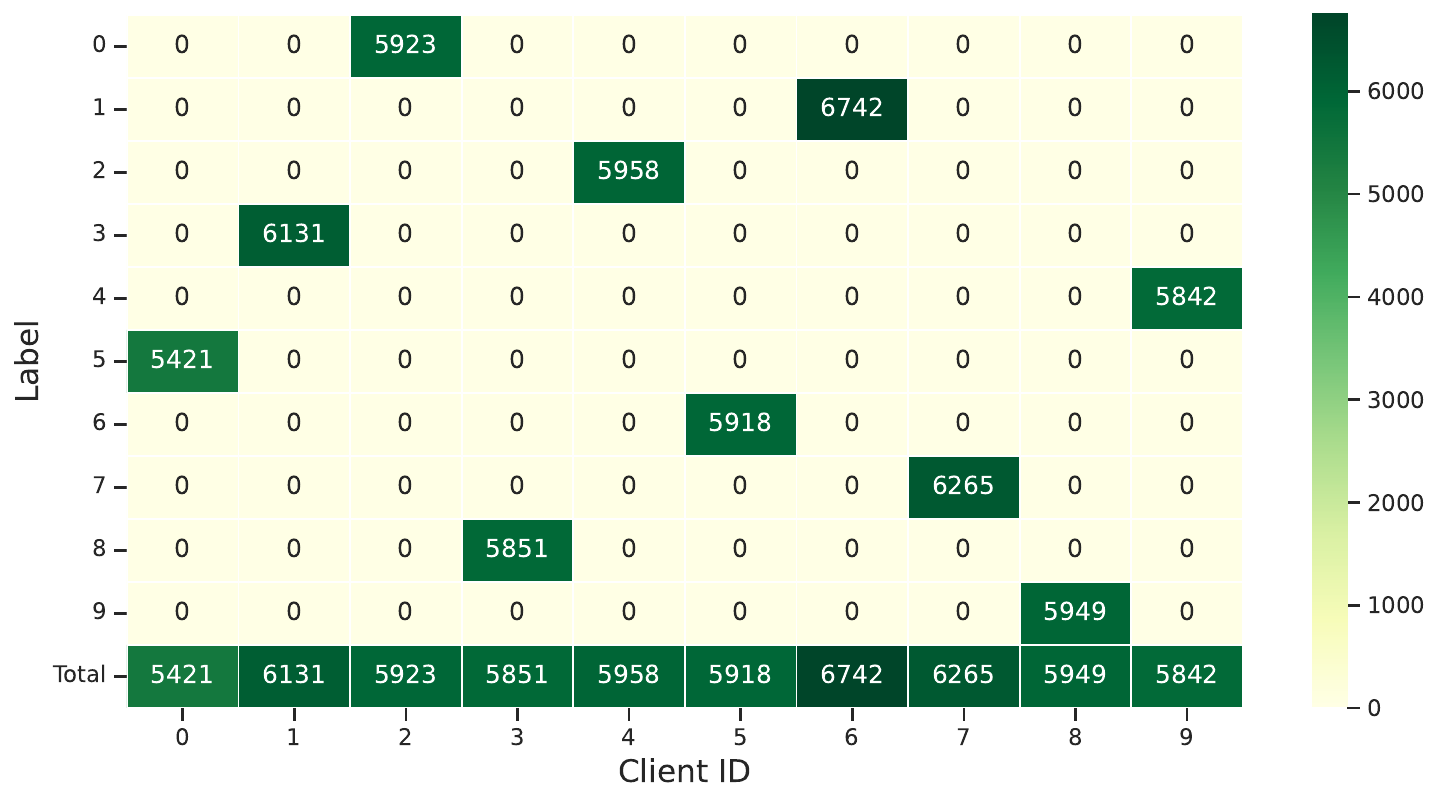}
		\caption{10 clients, $\text{ExDir}(1,100.0)$}
	\end{subfigure}
	\centering
	\begin{subfigure}{0.325\linewidth}
		\centering
		\includegraphics[width=1\linewidth]{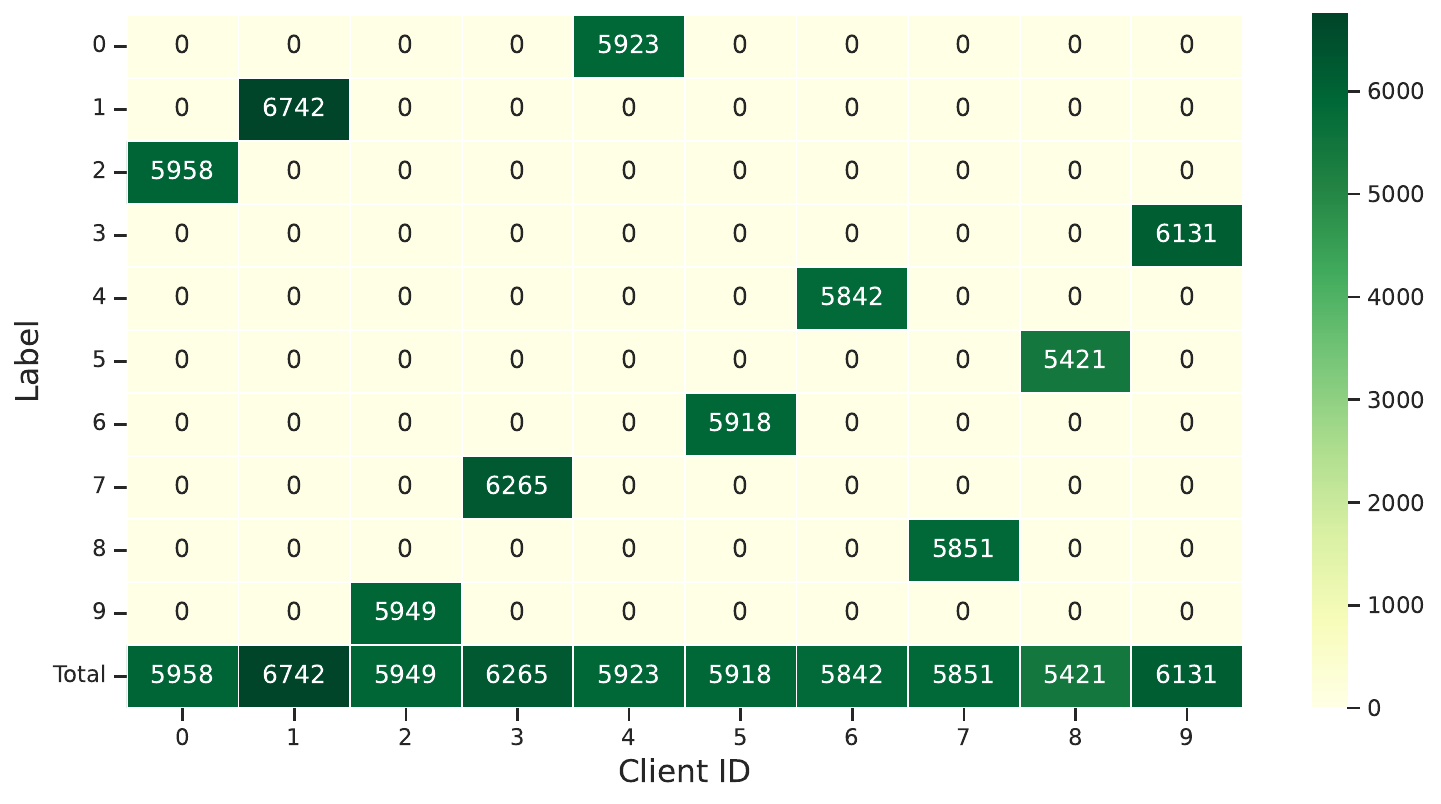}
		\caption{10 clients, $\text{ExDir}(1,10.0)$}
	\end{subfigure}
	\centering
	\begin{subfigure}{0.325\linewidth}
		\centering
		\includegraphics[width=1\linewidth]{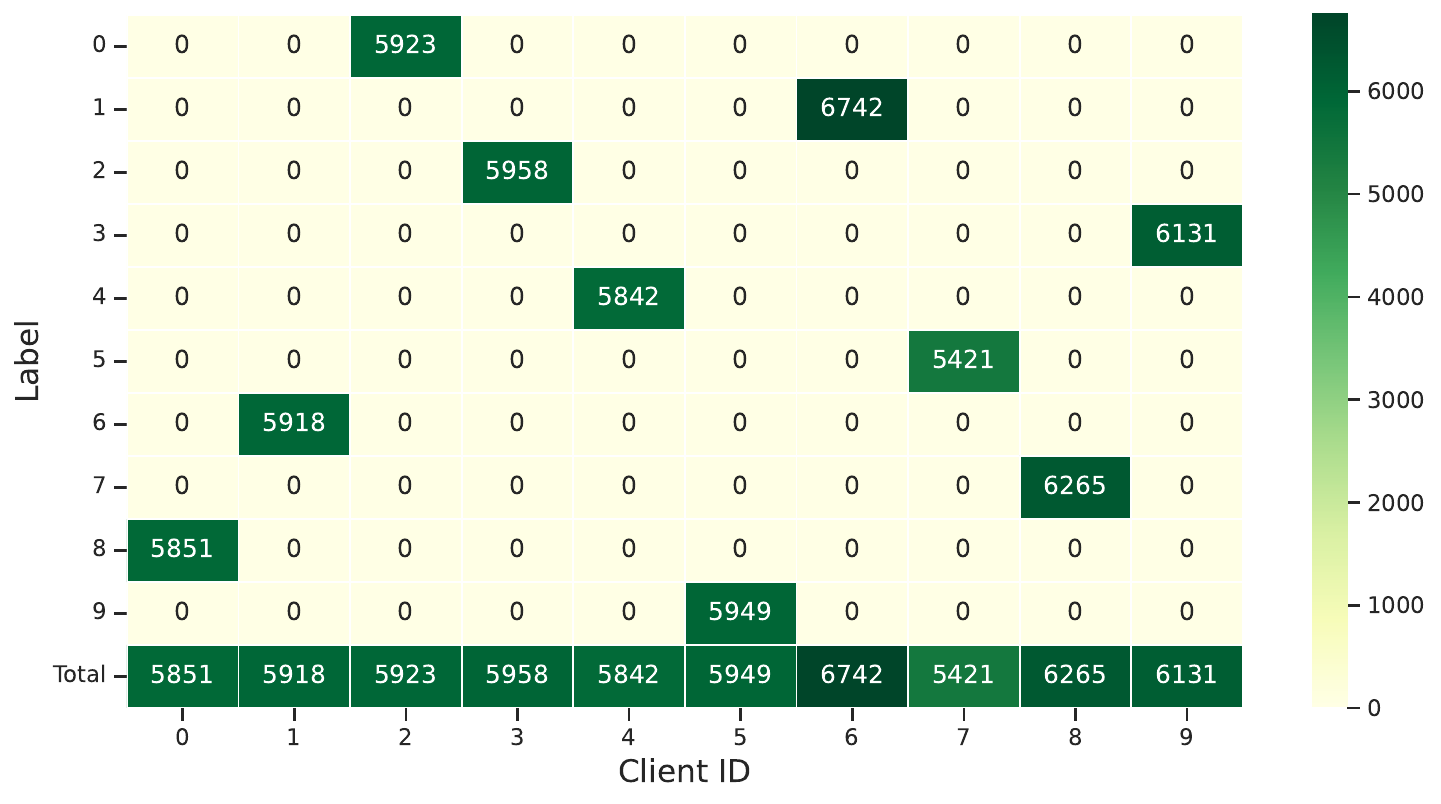}
		\caption{10 clients, $\text{ExDir}(1,1.0)$}
	\end{subfigure}	
	\centering
	\begin{subfigure}{0.325\linewidth}
		\centering
		\includegraphics[width=1\linewidth]{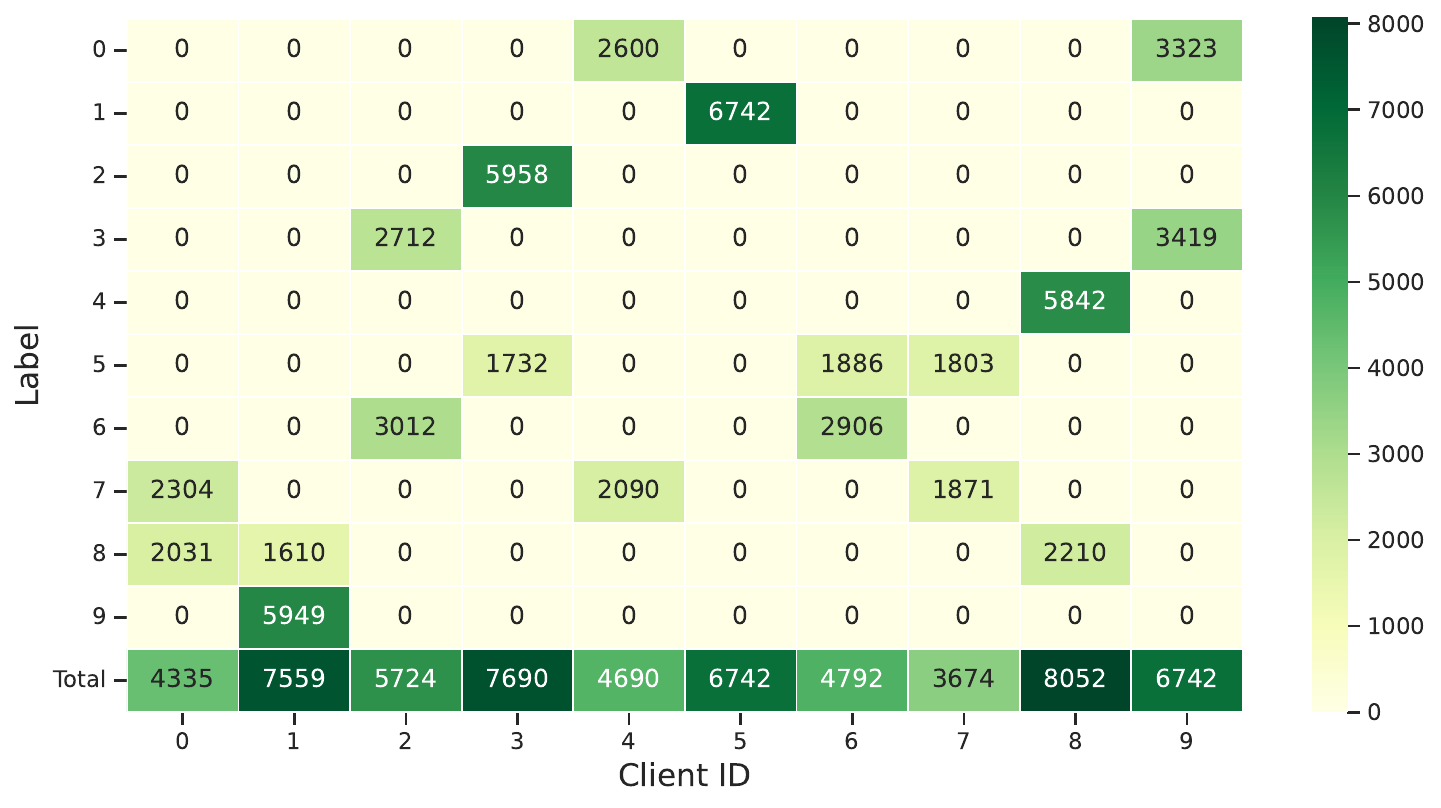}
		\caption{10 clients, $\text{ExDir}(2,100.0)$}
	\end{subfigure}
	\centering
	\begin{subfigure}{0.325\linewidth}
		\centering
		\includegraphics[width=1\linewidth]{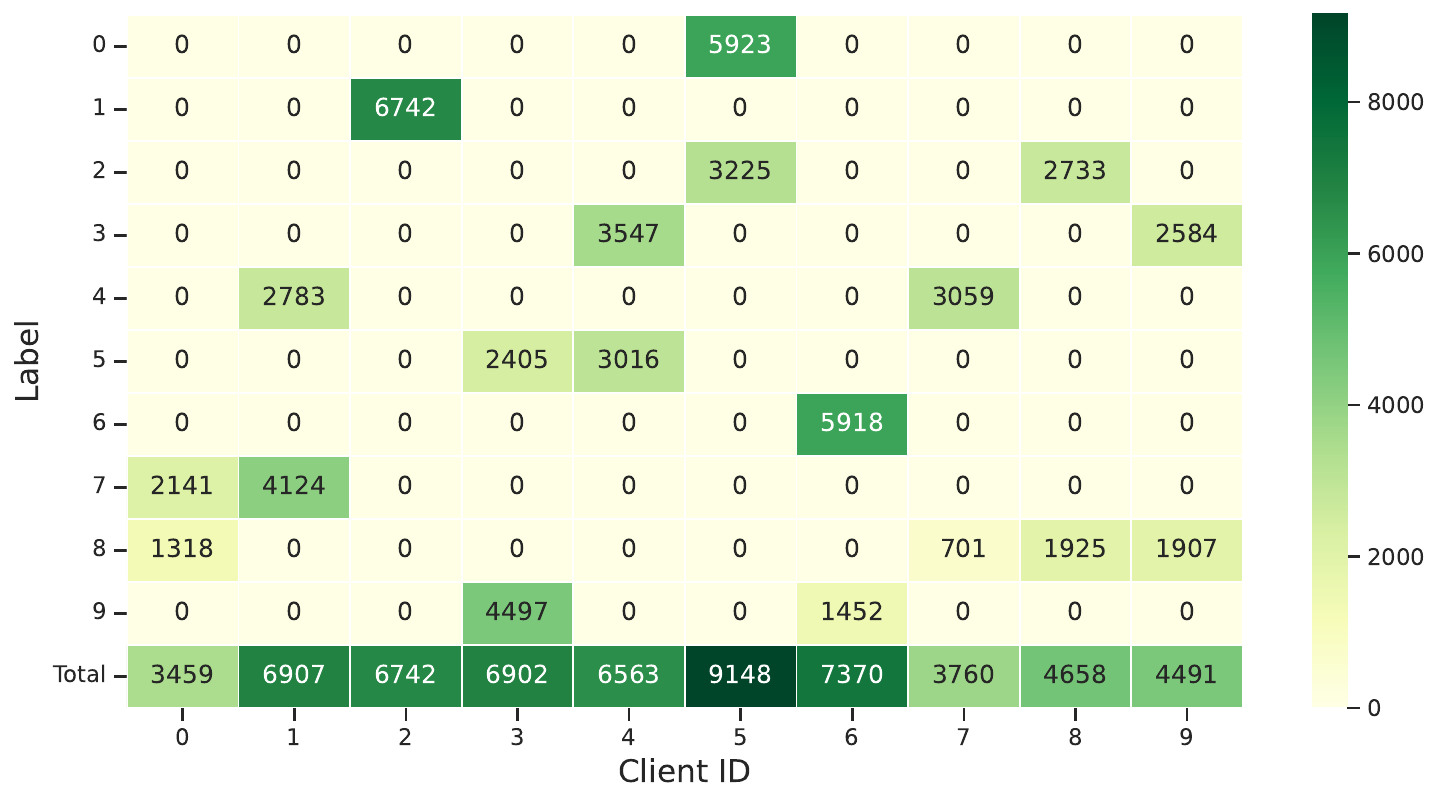}
		\caption{10 clients, $\text{ExDir}(2,10.0)$}
	\end{subfigure}
	\centering
	\begin{subfigure}{0.325\linewidth}
		\centering
		\includegraphics[width=1\linewidth]{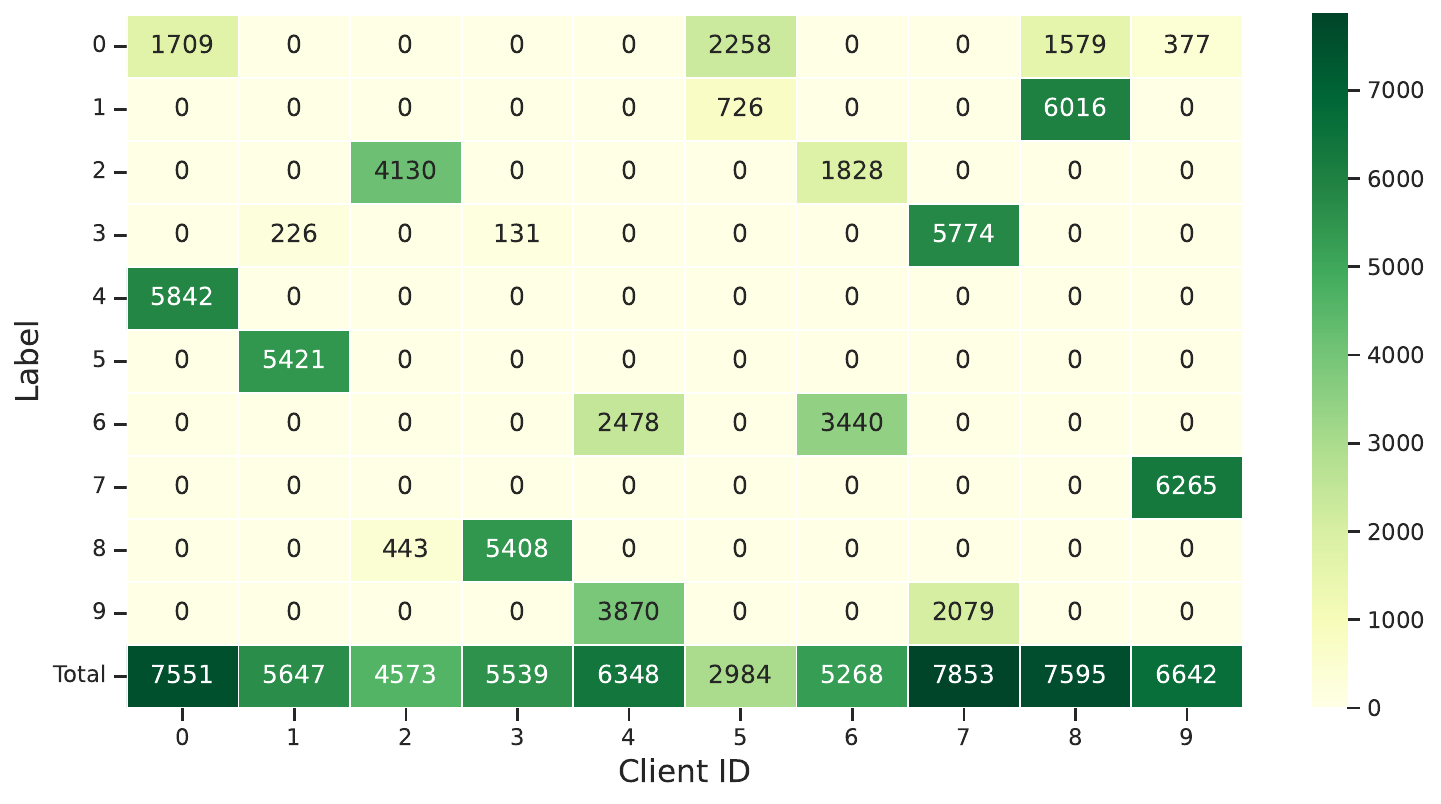}
		\caption{10 clients, $\text{ExDir}(2,1.0)$}
	\end{subfigure}	
	\centering
	\begin{subfigure}{0.325\linewidth}
		\centering
		\includegraphics[width=1\linewidth]{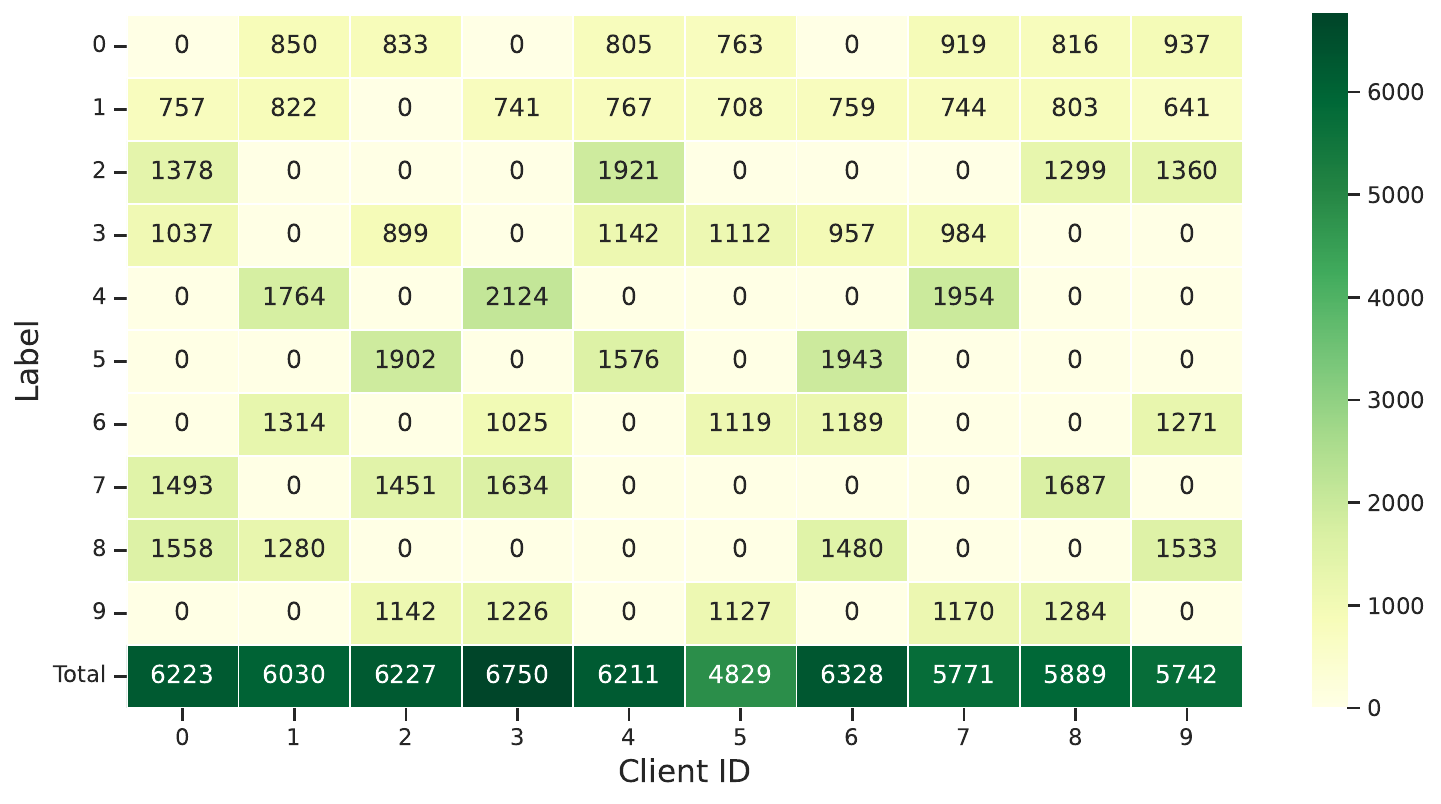}
		\caption{10 clients, $\text{ExDir}(5,100.0)$}
	\end{subfigure}
	\centering
	\begin{subfigure}{0.325\linewidth}
		\centering
		\includegraphics[width=1\linewidth]{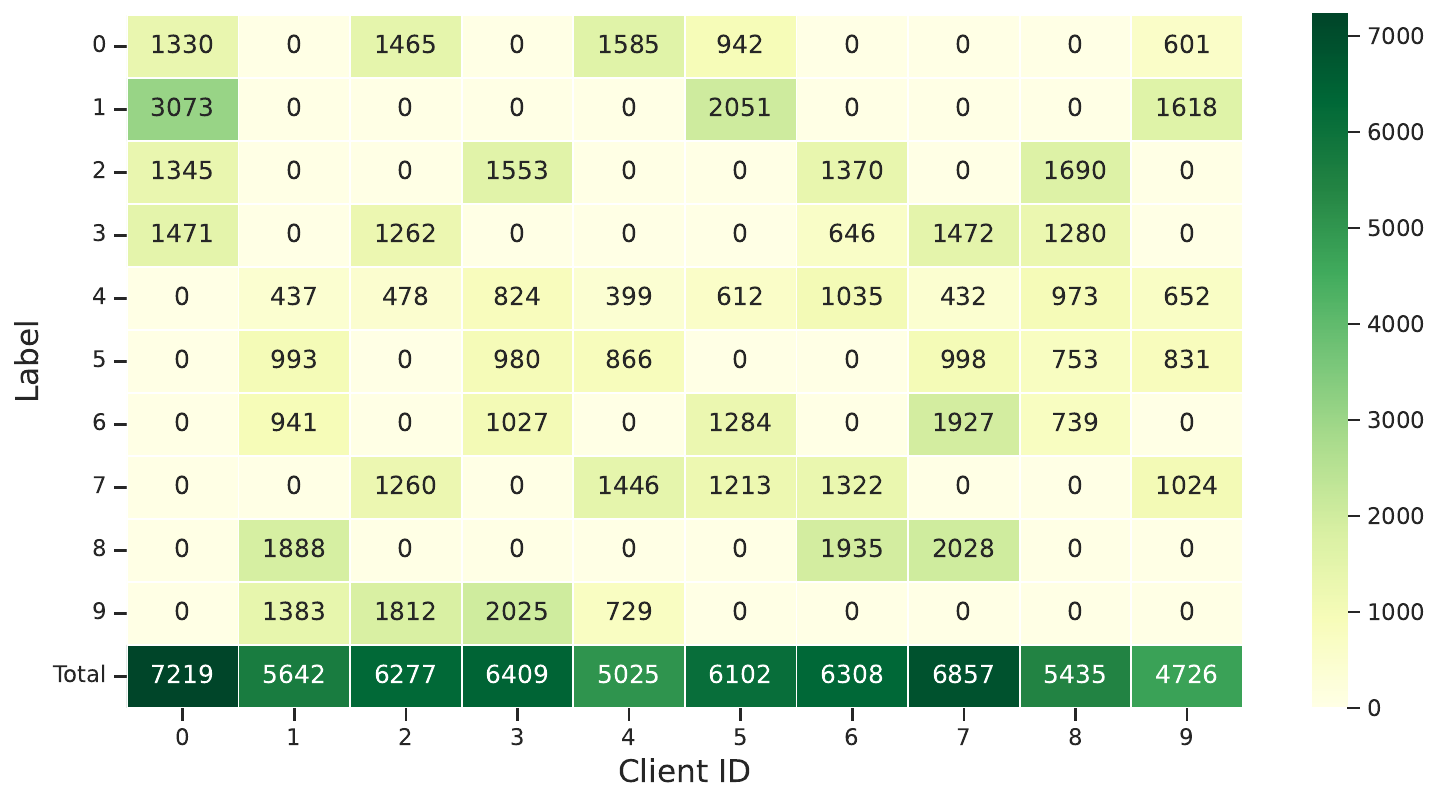}
		\caption{10 clients, $\text{ExDir}(5,10.0)$}
	\end{subfigure}
	\centering
	\begin{subfigure}{0.325\linewidth}
		\centering
		\includegraphics[width=1\linewidth]{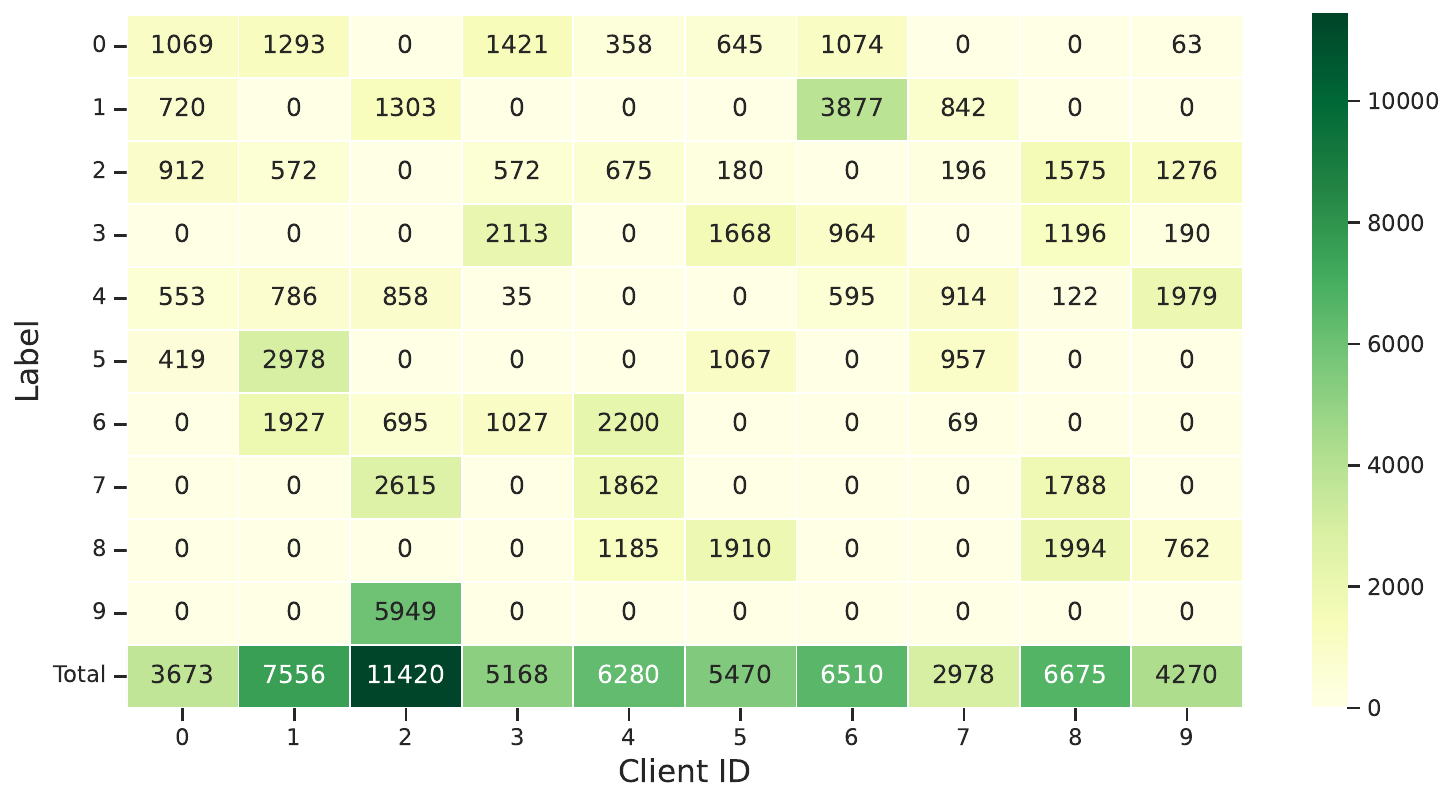}
		\caption{10 clients, $\text{ExDir}(5,1.0)$}
	\end{subfigure}	
	\centering
	\begin{subfigure}{0.325\linewidth}
		\centering
		\includegraphics[width=1\linewidth]{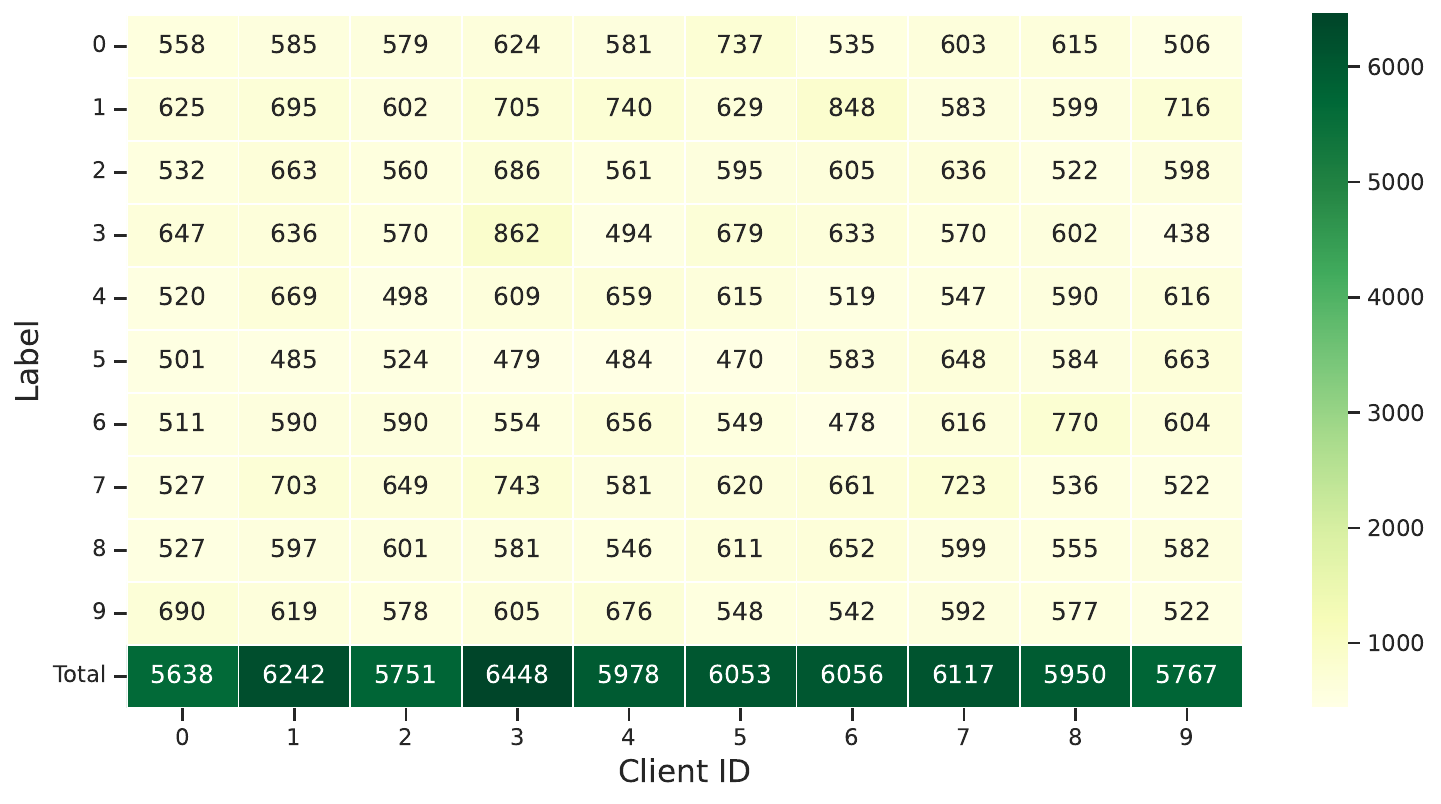}
		\caption{10 clients, $\text{ExDir}(10,100.0)$}
	\end{subfigure}
	\centering
	\begin{subfigure}{0.325\linewidth}
		\centering
		\includegraphics[width=1\linewidth]{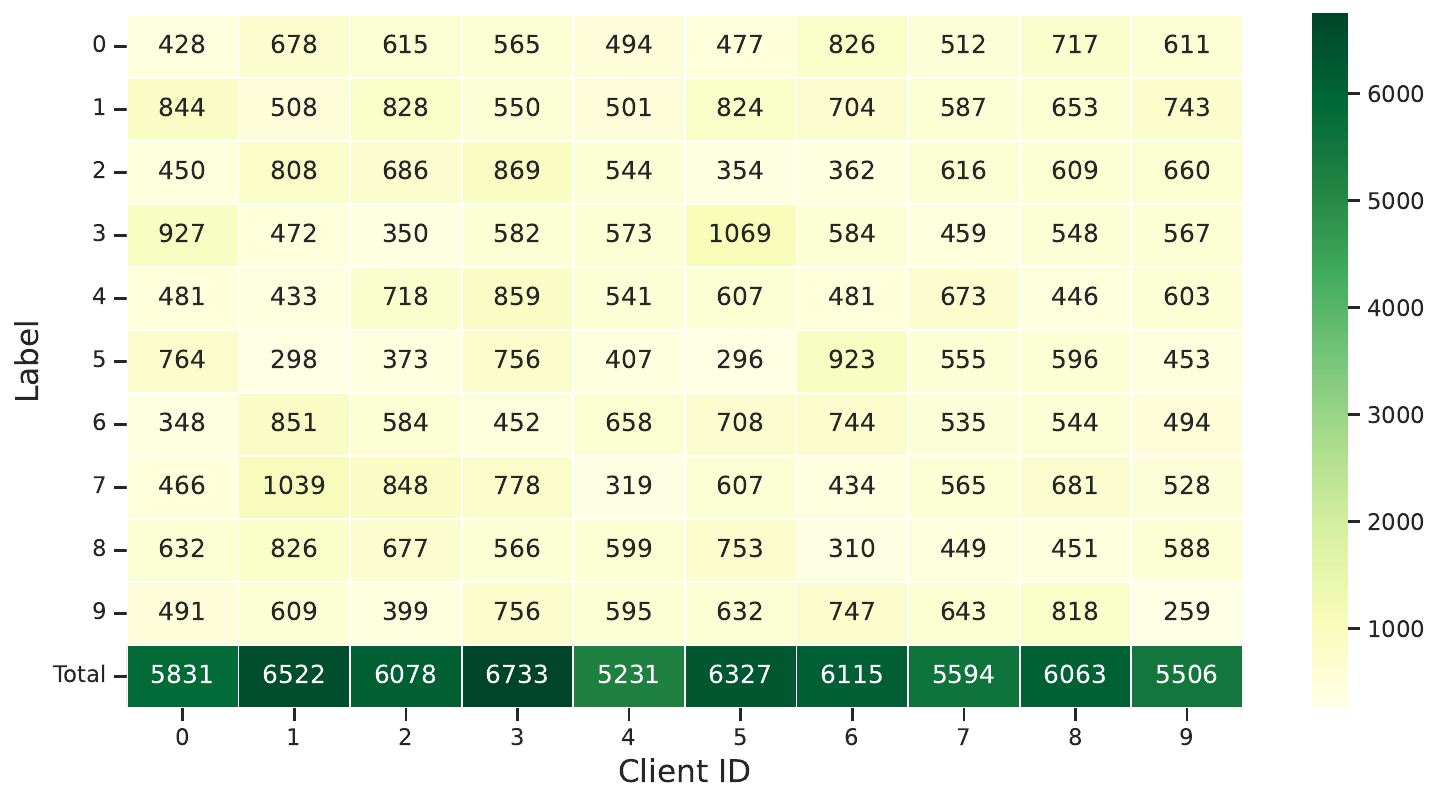}
		\caption{10 clients, $\text{ExDir}(10,10.0)$}
	\end{subfigure}
	\centering
	\begin{subfigure}{0.325\linewidth}
		\centering
		\includegraphics[width=1\linewidth]{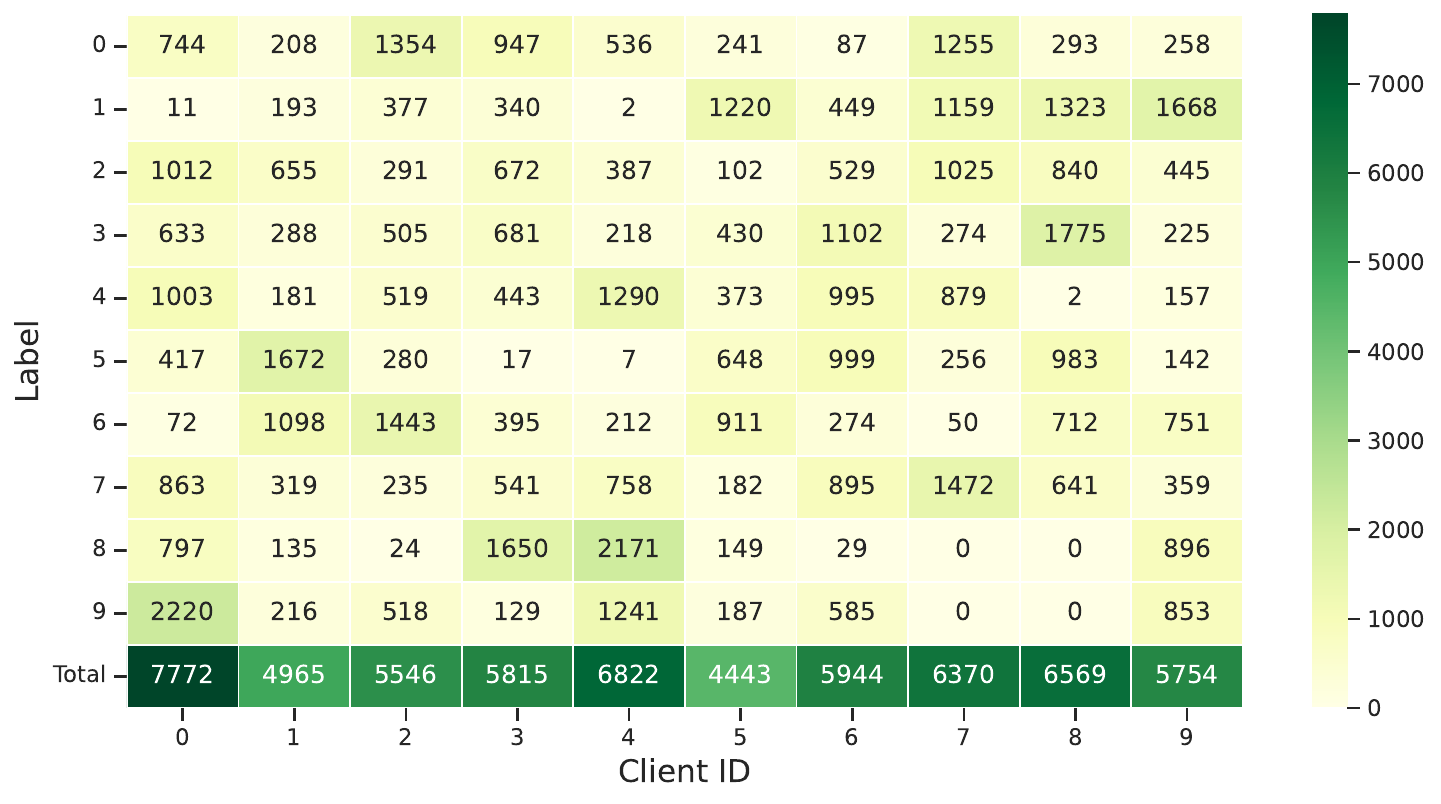}
		\caption{10 clients, $\text{ExDir}(10,1.0)$}
	\end{subfigure}	
	\caption{Visualization of data heterogeneity. The x-axis indicates client IDs and the y-axis indicates labels. The value in each cell is the number of data samples of a label belonging to that client. For the first row, there are only one possible results in the case where each client owns one label with 10 clients and 10 labels in total, so these three partitions are the same. For the second, third and forth rows, data heterogeneity increases from left to right.}
	\label{fig:extended dirichlet strategy}
\end{figure}

We use the Extended Dirichlet strategy to partition samples from MNIST to 10 clients. The results are shown in Figure~\ref{fig:extended dirichlet strategy}.


\subsection{Hyperparameter details}

\paragraph{Datasets and models.} We adopt the following models and datasets: (i) training LeNet-5 \citep{lecun1998gradient} on the MNIST dataset \citep{lecun1998gradient}; (ii) training LeNet-5 on the FMNIST dataset \citep{xiao2017fashion}; (iii) training VGG-11 \citep{simonyan2014very} on the CIFAR-10 dataset \citep{krizhevsky2009learning}. For SL, the LeNet-5 is split after the second 2D MaxPool layer, with 6\% of the entire model size retained in the client; the VGG-11 is split after the third 2D MaxPool layer, with 10\% of the entire model size at the client. Ideally, the split layer position has no effect on the performance of SL \citep{wang2022fedlite}.

\paragraph{Platform.} We conducted all experiments with three different seeds 1234, 666, 22. The experiments of MNIST and FMNIST are conducted on Nvidia GeForce RTX 3090 with cuda 11.7, python 3.7, pytorch 1.13.1. The experiments of CIFAR-10 are conducted on NVIDIA GeForce RTX 4090 (seed 1234), Nvidia GeForce RTX 3090 (seeds 666, 22) with cuda 12.0, python 3.10, pytorch, pytorch 2.0.0+cu118.

\begin{figure}[htbp]
	\begin{subfigure}{0.45\linewidth}
		\centering
		\includegraphics[width=1\linewidth]{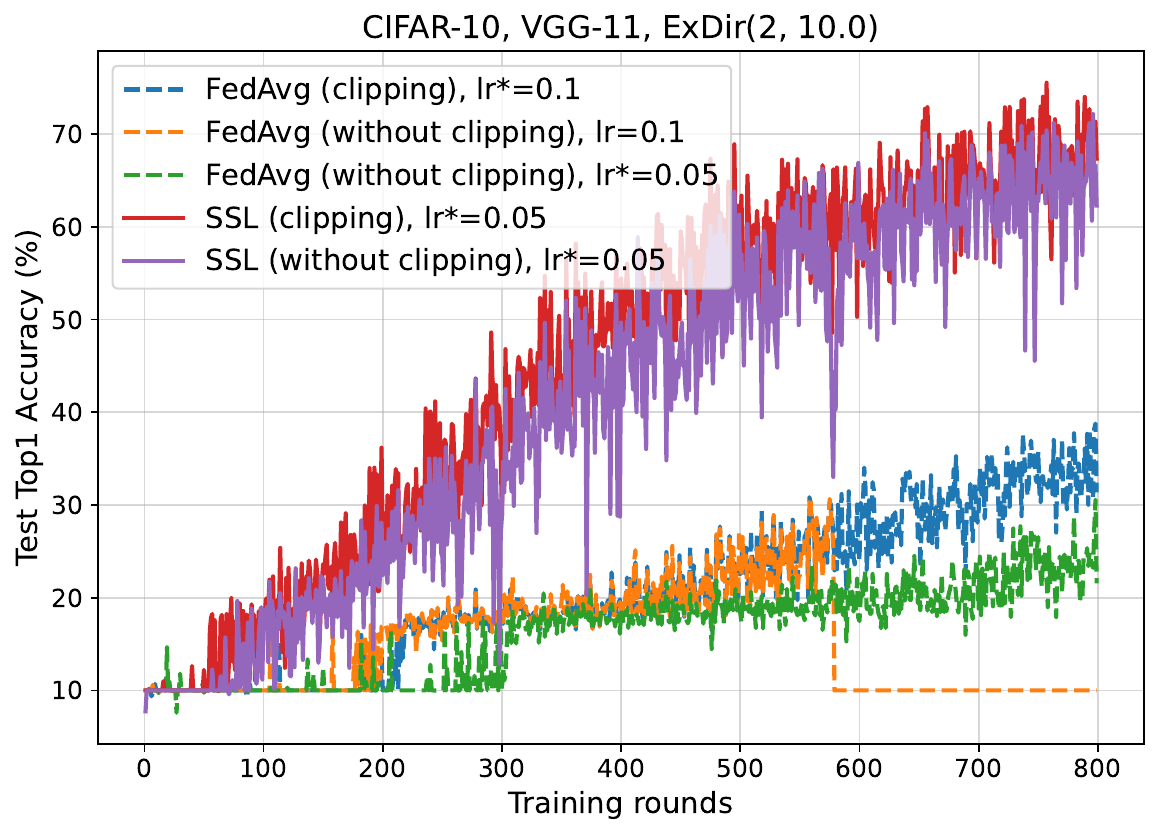}
		\caption{$\text{Exdir}(2, 10.0)$, $M=500$, $S=10$, $K=10$}
	\end{subfigure}
	\hspace{2em}
	\begin{subfigure}{0.45\linewidth}
		\centering
		\includegraphics[width=1\linewidth]{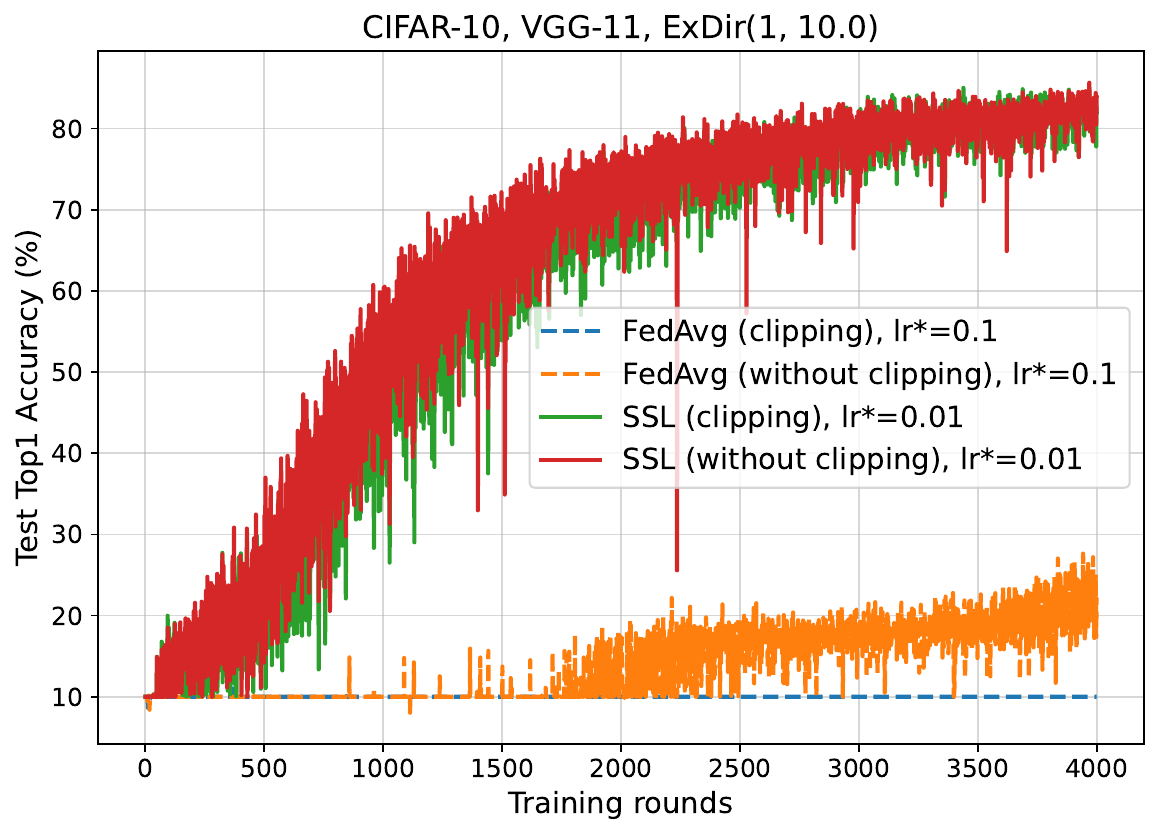}
		\caption{$\text{Exdir}(1, 10.0)$, $M=500$, $S=10$, $K=10$}
	\end{subfigure}
	\caption{Gradient clipping. ``lr'' in the legend represents the learning rate. ``lr*'' represents the best learning rate by the grid search. For the partitions $\text{ExDir}(2, 10.0)$, we see gradient clipping improves the stability. It can mitigate the gradient explosion (the loss turns ``Nan'' in our experiments). Yet for the partition $\text{ExDir}(1, 10.0)$, gradient clipping (max norm=10) hurts the convergence of both algorithms, especially \texttt{FedAvg}.}
	\label{fig:gradient clipping}
\end{figure}

\paragraph{Gradient clipping.} For $\text{ExDir}(2, 10.0)$, we use the use gradient clipping with max norm = 10, to improve stability of the algorithms as done in previous work \cite{acar2021federated, jhunjhunwala2023fedexp} (the left plot in Figure~\ref{fig:gradient clipping}). For $\text{ExDir}(1, 10.0)$, we do not use gradient clipping as it hurts the convergence (the right plot in Figure~\ref{fig:gradient clipping}).

\paragraph{Other hyperparameters} The local optimizer is SGD with momentem = 0 and weight decay = 1e-4. No learning rate decay is used for all experiments. We fix the number of participating clients to be 10 and the mini-batch size to be 20 by default.

\subsection{Grid search for learning rates}
For all the algorithms, we use the grid search to find the best learning rate. Note that one random seed ``1234'' is used for all grid searches. We summarize the best learning rates of different setups in Table~\ref{table:summary of best learning rates} and leave the details to the following subsubsections.

\begin{table}[ht]
	\renewcommand{\arraystretch}{1}
	\centering
	\caption{Best learning rates for different setups.}
	\label{table:summary of best learning rates}
	\begin{tabular}{cllllll}
		\toprule
		\multicolumn{3}{c}{\multirow{2}{*}{Setup}} &\multicolumn{2}{c}{$\text{ExDir}(1, 10.0)$} &\multicolumn{2}{c}{$\text{ExDir}(2, 10.0)$} \\ \cmidrule{4-7}
		& & &FedAvg  &SSL &FedAvg &SSL \\ \midrule
		\multirow{2}{*}{MNIST} &\multirow{2}{*}{$M=500$} &$K=10$ &0.001 &0.01 &0.05 &0.05 \\ 
		& &$K=20$ &0.0005 &0.01 &0.05 &0.05 \\ \midrule	
		\multirow{2}{*}{FMNIST} &\multirow{2}{*}{$M=500$} &$K=10$ &0.001 &0.01 &0.1 &0.1 \\ 
		& &$K=20$ &0.0005 &0.01 &0.1 &0.05 \\ \midrule
		\multirow{4}{*}{CIFAR-10} &\multirow{4}{*}{$M=10$} &$K=5$ &0.5 &0.01 &0.1 &0.05\\
		& &$K=10$ &0.1 &0.01 &0.1 &0.05 \\
		& &$K=20$ &0.1 &0.01 &0.1 &0.01 \\
		& &$K=30$ &0.1 &0.005 &0.1 &0.01\\ \midrule
		\multirow{4}{*}{CIFAR-10} &\multirow{4}{*}{$M=500$} &$K=5$ &0.1 &0.01 &0.5 &0.05 \\ 
		& &$K=10$ &0.1 &0.01 &0.1 &0.05 \\ 
		& &$K=20$ &0.1 &0.005 &0.1 &0.01 \\ 
		& &$K=30$ &- &- &0.1 &0.01 \\
		\bottomrule
	\end{tabular}
\end{table}

\clearpage
\subsection{More experimental results in cross-silo settings}

\begin{table}[ht]
	\renewcommand{\arraystretch}{1}
	\centering
	\caption{Test Accuracy results in cross-silo settings. We run 4000 rounds for both two partitions on CIFAR-10. Results are averaged across 3 random seeds and the last 20 rounds. The column 30\% and 75\% show the minimum number of rounds required to reach 30\% and 75\% accuracies.}
	\label{table:cross-silo settings}
	\resizebox{\linewidth}{!}{
		\begin{tabular}{llllllllll}
			\toprule
			\multicolumn{2}{c}{\multirow{3}{*}{Setup}} &\multicolumn{4}{c}{$\text{ExDir}(1, 10.0)$} &\multicolumn{4}{c}{$\text{ExDir}(2, 10.0)$} \\
			& &\multicolumn{2}{c}{FedAvg}  &\multicolumn{2}{c}{SSL} &\multicolumn{2}{c}{FedAvg} &\multicolumn{2}{c}{SSL} \\ \cmidrule(lr){3-6}\cmidrule(lr){7-10}
			& &ACC (\%) &30\%  &ACC (\%) &30\% &ACC (\%) &75\% &ACC (\%) &75\% \\ \cmidrule(lr){1-2}\cmidrule(lr){3-6}\cmidrule(lr){7-10}
			\multirow{4}{*}{CIFAR-10} 
			&$K=5$ &$69.28\pm{\color{gray}3.02}$ &956 
			&$84.79\pm{\color{gray}0.45}$ &495 
			&$80.51\pm{\color{gray}0.47}$ &2413 
			&$85.52\pm{\color{gray}0.58}$ &883\\
			
			&$K=10$ &$50.01\pm{\color{gray}4.06}$ &2254 &$85.64\pm{\color{gray}0.45}$ &401 
			&$80.91\pm{\color{gray}0.58}$ &1743 
			&$84.33\pm{\color{gray}0.66}$ &897\\
			
			&$K=20$ &$25.30\pm{\color{gray}2.72}$ &3516 &$84.75\pm{\color{gray}0.51}$ &712 
			&$81.24\pm{\color{gray}0.44}$ &1236 
			&$86.43\pm{\color{gray}0.44}$ &1096\\
			
			&$K=30$ &$32.07\pm{\color{gray}5.95}$ &3119 &$83.19\pm{\color{gray}0.79}$ &920 
			&$81.09\pm{\color{gray}0.52}$ &996 
			&$85.48\pm{\color{gray}0.61}$ &1186 \\
%
%
%
%
%
%
			\bottomrule
		\end{tabular}
	}
\end{table}

%

\begin{figure}[htbp]
	\centering
	\begin{subfigure}{0.45\linewidth}
		\centering
		\includegraphics[width=1\linewidth]{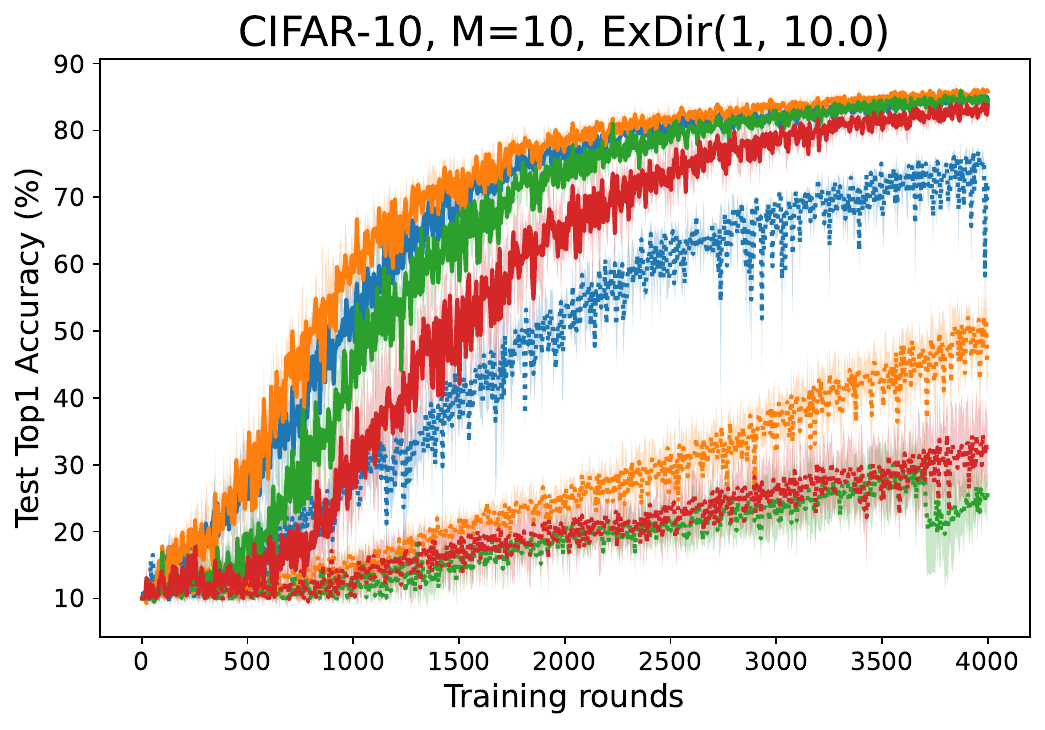}
	\end{subfigure}
	\hspace{2em}
	\begin{subfigure}{0.45\linewidth}
		\centering
		\includegraphics[width=1\linewidth]{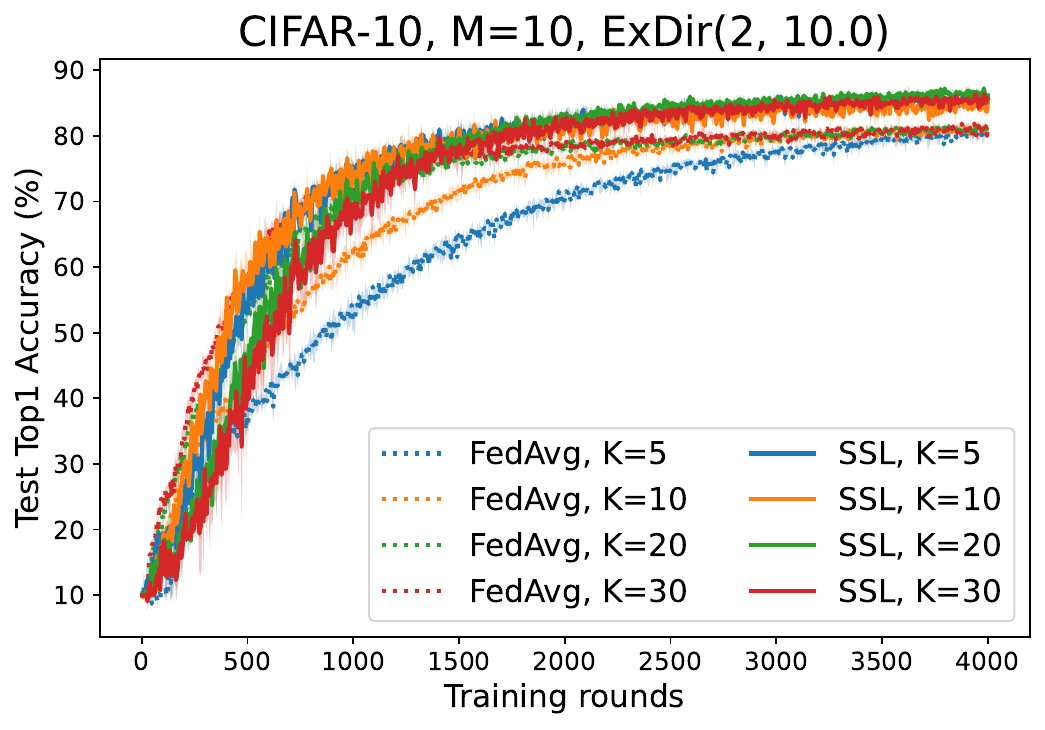}
	\end{subfigure}
	\caption{Test accuracy results with varying number of local steps on CIFAR-10 in cross-silo settings. We report the test accuracy results every 5 training rounds for clarity. The shaded areas show the standard deviation.}
	\label{fig:cifar10 cross-silo settings}
\end{figure}

\subsection{More experimental results in cross-device settings}

\begin{table}[ht]
	\renewcommand{\arraystretch}{1}
	\centering
	\caption{Test Accuracy results in cross-device settings. 1000 rounds are run for $\text{ExDir}(1, 10.0)$ and 400 rounds for $\text{ExDir}(2, 10.0)$ on MNIST, FMNIST. 4000 rounds are for both two partitions on CIFAR-10. Results are averaged across 3 random seeds and the last 20 rounds. The column 30\% and 75\% show the minimum number of rounds required to reach 30\% and 75\% accuracies.}
	\label{table:cross-device settings2}
	\resizebox{\linewidth}{!}{
		\begin{tabular}{llllllllll}
			\toprule
			\multicolumn{2}{c}{\multirow{3}{*}{Setup}} &\multicolumn{4}{c}{$\text{ExDir}(1, 10.0)$} &\multicolumn{4}{c}{$\text{ExDir}(2, 10.0)$} \\
			& &\multicolumn{2}{c}{FedAvg}  &\multicolumn{2}{c}{SSL} &\multicolumn{2}{c}{FedAvg} &\multicolumn{2}{c}{SSL} \\ \cmidrule(lr){3-6}\cmidrule(lr){7-10}
			& &ACC (\%) &30\%  &ACC (\%) &30\% &ACC (\%) &75\% &ACC (\%) &75\% \\ \cmidrule(lr){1-2}\cmidrule(lr){3-6}\cmidrule(lr){7-10}
			\multirow{4}{*}{CIFAR-10} 
			&$K=5$ &$29.06\pm{\color{gray}6.46}$ &3666 &$\textbf{81.34}\pm{\color{gray}1.35}$ &604 &$75.92\pm{\color{gray}1.17}$ &3210 &$\textbf{84.57}\pm{\color{gray}0.90}$ &1004\\
			&$K=10$ &$14.00\pm{\color{gray}5.66}$ &- &$\textbf{82.20}\pm{\color{gray}1.33}$ &483 &$81.17\pm{\color{gray}0.55}$ &2153 &$\textbf{82.95}\pm{\color{gray}1.70}$ &966\\
			&$K=20$ &$10.00\pm{\color{gray}0.00}$ &- &$\textbf{80.60}\pm{\color{gray}1.62}$ &673 &$81.26\pm{\color{gray}0.60}$ &1614 &$\textbf{85.24}\pm{\color{gray}0.79}$ &998\\
			&$K=30$ &- &- &- &- &$80.47\pm{\color{gray}0.69}$ &1618 &$\textbf{84.52}\pm{\color{gray}0.86}$ &966 \\\midrule
			\multirow{2}{*}{FMNIST}
			&$K=10$ &$39.90\pm{\color{gray}6.08}$ &492 &$\textbf{80.38}\pm{\color{gray}2.80}$ &19 &$\textbf{81.77}\pm{\color{gray}1.15}$ &127 &$78.25\pm{\color{gray}2.28}$ &162\\
			&$K=20$ &$38.78\pm{\color{gray}6.51}$ &490 &$\textbf{76.39}\pm{\color{gray}3.15}$ &25 &$\textbf{83.55}\pm{\color{gray}1.13}$ &87 &$76.58\pm{\color{gray}3.18}$ &103\\\midrule
			\multirow{2}{*}{MNIST}
			&$K=10$ &$56.43\pm{\color{gray}11.54}$ &274 &$\textbf{98.84}\pm{\color{gray}0.19}$ &16 &$98.15\pm{\color{gray}0.13}$ &25 &$98.57\pm{\color{gray}0.36}$ &10\\
			&$K=20$ &$51.86\pm{\color{gray}10.66}$ &274 &$\textbf{98.85}\pm{\color{gray}0.12}$ &14 &$98.35\pm{\color{gray}0.14}$ &18 &$98.56\pm{\color{gray}0.33}$ &8\\
			\bottomrule
		\end{tabular}
	}
\end{table}

\begin{figure}[htbp]
	\centering
	\begin{subfigure}{0.45\linewidth}
		\centering
		\includegraphics[width=1\linewidth]{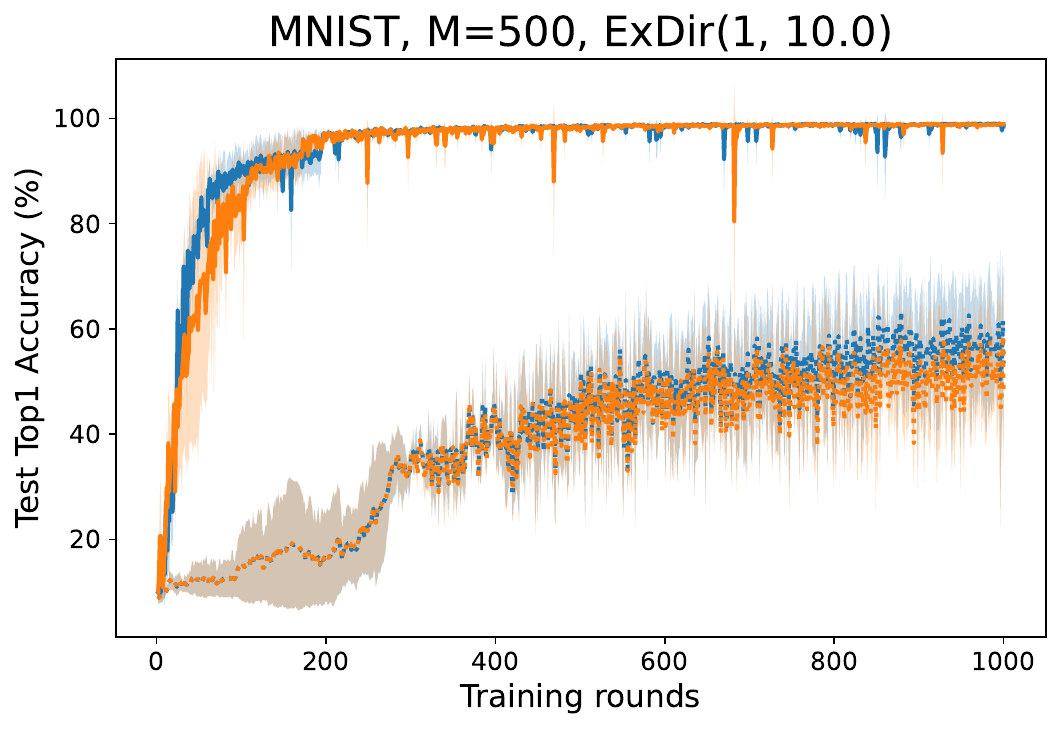}
	\end{subfigure}
	\hspace{2em}
	\begin{subfigure}{0.45\linewidth}
		\centering
		\includegraphics[width=1\linewidth]{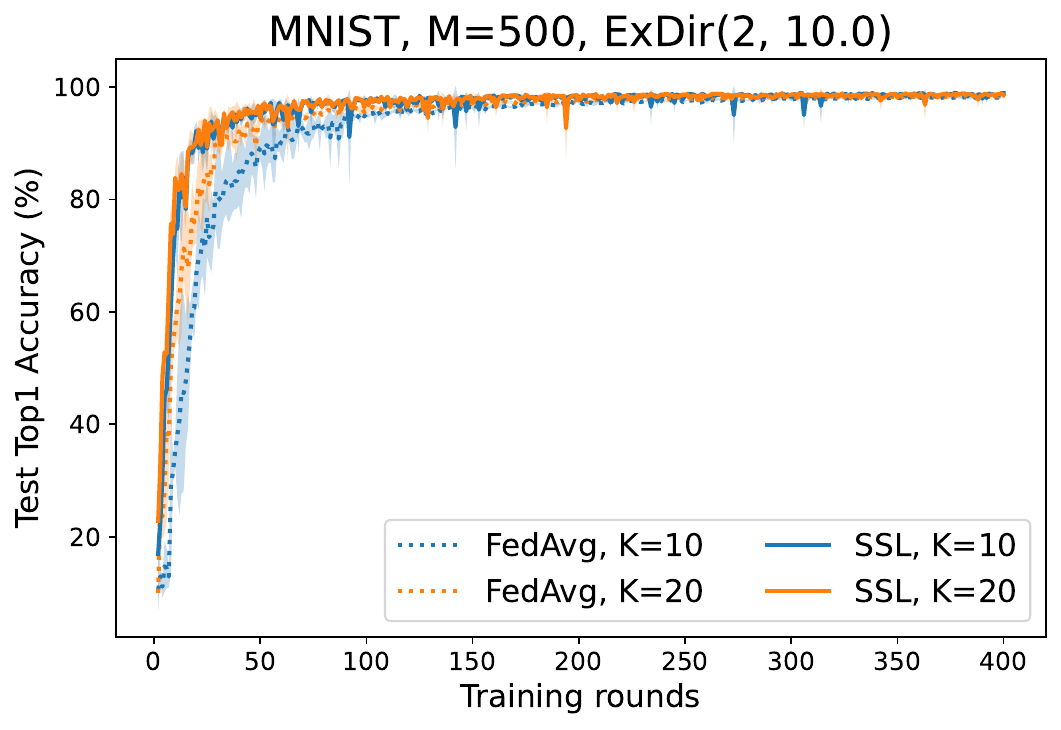}
	\end{subfigure}
	\caption{Test accuracy results with varying number of local steps on MNIST in cross-device settings. The shaded areas show the standard deviation.}
	\label{fig:mnist cross-device settings}
\end{figure}

\begin{figure}[htbp]
	\centering
	\begin{subfigure}{0.45\linewidth}
		\centering
		\includegraphics[width=1\linewidth]{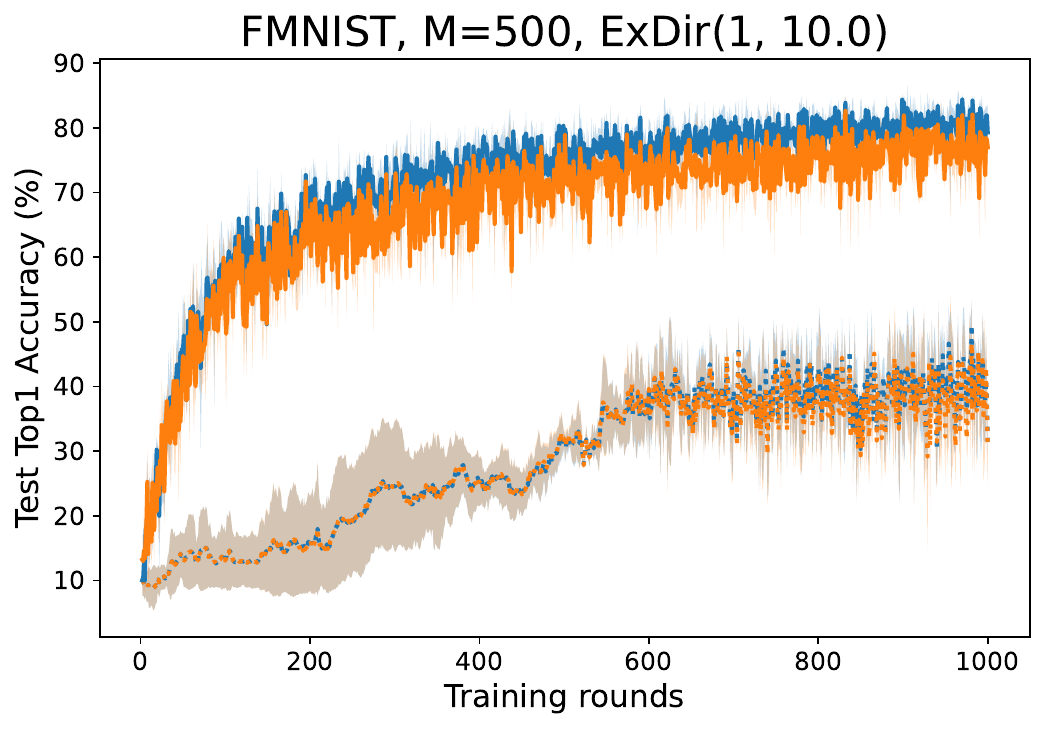}
	\end{subfigure}
	\hspace{2em}
	\begin{subfigure}{0.45\linewidth}
		\centering
		\includegraphics[width=1\linewidth]{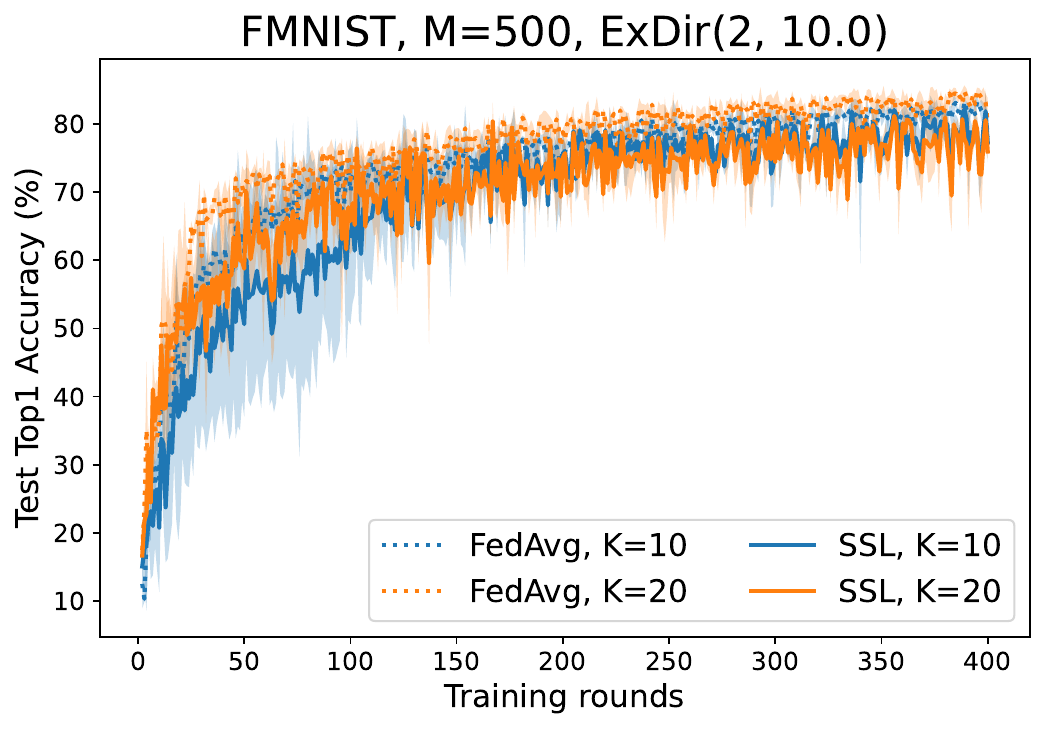}
	\end{subfigure}
	\caption{Test accuracy results with varying number of local steps on FMNIST in cross-device settings. The shaded areas show the standard deviation.}
	\label{fig:fashionmnist cross-device settings}
\end{figure}

\begin{figure}[htbp]
	\centering
	\begin{subfigure}{0.45\linewidth}
		\centering
		\includegraphics[width=1\linewidth]{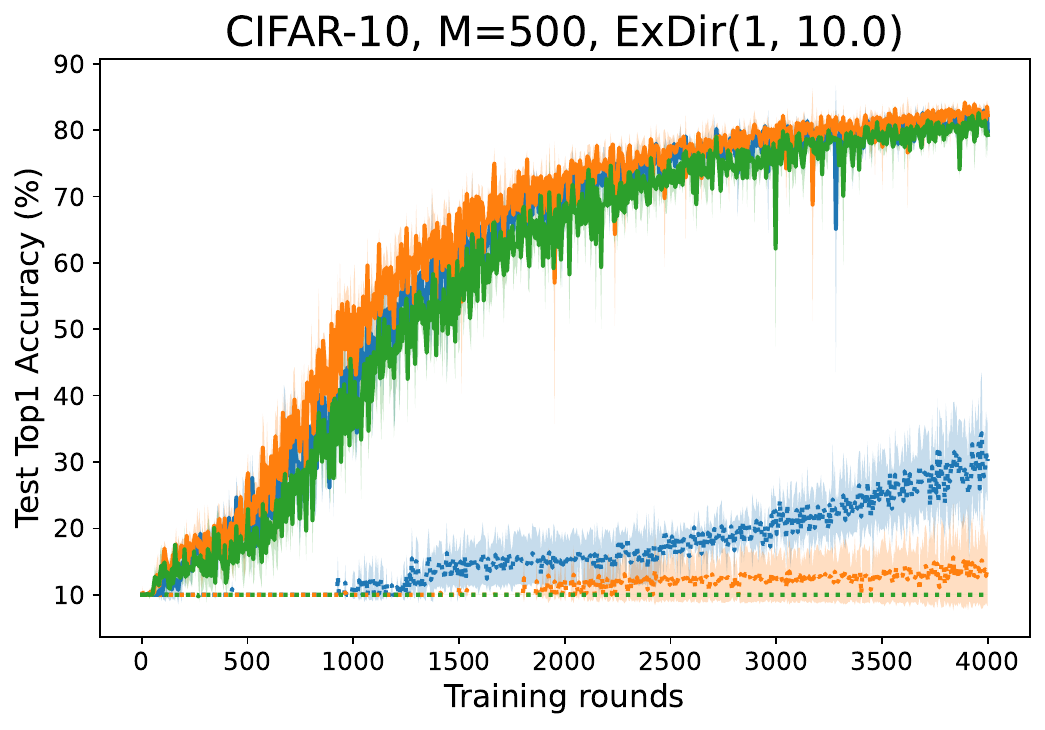}
	\end{subfigure}
	\hspace{2em}
	\begin{subfigure}{0.45\linewidth}
		\centering
		\includegraphics[width=1\linewidth]{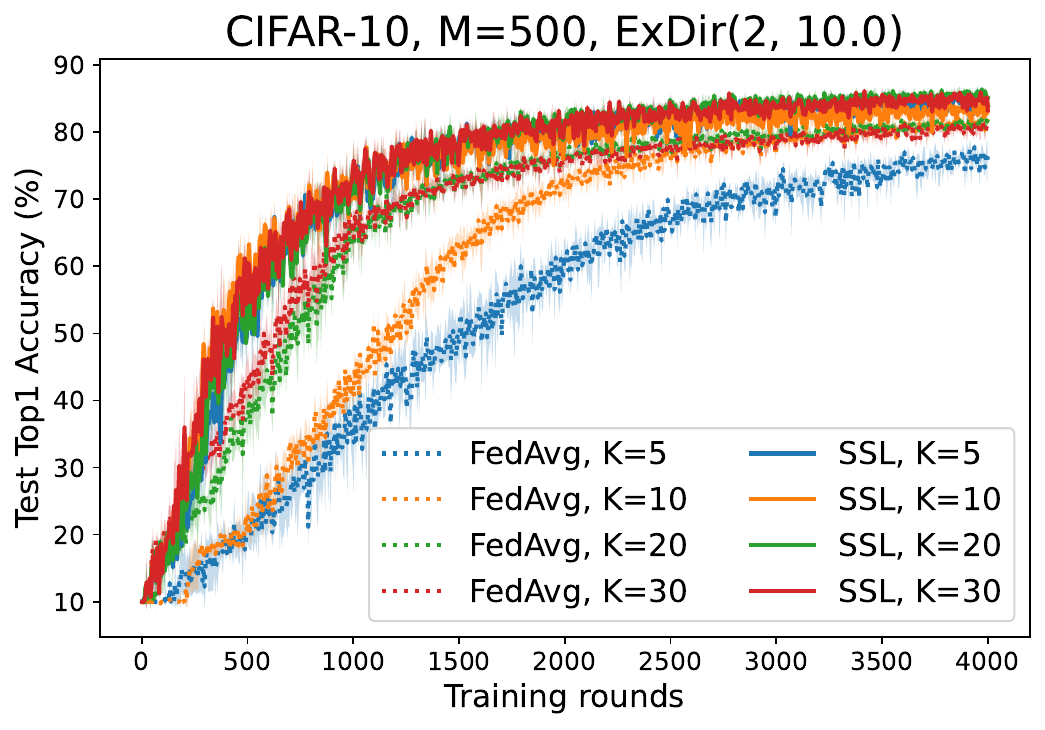}
	\end{subfigure}
	\caption{Test accuracy results with varying number of local steps on CIFAR-10 in cross-device settings. We report the test accuracy results every 5 training rounds for clarity. The shaded areas show the standard deviation.}
	\label{fig:cifar10 cross-device settings}
\end{figure}

\end{document}